\documentclass[runningheads]{llncs}
\usepackage{graphicx}
\usepackage[export]{adjustbox}
\usepackage{comment}
\usepackage{amsmath,amssymb} % define this before the line numbering.
\usepackage{booktabs}
\usepackage{xcolor}
\usepackage[caption=false]{subfig}
\usepackage{capt-of}
\usepackage{makecell}
\usepackage{hyperref}
\usepackage{blindtext}
\usepackage{fancyhdr}

\usepackage{array, multirow}
\newcolumntype{C}[1]{>{\centering\let\newline\\\arraybackslash\hspace{0pt}}m{#1}}

\usepackage[width=122mm,left=12mm,paperwidth=146mm,height=193mm,top=12mm,paperheight=217mm]{geometry}

\newcommand\vertarrowbox[3][6ex]{%
\begin{array}[t]{@{}c@{}} #2 \\
\left\uparrow\vcenter{\hrule height #1}\right.\kern-\nulldelimiterspace\\
\makebox[0pt]{\scriptsize#3}
\end{array}%
}

\newcommand{\bb}[1]{\mathbf{#1}}

\newcommand{\bx}{\bb{x}}

\newcommand{\by}{\bb{y}}

\newcommand{\bh}{\bb{h}}

\newcommand{\bu}{\bb{u}}

\newcommand{\bz}{\bb{z}}

\newcommand{\bT}{{\boldsymbol{\theta}}}

\newcommand{\pz}{p_{\bz}}
\newcommand{\pyGivenx}{p_{\by|\bx}}

\newcommand{\fT}{f_{\bT}}
\newcommand{\fTinv}{f_{\bT}^{-1}}
\newcommand{\gT}{g_{\bT}}

\newcommand{\downsc}{d_\downarrow   }

\newcommand{\parsection}[1]{\vspace{0.5mm}\noindent\textbf{#1:}~}

\fancypagestyle{firstpage}{%
\lhead{Accepted as Spotlight at ECCV 2020 (Extended Version)}
}

\begin{document}
\pagestyle{headings} \mainmatter \def\ECCVSubNumber{*} % Insert your submission number here

\title{SRFlow: Learning the Super-Resolution Space with Normalizing Flow}

\titlerunning{SRFlow}
\author{Andreas Lugmayr \and
Martin Danelljan \and
Luc Van Gool \and
Radu Timofte}

\authorrunning{ECCV 2020}
%\authorrunning{A. Lugmayr et al.}
\institute{Computer Vision Laboratory, ETH Zurich
\email{\{andreas.lugmayr,martin.danelljan,vangool,radu.timofte\}@vision.ee.ethz.ch}}
%******************

\maketitle
\thispagestyle{firstpage}

\begin{abstract}
Super-resolution is an ill-posed problem, since it allows for multiple predictions for a given low-resolution image. 
This fundamental fact is largely ignored by state-of-the-art deep learning based approaches.
These methods instead train a deterministic mapping using combinations of reconstruction and adversarial losses.
In this work, we therefore propose \textbf{SRFlow}: a normalizing flow based super-resolution method capable of learning the conditional distribution of the output given the low-resolution input.
Our model is trained in a principled manner using a single loss, namely the negative log-likelihood.
SRFlow therefore directly accounts for the ill-posed nature of the problem, and learns to predict diverse photo-realistic high-resolution images.
Moreover, we utilize the strong image posterior learned by SRFlow to design flexible image manipulation techniques, capable of enhancing super-resolved images by, e.g., transferring content from other images.
We perform extensive experiments on faces, as well as on super-resolution in general.
SRFlow outperforms state-of-the-art GAN-based approaches in terms of both PSNR and perceptual quality metrics, while allowing for diversity through the exploration of the space of super-resolved solutions.
Code and trained models will be available at: \href{https://www.git.io/SRFlow}{git.io/SRFlow}
\vspace{-2pt}

\end{abstract}

\section{Introduction}

\begin{figure}[b]
    \centering
    \includegraphics*[width=\textwidth]{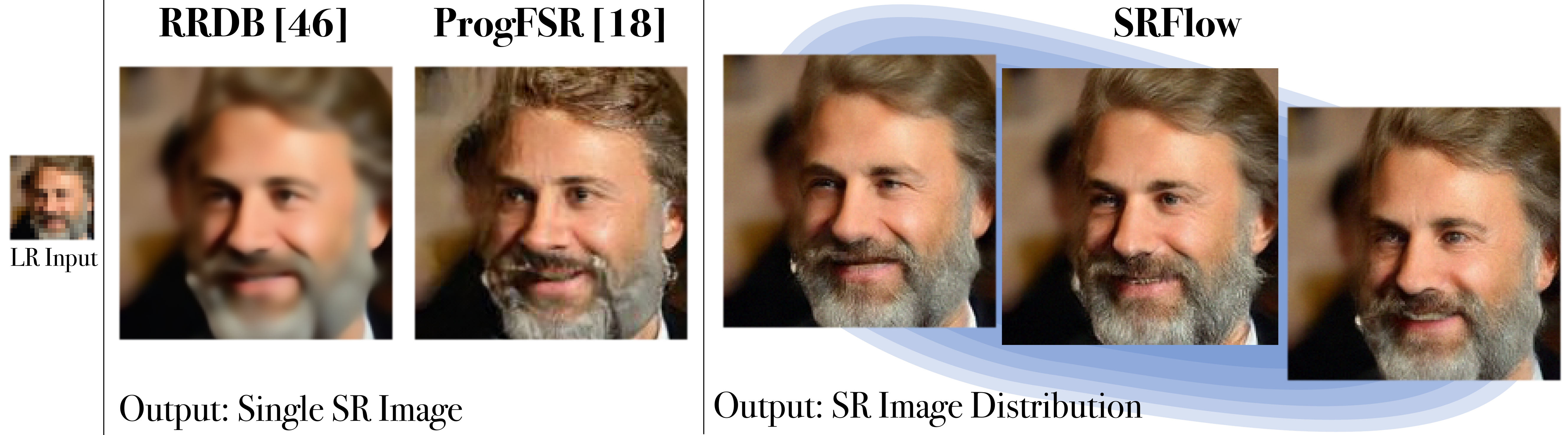}%
    \caption{While prior work trains a deterministic mapping, SRFlow learns the distribution of photo-realistic HR images for a given LR image. This allows us to explicitly account for the ill-posed nature of the SR problem, and to sample diverse images. ($8\times$ upscaling)}
    \label{fig:intro_sr_distribution}
\end{figure}

Single image super-resolution (SR) is an active research topic with several important applications. It aims to enhance the resolution of a given image by adding missing high-frequency information. Super-resolution is therefore a fundamentally ill-posed problem. In fact, for a given low-resolution (LR) image, there exist infinitely many compatible high-resolution (HR) predictions. This poses severe challenges when designing deep learning based super-resolution approaches.

Initial deep learning approaches~\cite{DongLHT14LearningDeepConv,dong2016image,kim2016accurate,lai2017deep,lim2017EDSR} employ feed-forward architectures trained using standard $L_2$ or $L_1$ reconstruction losses.
While these methods achieve impressive PSNR, they tend to generate blurry predictions.
This shortcoming stems from discarding the ill-posed nature of the SR problem.
The employed $L_2$ and $L_1$ reconstruction losses favor the prediction of an \emph{average} over the plausible HR solutions, leading to the significant reduction of high-frequency details.
To address this problem, more recent approaches~\cite{ahn2018image,haris2018deep,ledig2017photo,Sajjadi17EnhanceNet,wang2018esrgan,URDGN} integrate adversarial training and perceptual loss functions.
While achieving sharper images with better perceptual quality, such methods only predict a \emph{single} SR output, which does not fully account for the ill-posed nature of the SR problem.

We address the limitations of the aforementioned approaches by learning the conditional \emph{distribution} of plausible HR images given the input LR image. To this end, we design a conditional normalizing flow \cite{Dinh2017RealNVP,RezendeM15VarInferenceWithNF} architecture for image super-resolution. Thanks to the exact log-likelihood training enabled by the flow formulation, our approach can model expressive distributions over the HR image space. This allows our network to learn the generation of photo-realistic SR images that are consistent with the input LR image, without any additional constraints or losses. Given an LR image, our approach can sample multiple diverse SR images from the learned distribution. In contrast to conventional methods, our network can thus explore the space of SR images (see Fig.~\ref{fig:intro_sr_distribution}).

Compared to standard Generative Adversarial Network (GAN) based SR approaches~\cite{ledig2017photo,wang2018esrgan}, the proposed flow-based solution exhibits a few key advantages. First, our method naturally learns to generate diverse SR samples without suffering from mode-collapse, which is particularly problematic in the conditional GAN setting \cite{IsolaZZE17pix2pix,MathieuCL15VideoMSE}. Second, while GAN-based SR networks require multiple losses with careful parameter tuning, our network is stably trained with a single loss: the negative log-likelihood. Third, the flow network employs a fully invertible encoder, capable of mapping any input HR image to the latent flow-space and ensuring \emph{exact} reconstruction. This allows us to develop powerful image manipulation techniques for editing the predicted SR or any existing HR image. 

\parsection{Contributions}
We propose \textbf{SRFlow}, a flow-based super-resolution network capable of accurately learning the distribution of realistic HR images corresponding to the input LR image. In particular, the main contributions of this work are as follows: \textbf{(i)} We are the first to design a conditional normalizing flow architecture that achieves state-of-the-art super-resolution quality. 
\textbf{(ii)} We harness the strong HR distribution learned by SRFlow to develop novel techniques for controlled image manipulation and editing. \textbf{(iii)} Although only trained for super-resolution, we show that SRFlow is capable of image denoising and restoration. \textbf{(iv)} Comprehensive experiments for face and general image super-resolution show that our approach outperforms state-of-the-art GAN-based methods for both perceptual and reconstruction-based metrics.

\section{Related Work}

\parsection{Single image SR}
Super-resolution has long been a fundamental challenge in computer vision due to its ill-posed nature. Early learning-based methods mainly employed sparse coding based techniques~\cite{DaiTG15JointlyOptimizedRegressorsForSR,SunH12SRFromInternetScaleSceneMatching,YangWHM08SRAsSparseRepresentationOfRawPatches,YangWHM10SRViaSparseRep} or local linear regression~\cite{Timofte2014a+,Timofte13AnchNeighReg,YangY13SimpleFuncSR}. The effectiveness of example-based deep learning for super-resolution was first demonstrated by SRCNN~\cite{DongLHT14LearningDeepConv}, which further led to the development of more effective network architectures~\cite{dong2016image,kim2016accurate,lai2017deep,lim2017EDSR}. However, these methods do not reproduce the sharp details present in natural images due to their reliance on $L_2$ and $L_1$ reconstruction losses. This was addressed in URDGN~\cite{URDGN}, SRGAN~\cite{ledig2017photo} and more recent approaches~\cite{ahn2018image,haris2018deep,Sajjadi17EnhanceNet,wang2018esrgan} by adopting a conditional GAN based architecture and training strategy. While these works aim to predict \emph{one} example, we undertake the more ambitious goal of learning the distribution of \emph{all} plausible reconstructions from the natural image manifold.

\parsection{Stochastic SR}
The problem of generating diverse super-resolutions has received relatively little attention. This is partly due to the challenging nature of the problem. While GANs provide an method for learning a distribution over data~\cite{Goodfellow14GAN}, conditional GANs are known to be extremely susceptible to mode collapse since they easily learn to ignore the stochastic input signal~\cite{IsolaZZE17pix2pix,MathieuCL15VideoMSE}. Therefore, most conditional GAN based approaches for super-resolution and image-to-image translation resort to purely deterministic mappings~\cite{ledig2017photo,PathakKDDE16i2iInpainting,wang2018esrgan}.
A few recent works \cite{bahat2019explorableSR,buhler2020deepsee,menon2020pulse} address GAN-based stochastic SR by exploring techniques to avoid mode collapse and explicitly enforcing low-resolution consistency.
In contrast to those works, we design a flow-based architecture trained using the negative log-likelihood loss. This allows us to learn the conditional distribution of HR images, without any additional constraints, losses, or post-processing techniques to enforce low-resolution consistency.
A different line of research~\cite{Bell19InternalGAN,shaham2019singan,shocher2018zssr} exploit the internal patch recurrence by only training the network on the input image itself.
Recently~\cite{shaham2019singan} employed this strategy to learn a GAN capable of stochastic SR generation.
While this is an interesting direction, our goal is to exploit large image datasets to learn a general distribution over the image space.

\parsection{Normalizing flow}
Generative modelling of natural images poses major challenges due to the high dimensionality and complex structure of the underlying data distribution. While GANs~\cite{Goodfellow14GAN} have been explored for several vision tasks, Normalizing Flow based models~\cite{DinhKB14NICENonLinIndependentComponetsEst,Dinh2017RealNVP,KingmaD18Glow,RezendeM15VarInferenceWithNF} have received much less attention.
These approaches parametrize a complex distribution $p_\by(\by|\bT)$ using an invertible neural network $f_\bT$, which maps samples drawn from a simple (e.g.\ Gaussian) distribution $\pz(\bz)$ as $\by=f^{-1}_\bT (\bz)$. This allows the \emph{exact} negative log-likelihood $-\log p_\by(\by|\bT)$ to be computed by applying the change-of-variable formula. The network can thus be trained by directly minimizing the negative log-likelihood using standard SGD-based techniques. 
Recent works have investigated conditional flow models for point cloud generation~\cite{pumarola2020cflow,Yang2019PointFlow} as well as class~\cite{LiuLGWL19CondAdversarialGlow} and image~\cite{Ardizzone19cINNcolor,Winkler2020CondNormalizingFlowSR} conditional generation of images.
The latter works~\cite{Ardizzone19cINNcolor,Winkler2020CondNormalizingFlowSR} adapt the widely successful Glow architecture~\cite{KingmaD18Glow} to conditional image generation by concatenating the encoded conditioning variable in the affine coupling layers~\cite{DinhKB14NICENonLinIndependentComponetsEst,Dinh2017RealNVP}.
The concurrent work~\cite{Winkler2020CondNormalizingFlowSR} consider the SR task as an example application, but only addressing $2\times$ magnification and without comparisons with state-of-the-art GAN-based methods.
While we also employ the conditional flow paradigm for its theoretically appealing properties, our work differs from these previous approaches in several aspects.
Our work is first to develop a conditional flow architecture for SR that provides favorable or superior results compared to state-of-the-art GAN-based methods.
Second, we develop powerful flow-based image manipulation techniques, applicable for guided SR and to editing existing HR images.
Third, we introduce new training and architectural considerations.
Lastly, we demonstrate the generality and strength of our learned image posterior by applying SRFlow to image restoration tasks, unseen during training.

\section{Proposed Method: SRFlow}

We formulate super-resolution as the problem of learning a conditional probability distribution over high-resolution images, given an input low-resolution image. This approach explicitly addresses the ill-posed nature of the SR problem by aiming to capture the full diversity of possible SR images from the natural image manifold. To this end, we design a conditional normalizing flow architecture, allowing us to learn rich distributions using exact log-likelihood based training.

\subsection{Conditional Normalizing Flows for Super-Resolution}

The goal of super-resolution is to predict higher-resolution versions $\by$ of a given low-resolution image $\bx$ by generating the absent high-frequency details. While most current approaches learn a deterministic mapping $\bx \mapsto \by$, we aim to capture the full conditional distribution $\pyGivenx(\by | \bx, \bT)$ of natural HR images $\by$ corresponding to the LR image $\bx$. This constitutes a more challenging task, since the model must span a variety of possible HR images, instead of just predicting a single SR output. Our intention is to train the parameters $\bT$ of the distribution in a purely data-driven manner, given a large set of LR-HR training pairs $\{(\bx_i, \by_i)\}_{i=1}^M$.

The core idea of normalizing flow \cite{DinhKB14NICENonLinIndependentComponetsEst,RezendeM15VarInferenceWithNF} is to parametrize the distribution $\pyGivenx$ using an invertible neural network $\fT$. In the conditional setting, $\fT$ maps an HR-LR image pair to a latent variable $\bz = \fT(\by; \bx)$. We require this function to be invertible w.r.t.\ the first argument $\by$ for any LR image $\bx$. That is, the HR image $\by$ can always be exactly reconstructed from the latent encoding $\bz$ as $\by = \fT^{-1}(\bz; \bx)$. By postulating a simple distribution $\pz(\bz)$ (e.g.\ a Gaussian) in the latent space $\bz$, the conditional distribution $\pyGivenx(\by | \bx, \bT)$ is implicitly defined by the mapping $\by = \fT^{-1}(\bz; \bx)$ of samples $\bz \sim \pz$. The key aspect of normalizing flows is that the probability density $\pyGivenx$ can be explicitly computed as,
\begin{equation}
    \label{eq:conddens}
    \pyGivenx(\by | \bx, \bT) = \pz\big(\fT(\by; \bx)\big) \left| \det \frac{\partial\fT}{\partial\by}(\by; \bx) \right| \,.
\end{equation}
It is derived by applying the change-of-variables formula for densities, where the second factor is the resulting volume scaling given by the determinant of the Jacobian $\frac{\partial\fT}{\partial\by}$. The expression \eqref{eq:conddens} allows us to train the network by minimizing the negative log-likelihood (NLL) for training samples pairs $(\bx, \by)$,
\begin{equation}
    \label{eq:nll}
    \mathcal{L}(\bT; \bx, \by) = - \log \pyGivenx(\by | \bx, \bT) = -\log \pz\big(\fT(\by; \bx)\big) - \log \left| \det \frac{\partial\fT}{\partial\by}(\by; \bx) \right| \,.
\end{equation}
HR image samples $\by$ from the learned distribution $\pyGivenx(\by | \bx, \bT)$ are generated by applying the inverse network $\by = \fT^{-1}(\bz; \bx)$ to random latent variables $\bz \sim \pz$.

\begin{figure}[t]
    \centering
    \includegraphics[width=1\linewidth]{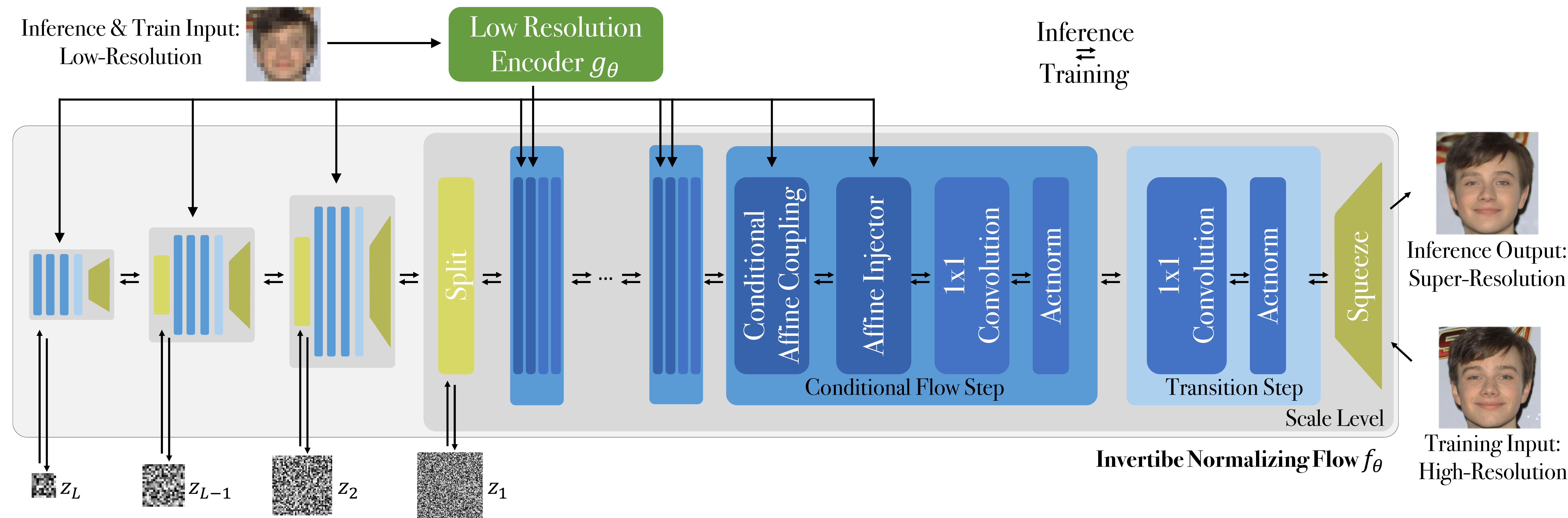}

    \caption{\textbf{SRFlow's conditional normalizing flow architecture.}
    Our model consists of an invertible flow network $\fT$, conditioned on an encoding (green) of the low-resolution image. The flow network operates at multiple scale levels (gray). The input is processed through a series of flow-steps (blue), each consisting of four different layers.
    Through exact log-likelihood training, our network learns to transform a Gaussian density $\pz(\bz)$ to the conditional HR-image distribution $\pyGivenx(\by|\bx,\bT)$. During training, an LR-HR $(\bx,\by)$ image pair is input in order to compute the negative log-likelihood loss. During inference, the network operates in the reverse direction by inputting the LR image along with a random variables $\bz = (\bz_l)_{l=1}^L \sim \pz$, which generates sample SR images from the learned distribution $\pyGivenx$.
    }

    \label{fig:method_architecture}
\end{figure}

In order to achieve a tractable expression of the second term in \eqref{eq:nll}, the neural network $\fT$ is decomposed into a sequence of $N$ invertible layers $\bh^{n+1} = \fT^n(\bh^{n}; \gT(\bx))$, where $\bh^0 = \by$ and $\bh^N = \bz$. We let the LR image to first be encoded by a shared deep CNN $\gT(\bx)$ that extracts a rich representation suitable for conditioning in all flow-layers, as detailed in Sec.~\ref{sec:architecture}. By applying the chain rule along with the multiplicative property of the determinant~\cite{Dinh2017RealNVP}, the NLL objective in \eqref{eq:nll} can be expressed as
\begin{equation}
    \label{eq:nll-layers}
    \mathcal{L}(\bT; \bx, \by) = -\log \pz(\bz) - \sum_{n=0}^{N-1} \log \left| \det \frac{\partial\fT^n}{\partial\bh^{n}}(\bh^{n}; \gT(\bx)) \right| \,.
\end{equation}
We thus only need to compute the log-determinant of the Jacobian $\frac{\partial\fT^n}{\partial\bh^{n}}$ for each individual flow-layer $\fT^n$.
To ensure efficient training and inference, the flow layers $\fT^n$ thus need to allow efficient inversion and a tractable Jacobian determinant. This is further discussed next, where we detail the employed conditional flow layers $\fT^n$ in our SR architecture. Our overall network architecture for flow-based super-resolution is depicted in Fig.~\ref{fig:method_architecture}.

\subsection{Conditional Flow Layers}
\label{sec:layers}

The design of flow-layers $\fT^n$ requires care in order to ensure a well-conditioned inverse and a tractable Jacobian determinant.
This challenge was first addressed in~\cite{DinhKB14NICENonLinIndependentComponetsEst,Dinh2017RealNVP} and has recently spurred significant interest~\cite{BehrmannGCDJ19iResNets,DurkanB0P19NeuralSplineFlows,KingmaD18Glow}.
We start from the unconditional Glow architecture~\cite{KingmaD18Glow}, which is itself based on the RealNVP~\cite{Dinh2017RealNVP}.
The flow layers employed in these architectures can be made conditional in a straight-forward manner~\cite{Ardizzone19cINNcolor,Winkler2020CondNormalizingFlowSR}.
We briefly review them here along with our introduced Affine Injector layer.

\parsection{Conditional Affine Coupling}
The affine coupling layer~\cite{DinhKB14NICENonLinIndependentComponetsEst,Dinh2017RealNVP} provides a simple and powerful strategy for constructing flow-layers that are easily invertible. It is trivially extended to the conditional setting as follows,
\begin{equation}
    \label{eq:couplingMethod}
        \bh^{n+1}_A = \bh^n_A \;,\qquad
        \bh^{n+1}_B = \exp\big(f_{\bT,\text{s}}^n(\bh^n_A; \bu)\big) \cdot \bh^n_B + f_{\bT,\text{b}}^n(\bh^n_A; \bu) \,.
\end{equation}
Here, $\bh^n = (\bh^n_A, \bh^n_B)$ is a partition of the activation map in the channel dimension. Moreover, $\bu$ is the conditioning variable, set to the encoded LR image $\bu = \gT(\bx)$ in our work. Note that $f_{\bT,\text{s}}^n$ and $f_{\bT,\text{b}}^n$ represent \emph{arbitrary} neural networks that generate the scaling and bias of $\bh^n_B$. The Jacobian of \eqref{eq:couplingMethod} is triangular, enabling the efficient computation of its log-determinant as $\sum_{ijk} f_{\bT,\text{s}}^n(\bh^n_A; \bu)_{ijk}$.

\parsection{Invertible $1 \times 1$ Convolution}
General convolutional layers are often intractable to invert or evaluate the determinant of. However, \cite{KingmaD18Glow} demonstrated that a $1 \times 1$ convolution $\bh^{n+1}_{ij} = W \bh^{n}_{ij}$ can be efficiently integrated since it acts on each spatial coordinate $(i,j)$ independently, which leads to a block-diagonal structure. We use the non-factorized formulation in \cite{KingmaD18Glow}.

\parsection{Actnorm}
This provides a channel-wise normalization through a learned scaling and bias.
We keep this layer in its standard un-conditional form~\cite{KingmaD18Glow}.

\parsection{Squeeze}
It is important to process the activations at different scales in order to capture correlations and structures over larger distances.
The squeeze layer~\cite{KingmaD18Glow} provides an invertible means to halving the resolution of the activation map $\bh^n$ by reshaping each spatial $2\times2$ neighborhood into the channel dimension.

\parsection{Affine Injector}
To achieve more direct information transfer from the low-resolution image encoding $\bu = \gT(\bx)$ to the flow branch, we additionally introduce the affine injector layer.
In contrast to the conditional affine coupling layer, our affine injector layer directly affects all channels and spatial locations in the activation map $\bh^n$.
This is achieved by predicting an element-wise scaling and bias using only the conditional encoding $\bu$,
\begin{equation}
    \label{eq:condaffine}
    \bh^{n+1} = \exp\!\big(f_{\bT,\text{s}}^n(\bu)\big) \cdot \bh^n + f_{\bT,\text{b}}(\bu) \,.
\end{equation}
Here, $f_{\bT,\text{s}}$ and $f_{\bT,\text{s}}$ can be any network.
The inverse of~\eqref{eq:condaffine} is trivially obtained as $\bh^n = \exp(-f_{\bT,\text{s}}^n(\bu)) \cdot (\bh^{n+1} - f_{\bT,\text{b}}^n(\bu))$ and the log-determinant is given by $\sum_{ijk} f_{\bT,\text{s}}^n(\bu)_{ijk}$.
Here, the sum ranges over all spatial $i,j$ and channel indices $k$.

\subsection{Architecture}
\label{sec:architecture}

Our SRFlow architecture, depicted in Fig.~\ref{fig:method_architecture}, consists of the invertible flow network $\fT$ and the LR encoder $\gT$.
The flow network is organized into $L$ levels, each operating at a resolution of $\frac{H}{2^l} \times \frac{W}{2^l}$, where $l \in \{1, \ldots, L\}$ is the level number and $H \times W$ is the HR resolution.
Each level itself contains $K$ number of flow-steps.

\parsection{Flow-step}
Each flow-step in our approach consists of four different layers, as visualized in Fig.~\ref{fig:method_architecture}.
The Actnorm if applied first, followed by the $1\times1$ convolution.
We then apply the two conditional layers, first the Affine Injector followed by the Conditional Affine Coupling.

\parsection{Level transitions}
Each level first performs a squeeze operation that effectively halves the spatial resolution.
We observed that this layer can lead to checkerboard artifacts in the reconstructed image, since it is only based on pixel re-ordering.
To learn a better transition between the levels, we therefore remove the conditional layers first few flow steps after the squeeze (see Fig.~\ref{fig:method_architecture}).
This allows the network to learn a linear invertible interpolation between neighboring pixels.
Similar to~\cite{KingmaD18Glow}, we split off $50\%$ of the channels before the next squeeze layer.
Our latent variables $(z_l)_{l=1}^L$ thus model variations in the image at different resolutions, as visualized in Fig.~\ref{fig:method_architecture}.

\parsection{Low-resolution encoding network $\gT$}
SRFlow allows for the use of any differentiable architecture for the LR encoding network $\gT$, since it does not need to be invertible.
Our approach can therefore benefit from the advances in standard feed-forward SR architectures.
In particular, we adopt the popular CNN architecture based on Residual-in-Residual Dense Blocks (RRDB)~\cite{wang2018esrgan}, which builds upon \cite{ledig2017photo,lim2017EDSR}.
It employs multiple residual and dense skip connections, without any batch normalization layers.
We first discard the final upsampling layers in the RRDB architecture since we are only interested in the underlying representation and not the SR prediction.
In order to capture a richer representation of the LR image at multiple levels, we additionally concatenate the activations after each RRDB block to form the final output of $\gT$.

\parsection{Details}
We employ $K=16$ flow-steps at each level, with two additional unconditional flow-steps after each squeeze layer (discussed above).
We use $L=3$ and $L=4$ levels for SR factors $4\times$ and $8\times$ respectively.
For general image SR, we use the standard 23-block RRDB architecture~\cite{wang2018esrgan} for the LR encoder $\gT$.
For faces, we reduce to 8 blocks for efficiency.
The networks $f_{\bT,\text{s}}^n$ and $f_{\bT,\text{b}}^n$ in the conditional affine coupling~\eqref{eq:couplingMethod} and
the affine injector~\eqref{eq:condaffine} are constructed using two shared convolutional layers with ReLU, followed by a final convolution.

\subsection{Training Details}
\label{sec:trainingDetails}
We train our entire SRFlow network using the negative log-likelihood loss~\eqref{eq:nll-layers}.
We sample batches of 16 LR-HR image pairs $(\bx, \by)$.
During training, we use an HR patch size of $160\times 160$.
As optimizer we use Adam with a starting learning rate of $5\cdot10^{-4}$, which is halved at $50\%, 75\%, 90\%$ and $95\%$ of the total training iterations.
To increase training efficiency, we first pre-train the LR encoder $\gT$ using an $L_1$ loss for $200$k iterations.
We then train our full SRFlow architecture using only the loss~\eqref{eq:nll-layers} for $200$k iterations.
Our network takes 5 days to train on a single NVIDIA V100 GPU.
Further details are provided in the appendix.

\parsection{Datasets}
For face super-resolution, we use the CelebA~\cite{liu2015CelebA} dataset.
Similar to~\cite{KingmaD18Glow,Kim19ProgressFSR}, we pre-process the dataset by cropping aligned patches, which are resized to the HR resolution of $160\times 160$.
We employ the full train split ($160$k images).
For general SR, we use the same training data as ESRGAN~\cite{wang2018esrgan}, consisting of the train split of 800 DIV2k~\cite{div2k} along with 2650 images from Flickr2K.
The LR images are constructed using the standard MATLAB bicubic kernel. 

\section{Applications and Image Manipulations}
\label{sec:applications}

\begin{figure}[t]
    \centering%
    \begin{minipage}{0.45\textwidth}
        \includegraphics[width=\textwidth]{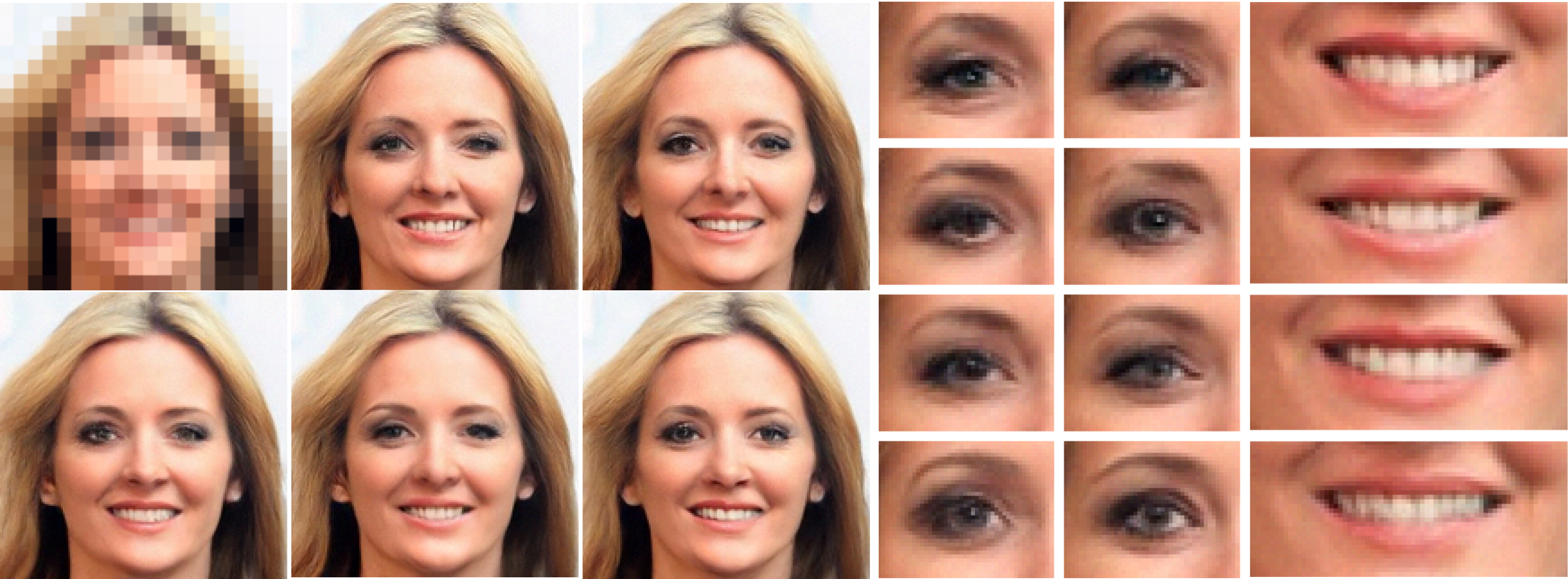}%
        \caption{Random $8\times$ SR samples generated by SRFlow using a temperature $\tau=0.8$. LR image is shown in top left.}
        \label{fig:srsamples}
    \end{minipage}~~~%
    \begin{minipage}{0.55\textwidth}
        \includegraphics[width=\linewidth]{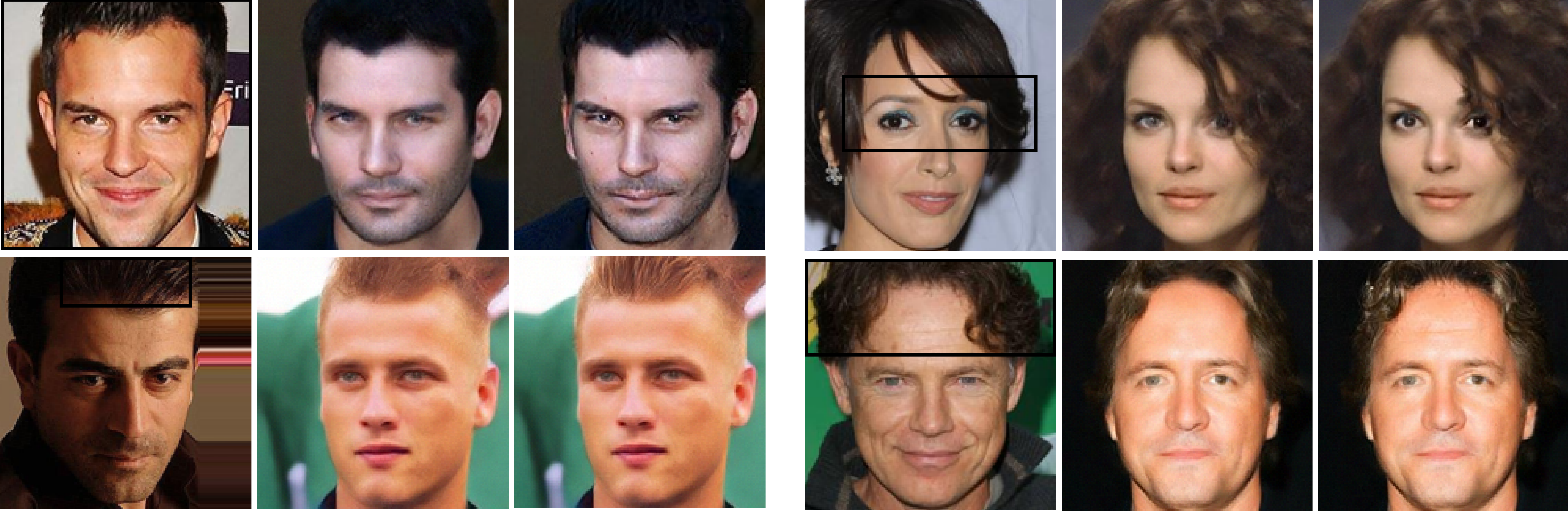}
        \resizebox{\linewidth}{!}{
        \begin{tabular}{ C{2cm} C{2cm} C{2cm}  C{0.5cm} C{2cm} C{2cm} C{2cm}}
            Source & Target & Transferred & & Source & Target & Transferred
        \end{tabular}
        }
        \caption{Latent space transfer from the region marked by the box to the target image. (8$\times$)}
        \label{fig:stylesamples}
    \end{minipage}%
\end{figure}

In this section, we explore the use of our SRFlow network for a variety of applications and image manipulation tasks.
Our techniques exploit two key advantages of our SRFlow network, which are not present in GAN-based super-resolution approaches~\cite{wang2018esrgan}.
First, our network models a distribution $\pyGivenx(\by | \bx, \bT)$ in HR-image space, instead of only predicting a single image.
It therefore possesses great flexibility by capturing a variety of possible HR predictions.
This allows different predictions to be explored using additional guiding information or random sampling.
Second, the flow network $\fT(\by; \bx)$ is a fully invertible encoder-decoder.
Hence, \emph{any} HR image $\tilde{\by}$ can be encoded into the latent space as $\tilde{\bz} = \fT(\tilde{\by}; \bx)$
and \emph{exactly} reconstructed as $\tilde{\by} = \fT^{-1}(\tilde{\bz}; \bx)$.
This bijective correspondence allows us to flexibly operate in both the latent and image space.

\subsection{Stochastic Super-resolution}

The distribution $\pyGivenx(\by | \bx, \bT)$ learned by our SRFlow can be explored by sampling different SR predictions as $\by^{(i)} = \fT^{-1}(\bz^{(i)}; \bx),\, \bz^{(i)}\! \sim \pz$ for a given LR image $\bx$.
As commonly observed for flow-based models, the best results are achieved when sampling with a slightly lower variance~\cite{KingmaD18Glow}. We therefore use a Gaussian $\bz^{(i)}\! \sim \mathcal{N}(0, \tau)$ with variance $\tau$ (also called temperature).
Results are visualized in Fig.~\ref{fig:srsamples} for $\tau=0.8$. Our approach generates a large variety of SR images, including differences in e.g.\ hair and facial attributes, while preserving consistency with the LR image.
Since our latent variables $\bz_{ijkl}$ are spatially localized, specific parts can be re-sampled, enabling more controlled interactive editing and exploration of the SR image.

\subsection{LR-Consistent Style Transfer}
Our SRFlow allows transferring the style of an existing HR image $\tilde{\by}$ when super-resolving an LR image $\bx$.
This is performed by first encoding the source HR image as $\tilde{\bz} = \fT(\tilde{\by}; \downsc(\tilde{\by}))$, where $\downsc$ is the down-scaling operator.
The encoding $\tilde{\bz}$ can then be used to as the latent variable for the super-resolution of $\bx$ as $\by = \fT^{-1}(\tilde{\bz}; \bx)$.
This operation can also be performed on local regions of the image.
Examples in Fig.~\ref{fig:stylesamples} show the transfer in the style of facial characteristics, hair and eye color.
Our SRFlow network automatically aims to ensure consistency with the LR image without any additional constraints.

\subsection{Latent Space Normalization}
\label{sec:latentnorm}

We develop more advanced image manipulation techniques by taking advantage of the invertability of the SRFlow network $\fT$ and the learned super-resolution posterior. The core idea of our approach is to map any HR image containing desired content to the latent space, where the latent statistics can be normalized in order to make it consistent with the low-frequency information in the given LR image. Let $\bx$ be a low-resolution image and $\tilde{\by}$ be \emph{any} high-resolution image, not necessarily consistent with the LR image $\bx$. For example, $\tilde{\by}$ can be an edited version of a super-resolved image or a guiding image for the super-resolution image. Our goal is to achieve an HR image $\by$, containing image content from $\tilde{\by}$, but that is consistent with the LR image $\bx$.

The latent encoding for the given image pair is computed as $\tilde{\bz} = \fT(\tilde{\by}; \bx)$. Note that our network is trained to predict consistent and natural SR images for latent variables sampled from a standard Gaussian distribution $\pz = \mathcal{N}(0, I)$. Since $\tilde{\by}$ is not necessarily consistent with the LR image $\bx$, the latent variables $\tilde{\bz}_{ijkl}$ do not have the same statistics as if independently sampled from $\bz_{ijkl} \sim \mathcal{N}(0, \tau)$. Here, $\tau$ denotes an additional temperature scaling of the desired latent distribution. In order to achieve desired statistics, we normalize the first two moments of collections of latent variables. In particular, if $\{z_i\}_1^N \sim \mathcal{N}(0, \tau)$ are independent, then it is well known \cite{MurphyML} that their empirical mean $\hat{\mu}$ and variance $\hat{\sigma}^2$ are distributed according to,
\begin{equation}
\label{eq:moment-distr}
\hat{\mu} = \frac{1}{N} \sum_{i=1}^N z_i \sim \mathcal{N}\left(0, \frac{\tau}{N}\right) , \;\, \hat{\sigma}^2 = \frac{1}{N\!-\!1} \sum_{i=1}^N (z_i - \hat{\mu})^2 \sim \Gamma\left(\frac{N\!-\!1}{2}, \frac{2 \tau}{N\!-\!1}\right) .
\end{equation}
Here, $\Gamma(k,\theta)$ is a gamma distribution with shape and scale parameters $k$ and $\theta$ respectively. For a given collection $\tilde{\mathcal{Z}} \subset \{\bz_{ijkl}\}$ of latent variables, we normalize their statistics by first sampling a new mean $\hat{\mu}$ and variance $\hat{\sigma}^2$ according to \eqref{eq:moment-distr}, where $N = |\tilde{\mathcal{Z}}|$ is the size of the collection. The latent variables in the collection are then normalized as,
\begin{equation}
\label{eq:normalization}
\hat{z} = \frac{\hat{\sigma}}{\tilde{\sigma}} (\tilde{z} - \tilde{\mu}) + \hat{\mu} \,,\quad \forall\tilde{z} \in \tilde{\mathcal{Z}} \,.
\end{equation}
Here, $\tilde{\mu}$ and $\tilde{\sigma}^2$ denote the empirical mean and variance of the collection $\tilde{\mathcal{Z}}$.

The normalization in \eqref{eq:normalization} can be performed using different collections $\tilde{\mathcal{Z}}$. We consider three different strategies in this work. \textbf{Global normalization} is performed over the entire latent space, using $\tilde{\mathcal{Z}} = \{\bz_{ijkl}\}_{ijkl}$. For \textbf{local normalization}, each spatial position $i,j$ in each level $l$ is normalized independently as $\tilde{\mathcal{Z}}_{ijl} = \{\bz_{ijkl}\}_{k}$. This better addresses cases where the statistics is spatially varying. \textbf{Spatial normalization} is performed independently for each feature channel $k$ and level $l$, using $\tilde{\mathcal{Z}}_{kl} = \{\bz_{ijkl}\}_{ij}$. It addresses global effects in the image that activates certain channels, such as color shift or noise.
In all three cases, normalized latent variable $\hat{\bz}$ is obtained by applying \eqref{eq:normalization} for all collections, which is an easily parallelized computation. The final HR image is then reconstructed as $\hat{\by} = \fTinv(\hat{\bz},\bx)$.
Note that our normalization procedure is stochastic, since a new mean $\hat{\mu}$ and variance $\hat{\sigma}^2$ are sampled independently for every collection of latent variables $\tilde{\mathcal{Z}}$. This allows us to sample from the natural diversity of predictions $\hat{\by}$, that integrate content from $\tilde{\by}$. Next, we explore our latent space normalization technique for different applications.

\begin{figure}[t]
    \centering%
    \newcommand{\imggt}[1]{%
    \newcommand{\size}{0.25}%
    \includegraphics[width=\size\linewidth]{figures/rgbGtTrans/#1_source}%
    \includegraphics[width=\size\linewidth]{figures/rgbGtTrans/#1_target}%
    \includegraphics[width=\size\linewidth]{figures/rgbGtTrans/#1_rgb}%
    \includegraphics[width=\size\linewidth]{figures/rgbGtTrans/#1_mod}%
    }%
    \newcommand{\imgsr}[1]{%
    \includegraphics[width=\size\linewidth]{figures/rgbSrTrans/#1_source}%
    \includegraphics[width=\size\linewidth]{figures/rgbSrTrans/#1_sr_orig}%
    \includegraphics[width=\size\linewidth]{figures/rgbSrTrans/#1_rgb}%
    \includegraphics[width=\size\linewidth]{figures/rgbSrTrans/#1_sr}%
    }%
    \begin{minipage}{0.60\textwidth}
        \imggt{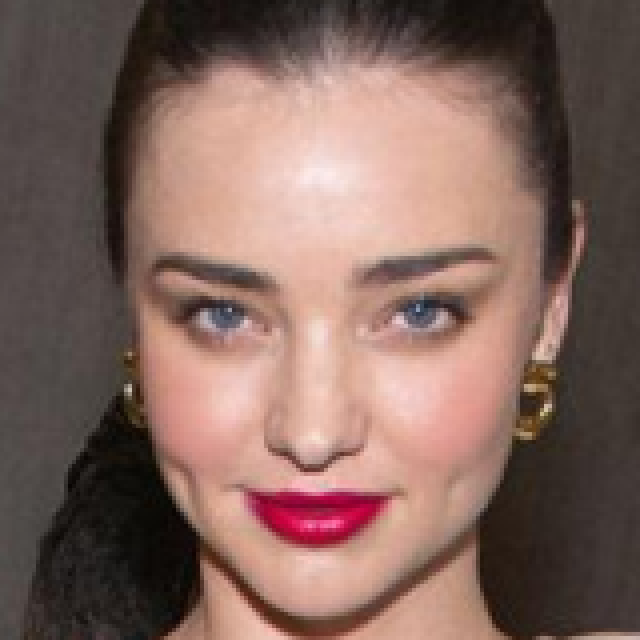}
        \imgsr{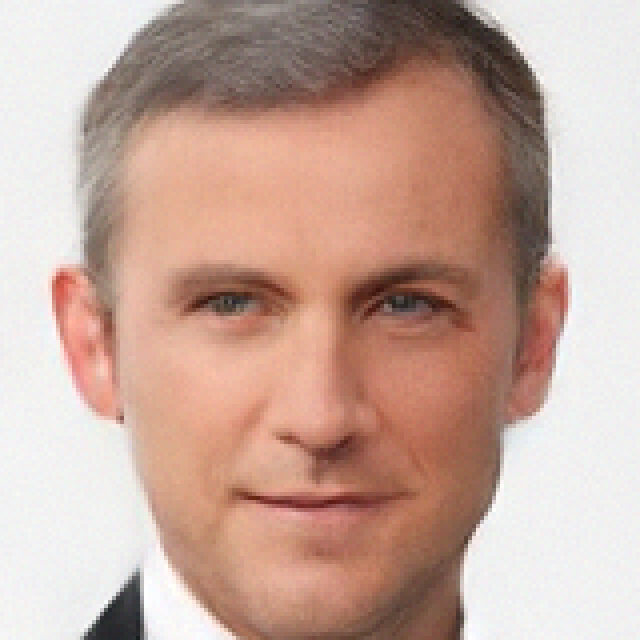}
        \resizebox{0.98\linewidth}{!}{
        \hspace{-2mm}\begin{tabular}{@{}C{2cm} C{2cm} C{2cm} C{2cm}}
                         Source & Target $\by$ & Input $\tilde{\by}$ & Transferred $\hat{\by}$
        \end{tabular}}
        \caption{Image content transfer for an existing HR image (top) and an SR prediction (bottom). Content from the source is applied directly to the target. By applying latent space normalization in our SRFlow, the content is integrated and harmonized.}
        \label{fig:content-transfer}
    \end{minipage}~~~%
    \begin{minipage}{0.37\textwidth}
        \resizebox{\textwidth}{!}{%
        \begin{tabular}{c@{~~~}l@{~~~}c@{~~~}c@{~~~}c}
\toprule
&& Original & Super-Resloved & Restored \\
\midrule
\parbox[t]{2mm}{\multirow{3}{*}{\rotatebox[origin=c]{90}{DIV2K}}}
&PSNR$\uparrow$    & 22.48 & 23.19 & 27.81
\\
&SSIM$\uparrow$    &  0.49 &  0.51 & 0.73  \\
&LPIPS$\downarrow$ & 0.370 & 0.364 & 0.255 \\
\midrule
\parbox[t]{2mm}{\multirow{3}{*}{\rotatebox[origin=c]{90}{CelebA}}}
&PSNR$\uparrow$    & 22.52 & 24.25 & 27.62 \\
&SSIM$\uparrow$    &  0.48 &  0.63 &  0.78 \\
&LPIPS$\downarrow$ & 0.326 & 0.172 & 0.143 \\
\bottomrule
\end{tabular}

        }
        \includegraphics[width=\linewidth]{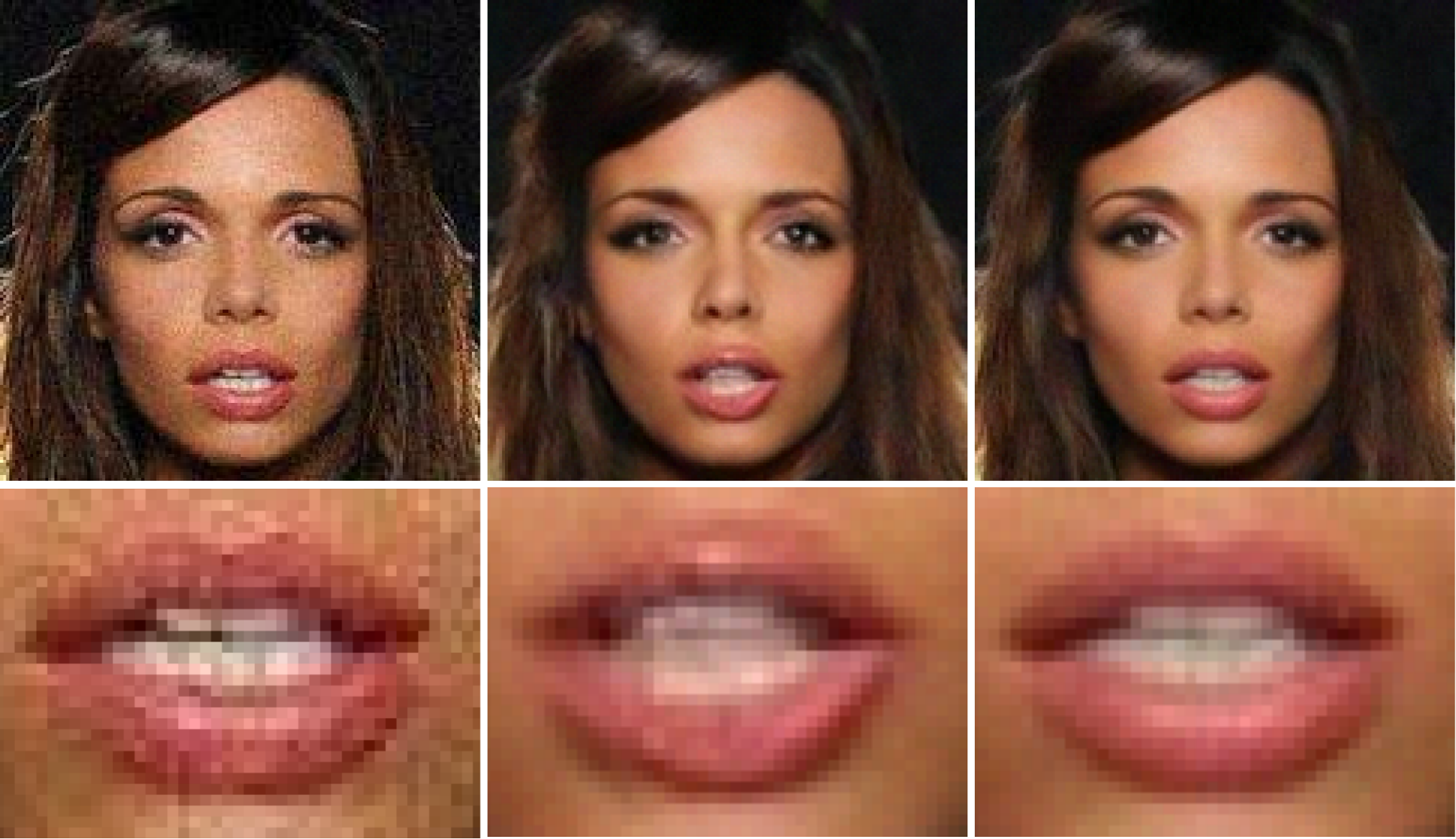}
        \resizebox{\linewidth}{!}{
        \begin{tabular}{ C{2cm} C{2cm} C{2cm} }
            Original & Direct SR & Restored
        \end{tabular}}
        \caption{Comparision of super-resolving the LR of the original and normalizing the latent space for image restoration.}
        \label{fig:restoration}
    \end{minipage}%
\end{figure}

\subsection{Image Content Transfer}

Here, we aim to manipulate an HR image by transferring content from other images.
Let $\bx$ be an LR image and $\by$ a corresponding HR image.
If we are manipulating a super-resolved image, then $\by = \fTinv(\bz, \bx)$ is an SR sample of $\bx$.
However, we can also manipulate an existing HR image $\by$ by setting $\bx = \downsc(\by)$ to the down-scaled version of $\by$.
We then manipulate $\by$ directly in the image space by simply inserting content from other images, as visualized in Fig.~\ref{fig:content-transfer}.
To harmonize the resulting manipulated image $\tilde{\by}$ by ensuring consistency with the LR image $\bx$,
we compute the latent encoding $\tilde{\bz} = \fT(\tilde{\by}; \bx)$ and perform \emph{local normalization} of the latent variables as described in Sec.~\ref{sec:latentnorm}.
We only normalize the affected regions of the image in order to preserve the non-manipulated content. 
Results are shown in Fig.~\ref{fig:content-transfer}.
If desired, the emphasis on LR-consistency can be reduced by training SRFlow with randomly misaligned HR-LR pairs, which allows increased manipulation flexibility (see Appendix).

\subsection{Image Restoration}

We demonstrate the strength of our learned image posterior by applying it for image restoration tasks.
Note that we here employ the \emph{same} SRFlow network, that is trained only for super-resolution, and not for the explored tasks.
In particular, we investigate degradations that mainly affect the high frequencies in the image, such as noise and compression artifacts.
Let $\tilde{\by}$ be a degraded image.
Noise and other high-frequency degradations are largely removed when down-sampled $\bx = \downsc(\tilde{\by})$.
Thus a cleaner image can be obtained by applying any super-resolution method to $\bx$.
However, this generates poor results since important image information is lost in the down-sampling process (Fig.~\ref{fig:restoration}, center).

Our approach can go beyond this result by directly utilizing the original image $\tilde{\by}$.
The degraded image along with its down-sampled variant are input to our SRFlow network to generate the latent variable $\tilde{\bz} = \fT(\tilde{\by}; \bx)$.
We then perform first \emph{spatial} and then \emph{local} normalization of $\tilde{\bz}$, as described in Sec.~\ref{sec:latentnorm}.
The restored image is then predicted as $\hat{\by} = \fTinv(\hat{\bz},\bx)$.
By, denoting the normalization operation as $\hat{\bz} = \phi(\tilde{\bz})$, the full restoration mapping can be expressed as $\hat{\by} = \fTinv(\phi(\fT(\tilde{\by}; \downsc(\tilde{\by}))),\downsc(\tilde{\by}))$.
As shown visually and quantitatively in Fig.~\ref{fig:restoration}, this allows us to recover a substantial amount of details from the original image
Intuitively, our approach works by mapping the degraded image $\tilde{\by}$ to the \emph{closest} image within the learned distribution $\pyGivenx(\by | \bx, \bT)$.
Since SRFlow is not trained with such degradations, $\pyGivenx(\by | \bx, \bT)$ mainly models \emph{clean} images.
Our normalization therefore automatically restores the image when it is transformed to a more \emph{likely} image according to our SR distribution $\pyGivenx(\by | \bx, \bT)$.

\section{Experiments}

We perform comprehensive experiments for super-resolution of faces and of generic images in comparisons with current state-of-the-art and an ablative analysis.
Applications, such as image manipulation tasks, are presented in Sec.~\ref{sec:applications}, with additional results, analysis and visuals in the appendix.

\parsection{Evaluation Metrics}
To evaluate the perceptual distance to the Ground Truth, we report the default LPIPS~\cite{zhang2018unreasonable}.
It is a learned distance metric, based on the feature-space of a finetuned AlexNet.
We report the standard fidelity oriented metrics, Peak Signal to Noise Ratio (PSNR) and structural similarity index (SSIM)~\cite{WangBSS04SSIM}, although they are known to not correlate well with the human perception of image quality~\cite{ignatov2018pirm,ledig2017photo,RealWorldSR,AIM2019RWSRchallenge,timofte2017ntire,timofte2018ntire}.
Furthermore, we report the no-reference metrics NIQE~\cite{MittalSB13NIQE}, BRISQUE~\cite{mittal2011brisque} and PIQUE~\cite{NDBCM15piqe}.
In addition to visual quality, consistency with the LR image is an important factor.
We therefore evaluate this aspect by reporting the LR-PSNR, computed as the PSNR between the downsampled SR image and the original LR image.

\begin{table}[t]
    \centering%
    \caption{Results for $8\times$ SR of faces of CelebA.
    We compare using both the standard bicubic kernel and the progressive linear kernel from~\cite{Kim19ProgressFSR}.
    We also report the diversity in the SR output in terms of the pixel standard deviation $\sigma$.}%
    \resizebox{\textwidth}{!}{%
    \begin{tabular}{c@{~}l@{~~~}c@{~~}c@{~~}c@{~}|@{~}c@{~}|@{~}c@{~~}c@{~~}c@{~}|@{~}c}
\toprule
LR& & $\uparrow$PSNR & $\uparrow$SSIM & $\downarrow$LPIPS & $\uparrow$LR-PSNR & $\downarrow$NIQE & $\downarrow$BRISQUE & $\downarrow$PIQUE &
$\uparrow$Diversity $\sigma$ \\ \midrule
\parbox[t]{2mm}{\multirow{4}{*}{\rotatebox[origin=c]{90}{Bicubic}}}
& Bicubic                            & 23.15 & 0.63  & 0.517 & 35.19 & 7.82 & 58.6 & 99.97 & 0 \\
& RRDB~\cite{wang2018esrgan}         & 26.59 & 0.77  & 0.230 & 48.22 & 6.02 & 49.7 & 86.5  & 0 \\
& ESRGAN~\cite{wang2018esrgan}       & 22.88 & 0.63  & 0.120 & 34.04 & 3.46 & 23.7 & 32.4  & 0 \\
& \textbf{SRFlow} \small{$\tau=0.8$}    & 25.24 & 0.71  & 0.110 & 50.85 & 4.20 & 23.2 & 24.0  & 5.21 \\ \midrule
\parbox[t]{2mm}{\multirow{2}{*}{\rotatebox[origin=c]{90}{Prog.}}}
& ProgFSR~\cite{Kim19ProgressFSR}    & 23.97 & 0.67  & 0.129 & 41.95 & 3.49 & 28.6 & 33.2  & 0 \\
& \textbf{SRFlow} \small{$\tau=0.8$}    & 25.20 & 0.71  & 0.110 & 51.05 & 4.20 & 22.5 & 23.1  & 5.28 \\
\bottomrule
\end{tabular}
    }
    \label{tab:sotaCMbic8}
\end{table}

% Experiments figure
\begin{figure}[t]
    \centering%
    \newcommand{\size}{0.145}%
    \newcommand{\img}[1]{%
    \includegraphics[width=\size\linewidth]{figures/sotaCMbic8/#1/lq}~~~~%
    \includegraphics[width=\size\linewidth]{figures/sotaCMbic8/#1/RRDBn23.jpg}~~%
    \includegraphics[width=\size\linewidth]{figures/sotaCMbic8/#1/ESRGANn23px2e-1.jpg}~~%
    \includegraphics[width=\size\linewidth]{figures/sotaCMbic8/#1/Prog.jpg}~~%
    \includegraphics[width=\size\linewidth]{figures/sotaCMbic8/#1/ChristmasP.jpg}~~%
    \includegraphics[width=\size\linewidth]{figures/sotaCMbic8/#1/gt.jpg}%
    }%
    \img{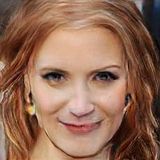}
    \img{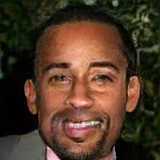}
    \img{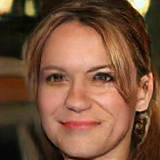}
    \img{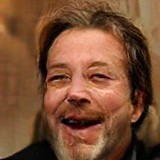}
    \resizebox{1.00\linewidth}{!}{%
    \begin{tabular}{ C{3.05cm} C{2.5cm} C{2.5cm} C{2.5cm} C{2.5cm} C{2.5cm} }
        LR  & RRDB~\cite{wang2018esrgan}~~ & ESRGAN~\cite{wang2018esrgan} & ProgFSR~\cite{Kim19ProgressFSR} & \textbf{SRFlow} \small{$\tau=0.8$} & Ground Truth
    \end{tabular}}%
    \caption{Comparison of our SRFlow with state-of-the-art for $8\times$ face SR on CelebA.}
    \label{fig:sotaCMbic8}
\end{figure}

\begin{table}[t]
    \centering%
    \caption{General image SR results on the 100 validation images of the DIV2K dataset.}
    \resizebox{\textwidth}{!}{%
    \begin{tabular}{l@{~~~}c@{~~}c@{~~}c@{~~}c@{~~}c@{~~}c@{~~}c@{~~}|@{~~}c@{~~}c@{~~}c@{~~}c@{~~}c@{~~}c@{~~}c}
\toprule
& \multicolumn{7}{c}{\textbf{DIV2K} $4\times$} & \multicolumn{7}{c}{\textbf{DIV2K} $8\times$} \\
& PSNR$\uparrow$ & SSIM$\uparrow$ & LPIPS$\downarrow$ & LR-PSNR$\uparrow$ & NIQE$\downarrow$ & BRISQUE$\downarrow$ & PIQUE$\downarrow$ & PSNR$\uparrow$ & SSIM$\uparrow$ & LPIPS$\downarrow$ & LR-PSNR$\uparrow$ & NIQE$\downarrow$ & BRISQUE$\downarrow$ & PIQUE$\downarrow$ \\
\midrule
Bicubic                             &  26.70 &  0.77 &  0.409 & 38.70 & 5.20 & 53.8 & 86.6 &  23.74 & 0.63 &  0.584 & 37.16 & 6.65 & 60.3 & 97.6 \\
EDSR~\cite{lim2017EDSR}             &  28.98 &  0.83 &  0.270 & 54.89 & 4.46 & 43.3 & 77.5 &     -  &    - &      - &    -  &    - &    - &    - \\
RRDB~\cite{wang2018esrgan}          &  29.44 &  0.84 &  0.253 & 49.20 & 5.08 & 52.4 & 86.7 &  25.50 & 0.70 &  0.419 & 45.43 & 4.35 & 42.4 & 79.1 \\
RankSRGAN~\cite{zhang2019ranksrgan}  &  26.55 &  0.75 &  0.128 & 42.33 & 2.45 & 17.2 & 20.1 &     -  &  -   &      - &     - &    - &    - &    - \\
ESRGAN~\cite{wang2018esrgan}        &  26.22 &  0.75 &  0.124 & 39.03 & 2.61 & 22.7 & 26.2 &  22.18 & 0.58 &  0.277 & 31.35 & 2.52 & 20.6 & 25.8 \\
\textbf{SRFlow}  \small{$\tau=0.9$} &  27.09 &  0.76 &  0.120 & 49.96 & 3.57 & 17.8 & 18.6 &  23.05 & 0.57 &  0.272 & 50.00 & 3.49 & 20.9 & 17.1 \\
\bottomrule
\end{tabular}

    }
    \vspace{4mm}
    \label{tab:sotadiv2k}
\end{table}
\begin{figure}[t]
\centering%
\newcommand{\size}{0.117}%
\newcommand{\img}[1]{%
\includegraphics[width=\size\linewidth]{figures/sotaDiv2k4/images/#1_lq}
\includegraphics[width=\size\linewidth]{figures/sotaDiv2k4/images/#1_Bicubic_Df2k4.jpg}
\includegraphics[width=\size\linewidth]{figures/sotaDiv2k4/images/#1_EDSR.jpg}
\includegraphics[width=\size\linewidth]{figures/sotaDiv2k4/images/#1_RRDBn23.jpg}
\includegraphics[width=\size\linewidth]{figures/sotaDiv2k4/images/#1_ESRGAN.jpg}
\includegraphics[width=\size\linewidth]{figures/sotaDiv2k4/images/#1_RankSRGAN.jpg}
\includegraphics[width=\size\linewidth]{figures/sotaDiv2k4/images/#1_Christmas.jpg}
\includegraphics[width=\size\linewidth]{figures/sotaDiv2k4/images/#1_gt.jpg}%
}
\img{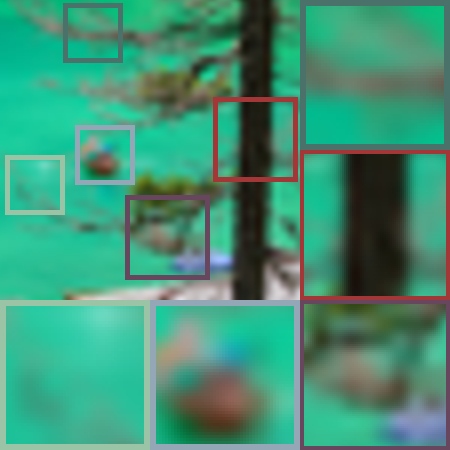}
\img{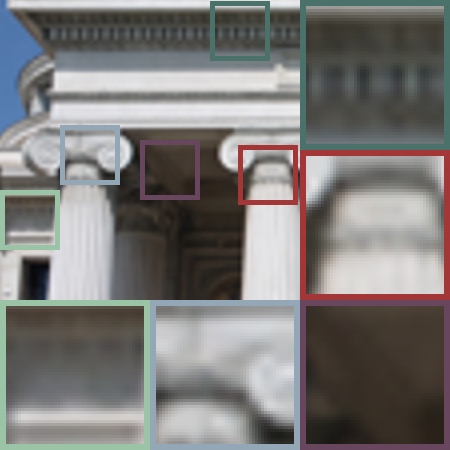}
\img{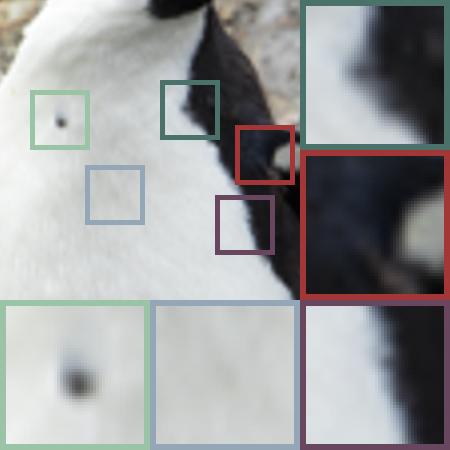}
\img{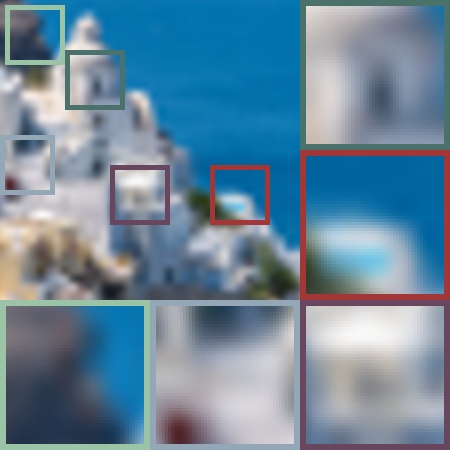}
\resizebox{\linewidth}{!}{
    \begin{tabular}{ C{2.5cm} C{2.5cm} C{2.5cm} C{2.5cm} C{2.5cm} C{2.5cm} C{2.5cm} C{2.5cm} }
        Low Resolution & Bicubic & EDSR \cite{lim2017EDSR} & RRDB \cite{wang2018esrgan} & ESRGAN \cite{wang2018esrgan} & RankSRGAN~\cite{zhang2019ranksrgan} & \textbf{SRFlow} \small{$\tau=0.9$} & Ground Truth
\end{tabular}
}
\caption{Comparison to state-of-the-art for general SR on the DIV2K validation set.}
\label{fig:sotaDiv2k4}
\end{figure}

\subsection{Face Super-Resolution}
We evaluate SRFlow for face SR ($8\times$) using 5000 images from the test split of the CelebA dataset.
We compare with bicubic, RRDB~\cite{wang2018esrgan}, ESRGAN~\cite{wang2018esrgan}, and ProgFSR~\cite{Kim19ProgressFSR}.
While the latter two are GAN-based, RRDB is trained using only $L_1$ loss.
ProgFSR is a very recent SR method specifically designed for faces, shown to outperform several prior face SR approaches in~\cite{Kim19ProgressFSR}. It is trained on the full train split of CelebA, but using a bilinear kernel.
For fair comparison, we therefore separately train and evaluate SRFlow on the same kernel.

Results are provided in Tab.~\ref{tab:sotaCMbic8} and Fig.~\ref{fig:sotaCMbic8}.
Since our aim is perceptual quality, we consider LPIPS the primary metric, as it has been shown to correlate much better with human opinions~\cite{ntire2020,zhang2018unreasonable}.
SRFlow achieves more than twice as good LPIPS distance compared to RRDB, at the cost of lower PSNR and SSIM scores.
As seen in the visual comparisons in Fig.~\ref{fig:sotaCMbic8}, RRDB generates extremely blurry results, lacking natural high-frequency details.
Compared to the GAN-based methods, SRFlow achieves significantly better results in all reference metrics.
Interestingly, even the PSNR is superior to ESRGAN and ProgFSR, showing that our approach preserves fidelity to the HR ground-truth, while achieving better perceptual quality.
This is partially explained by the hallucination artifacts that often plague GAN-based approaches, as seen in Fig.~\ref{fig:sotaCMbic8}.
Our approach generate sharp and natural images, while avoiding such artifacts.
Interestingly, our SRFlow achieves an LR-consistency that is even better than the fidelity-trained RRDB, while the GAN-based methods are comparatively in-consistent with the input LR image.

\subsection{General Super-Resolution}
Next, we evaluate our SRFlow for general SR on the DIV2K validation set.
We compare SRFlow to bicubic, EDSR~\cite{lim2017EDSR}, RRDB~\cite{wang2018esrgan}, ESRGAN~\cite{wang2018esrgan}, and RankSRGAN~\cite{zhang2019ranksrgan}.
Except for EDSR, which used DIV2K, all methods including SRFlow are trained on the train splits of DIV2K and Flickr2K (see Sec.~\ref{sec:architecture}).
For the $4\times$ setting, we employ the provided pre-trained models.
Due to lacking availability, we trained RRDB and ESRGAN for $8\times$ using the authors' code.

EDSR and RRDB are trained using only reconstruction losses, thereby achieving inferior results in terms of the perceptual LPIPS metric (Tab.~\ref{tab:sotadiv2k}).
Compared to the GAN-based methods \cite{wang2018esrgan,zhang2019ranksrgan}, our SRFlow achieves significantly better PSNR, LPIPS and LR-PSNR and favorable results in terms of PIQUE and BRISQUE.
Visualizations in Fig.~\ref{fig:sotaDiv2k4} confirm the perceptually inferior results of EDSR and RRDB, which generate little high-frequency details.
In contrast, SRFlow generates rich details, achieving favorable perceptual quality compared to ESRGAN.
The first row, ESRGAN generates severe discolored artifacts and ringing patterns at several locations in the image.
We find SRFlow to generate more stable and consistent results in these circumstances.

\begin{figure}[t]
    \centering%
    \newcommand{\sizeCMbic}{0.18}
    \newcommand{\imgCMbic}[1]{\includegraphics[width=\sizeCMbic\linewidth]{figures/ablationCMbic8/#1}}
    \newcommand{\imgHCMbic}[1]{\includegraphics[width=\sizeCMbic\linewidth]{figures/ablationH196/#1}}
    \begin{minipage}{0.55\textwidth}
        \imgCMbic{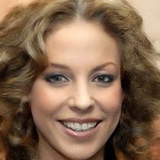}
        \imgCMbic{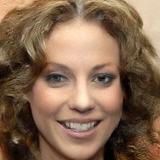}
        \imgCMbic{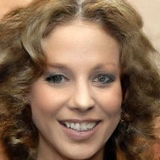}~~
        \imgHCMbic{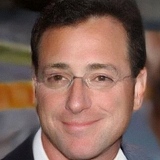}
        \imgHCMbic{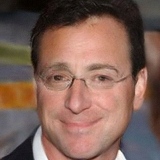} \\
        \imgCMbic{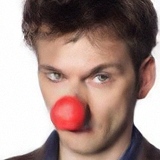}
        \imgCMbic{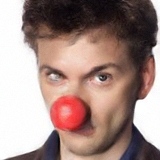}
        \imgCMbic{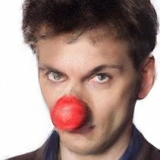}~~
        \imgHCMbic{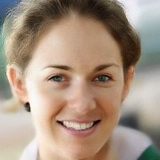}
        \imgHCMbic{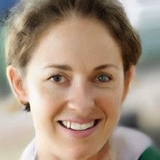} \\
        \resizebox{\textwidth}{!}{%
        \begin{tabular}{ C{2.3cm} C{2.3cm} C{2.3cm} C{0.3cm} C{2.3cm} C{2.3cm}}
            $K=16$ Steps & $K=8$ Steps & $K=4$ Steps & & 196 Channels & 64 Channels
        \end{tabular}}
        \caption{Analysis of number of flow steps and dimensionality in the conditional layers.}
        \label{fig:ablationCMbic8}
    \end{minipage}\hspace{0.02\textwidth}%
    \begin{minipage}{0.43\textwidth}
        \resizebox{\textwidth}{!}{%
        \begin{tabular}{l@{~~~}c@{~~}c@{~~}c@{~~}}
\toprule
DIV2K 4$\times$             & PSNR$\uparrow$ & SSIM$\uparrow$ & LPIPS$\downarrow$ \\
\midrule
No Lin. F-Step              & 26.96  & 0.759  & 0.125 \\
No Affine Inj.              & 26.81  & 0.756  & 0.126 \\
SRFlow                      & 27.09  & 0.763  & 0.125 \\
\bottomrule
\end{tabular}

        }
        \captionof{table}{Analysis of the impact of the transitional linear flow steps and the affine image injector.}
        \label{tab:ablation}
    \end{minipage}
\end{figure}

\subsection{Ablative Study}

To ablate the depth and width, we train our network with different number of flow-steps $K$ and hidden layers in two conditional layers \eqref{eq:coupling} and \eqref{eq:condaffine} respectively.
Figure~\ref{fig:ablationCMbic8} shows results on the CelebA dataset.
Decreasing the number of flow-steps $K$ leads to more artifacts in complex structures, such as eyes.
Similarly, a larger number of channels leads to better consistency in the reconstruction.
In Tab.~\ref{tab:ablation} we analyze architectural choices.
The Affine Image Injector increases the fidelity, while preserving the perceptual quality. We also observe the transitional linear flow steps (Sec.~\ref{sec:architecture}) to be beneficial.

\section{Conclusion}

We propose a flow-based method for super-resolution, called SRFlow. Contrary to conventional methods, our approach learns the distribution of photo-realistic SR images given the input LR image. This explicitly accounts for the ill-posed nature of the SR problem and allows for the generation of diverse SR samples.
Moreover, we develop techniques for image manipulation, exploiting the strong image posterior learned by SRFlow. In comprehensive experiments, our approach achieves improved results compared to state-of-the-art GAN-based approaches.

\parsection{Acknowledgements}
This work was supported by the ETH Z\"urich Fund (OK), a Huawei Technologies Oy (Finland) project, a Google GCP grant, an Amazon AWS grant, and an Nvidia GPU grant.

\bibliographystyle{splncs04} \bibliography{references}

\begin{thebibliography}{10}
\providecommand{\url}[1]{\texttt{#1}}
\providecommand{\urlprefix}{URL }
\providecommand{\doi}[1]{https://doi.org/#1}

\bibitem{div2k}
Agustsson, E., Timofte, R.: Ntire 2017 challenge on single image
  super-resolution: Dataset and study. In: CVPR Workshops (2017)

\bibitem{ahn2018image}
Ahn, N., Kang, B., Sohn, K.A.: Image super-resolution via progressive cascading
  residual network. In: CVPR (2018)

\bibitem{Ardizzone19cINNcolor}
Ardizzone, L., L{\"{u}}th, C., Kruse, J., Rother, C., K{\"{o}}the, U.: Guided
  image generation with conditional invertible neural networks. CoRR
  \textbf{abs/1907.02392} (2019), \url{http://arxiv.org/abs/1907.02392}

\bibitem{bahat2019explorableSR}
Bahat, Y., Michaeli, T.: Explorable super resolution. In: CVPR (2020)

\bibitem{BehrmannGCDJ19iResNets}
Behrmann, J., Grathwohl, W., Chen, R.T.Q., Duvenaud, D., Jacobsen, J.:
  Invertible residual networks. In: {ICML}. Proceedings of Machine Learning
  Research, vol.~97, pp. 573--582. {PMLR} (2019)

\bibitem{Bell19InternalGAN}
Bell{-}Kligler, S., Shocher, A., Irani, M.: Blind super-resolution kernel
  estimation using an internal-gan. In: NeurIPS. pp. 284--293 (2019),
  \url{http://papers.nips.cc/paper/8321-blind-super-resolution-kernel-estimation-using-an-internal-gan}

\bibitem{BlauM18PerceptionDistTradeoff}
Blau, Y., Michaeli, T.: The perception-distortion tradeoff. In: CVPR. pp.
  6228--6237 (2018). \doi{10.1109/CVPR.2018.00652},
  \url{http://openaccess.thecvf.com/content\_cvpr\_2018/html/Blau\_The\_Perception-Distortion\_Tradeoff\_CVPR\_2018\_paper.html}

\bibitem{buhler2020deepsee}
B{\"u}hler, M.C., Romero, A., Timofte, R.: Deepsee: Deep disentangled semantic
  explorative extreme super-resolution. arXiv preprint arXiv:2004.04433  (2020)

\bibitem{DaiTG15JointlyOptimizedRegressorsForSR}
Dai, D., Timofte, R., Gool, L.V.: Jointly optimized regressors for image
  super-resolution. Comput. Graph. Forum  \textbf{34}(2),  95--104 (2015).
  \doi{10.1111/cgf.12544}

\bibitem{DinhKB14NICENonLinIndependentComponetsEst}
Dinh, L., Krueger, D., Bengio, Y.: {NICE:} non-linear independent components
  estimation. In: 3rd International Conference on Learning Representations,
  {ICLR} 2015, San Diego, CA, USA, May 7-9, 2015, Workshop Track Proceedings
  (2015)

\bibitem{Dinh2017RealNVP}
Dinh, L., Sohl{-}Dickstein, J., Bengio, S.: Density estimation using real
  {NVP}. In: 5th International Conference on Learning Representations, {ICLR}
  2017, Toulon, France, April 24-26, 2017, Conference Track Proceedings (2017)

\bibitem{DongLHT14LearningDeepConv}
Dong, C., Loy, C.C., He, K., Tang, X.: Learning a deep convolutional network
  for image super-resolution. In: ECCV. pp. 184--199 (2014).
  \doi{10.1007/978-3-319-10593-2\_13}

\bibitem{dong2016image}
Dong, C., Loy, C.C., He, K., Tang, X.: Image super-resolution using deep
  convolutional networks. TPAMI  \textbf{38}(2),  295--307 (2016)

\bibitem{DurkanB0P19NeuralSplineFlows}
Durkan, C., Bekasov, A., Murray, I., Papamakarios, G.: Neural spline flows. In:
  Advances in Neural Information Processing Systems 32: Annual Conference on
  Neural Information Processing Systems 2019, NeurIPS 2019, 8-14 December 2019,
  Vancouver, BC, Canada. pp. 7509--7520 (2019)

\bibitem{Goodfellow14GAN}
Goodfellow, I.J., Pouget{-}Abadie, J., Mirza, M., Xu, B., Warde{-}Farley, D.,
  Ozair, S., Courville, A.C., Bengio, Y.: Generative adversarial nets. In:
  Advances in Neural Information Processing Systems 27: Annual Conference on
  Neural Information Processing Systems 2014, December 8-13 2014, Montreal,
  Quebec, Canada. pp. 2672--2680 (2014)

\bibitem{haris2018deep}
Haris, M., Shakhnarovich, G., Ukita, N.: Deep back-projection networks for
  super-resolution. In: CVPR (2018)

\bibitem{ignatov2018pirm}
Ignatov, A., Timofte, R., Van~Vu, T., Luu, T.M., Pham, T.X., Van~Nguyen, C.,
  Kim, Y., Choi, J.S., Kim, M., Huang, J., et~al.: Pirm challenge on perceptual
  image enhancement on smartphones: Report. arXiv preprint arXiv:1810.01641
  (2018)

\bibitem{IsolaZZE17pix2pix}
Isola, P., Zhu, J., Zhou, T., Efros, A.A.: Image-to-image translation with
  conditional adversarial networks. In: CVPR. pp. 5967--5976 (2017).
  \doi{10.1109/CVPR.2017.632}, \url{https://doi.org/10.1109/CVPR.2017.632}

\bibitem{Kim19ProgressFSR}
Kim, D., Kim, M., Kwon, G., Kim, D.: Progressive face super-resolution via
  attention to facial landmark. In: arxiv. vol. abs/1908.08239 (2019)

\bibitem{kim2016accurate}
Kim, J., Kwon~Lee, J., Mu~Lee, K.: Accurate image super-resolution using very
  deep convolutional networks. In: CVPR (2016)

\bibitem{KingmaD18Glow}
Kingma, D.P., Dhariwal, P.: Glow: Generative flow with invertible 1x1
  convolutions. In: Advances in Neural Information Processing Systems 31:
  Annual Conference on Neural Information Processing Systems 2018, NeurIPS
  2018, 3-8 December 2018, Montr{\'{e}}al, Canada. pp. 10236--10245 (2018)

\bibitem{lai2017deep}
Lai, W.S., Huang, J.B., Ahuja, N., Yang, M.H.: Deep laplacian pyramid networks
  for fast and accurate super-resolution. In: CVPR (2017)

\bibitem{ledig2017photo}
Ledig, C., Theis, L., Husz{\'a}r, F., Caballero, J., Cunningham, A., Acosta,
  A., Aitken, A.P., Tejani, A., Totz, J., Wang, Z., et~al.: Photo-realistic
  single image super-resolution using a generative adversarial network. CVPR
  (2017)

\bibitem{lim2017EDSR}
Lim, B., Son, S., Kim, H., Nah, S., Lee, K.M.: Enhanced deep residual networks
  for single image super-resolution. CVPR  (2017)

\bibitem{LiuLGWL19CondAdversarialGlow}
Liu, R., Liu, Y., Gong, X., Wang, X., Li, H.: Conditional adversarial
  generative flow for controllable image synthesis. In: {IEEE} Conference on
  Computer Vision and Pattern Recognition, {CVPR} 2019, Long Beach, CA, USA,
  June 16-20, 2019. pp. 7992--8001 (2019)

\bibitem{liu2015CelebA}
Liu, Z., Luo, P., Wang, X., Tang, X.: Deep learning face attributes in the
  wild. In: Proceedings of International Conference on Computer Vision (ICCV)
  (December 2015)

\bibitem{RealWorldSR}
Lugmayr, A., Danelljan, M., Timofte, R.: Unsupervised learning for real-world
  super-resolution. In: ICCVW. pp. 3408--3416. IEEE (2019)

\bibitem{ntire2020}
Lugmayr, A., Danelljan, M., Timofte, R.: Ntire 2020 challenge on real-world
  image super-resolution: Methods and results. In: Proceedings of the IEEE/CVF
  Conference on Computer Vision and Pattern Recognition (CVPR) Workshops (June
  2020)

\bibitem{AIM2019RWSRchallenge}
Lugmayr, A., Danelljan, M., Timofte, R., et~al.: Aim 2019 challenge on
  real-world image super-resolution: Methods and results. In: ICCV Workshops
  (2019)

\bibitem{MathieuCL15VideoMSE}
Mathieu, M., Couprie, C., LeCun, Y.: Deep multi-scale video prediction beyond
  mean square error. In: ICLR (2016), \url{http://arxiv.org/abs/1511.05440}

\bibitem{menon2020pulse}
Menon, S., Damian, A., Hu, S., Ravi, N., Rudin, C.: Pulse: Self-supervised
  photo upsampling via latent space exploration of generative models. In: CVPR
  (2020)

\bibitem{mittal2011brisque}
Mittal, A., Moorthy, A., Bovik, A.: Referenceless image spatial quality
  evaluation engine. In: 45th Asilomar Conference on Signals, Systems and
  Computers. vol.~38, pp. 53--54 (2011)

\bibitem{MittalSB13NIQE}
Mittal, A., Soundararajan, R., Bovik, A.C.: Making a "completely blind" image
  quality analyzer. {IEEE} Signal Process. Lett.  \textbf{20}(3),  209--212
  (2013)

\bibitem{MurphyML}
Murphy, K.P.: Machine Learning: A Probabilistic Perspective. The MIT Press
  (2012)

\bibitem{NDBCM15piqe}
N., V., D., P., Bh., M.C., Channappayya, S.S., Medasani, S.S.: Blind image
  quality evaluation using perception based features. In: {NCC}. pp.~1--6.
  {IEEE} (2015)

\bibitem{PathakKDDE16i2iInpainting}
Pathak, D., Kr{\"{a}}henb{\"{u}}hl, P., Donahue, J., Darrell, T., Efros, A.A.:
  Context encoders: Feature learning by inpainting. In: {CVPR}. pp. 2536--2544.
  {IEEE} Computer Society (2016)

\bibitem{pumarola2020cflow}
Pumarola, A., Popov, S., Moreno-Noguer, F., Ferrari, V.: C-flow: Conditional
  generative flow models for images and 3d point clouds. In: CVPR. pp.
  7949--7958 (2020)

\bibitem{RezendeM15VarInferenceWithNF}
Rezende, D.J., Mohamed, S.: Variational inference with normalizing flows. In:
  Proceedings of the 32nd International Conference on Machine Learning, {ICML}
  2015, Lille, France, 6-11 July 2015. pp. 1530--1538 (2015)

\bibitem{Sajjadi17EnhanceNet}
Sajjadi, M.S.M., Sch{\"{o}}lkopf, B., Hirsch, M.: Enhancenet: Single image
  super-resolution through automated texture synthesis. In: {IEEE}
  International Conference on Computer Vision, {ICCV} 2017, Venice, Italy,
  October 22-29, 2017. pp. 4501--4510. {IEEE} Computer Society (2017).
  \doi{10.1109/ICCV.2017.481}

\bibitem{shaham2019singan}
Shaham, T.R., Dekel, T., Michaeli, T.: Singan: Learning a generative model from
  a single natural image. In: ICCV. pp. 4570--4580 (2019)

\bibitem{shocher2018zssr}
Shocher, A., Cohen, N., Irani, M.: Zero-shot” super-resolution using deep
  internal learning. In: CVPR (2018)

\bibitem{SunH12SRFromInternetScaleSceneMatching}
Sun, L., Hays, J.: Super-resolution from internet-scale scene matching. In:
  ICCP (2012)

\bibitem{timofte2017ntire}
Timofte, R., Agustsson, E., Van~Gool, L., Yang, M.H., Zhang, L., Lim, B., Son,
  S., Kim, H., Nah, S., Lee, K.M., et~al.: Ntire 2017 challenge on single image
  super-resolution: Methods and results. CVPR Workshops  (2017)

\bibitem{Timofte2014a+}
Timofte, R., De~Smet, V., Van~Gool, L.: A+: Adjusted anchored neighborhood
  regression for fast super-resolution. In: ACCV. pp. 111--126. Springer (2014)

\bibitem{timofte2018ntire}
Timofte, R., Gu, S., Wu, J., Van~Gool, L.: Ntire 2018 challenge on single image
  super-resolution: methods and results. In: CVPR Workshops (2018)

\bibitem{Timofte13AnchNeighReg}
Timofte, R., Smet, V.D., Gool, L.V.: Anchored neighborhood regression for fast
  example-based super-resolution. In: ICCV. pp. 1920--1927 (2013).
  \doi{10.1109/ICCV.2013.241}, \url{https://doi.org/10.1109/ICCV.2013.241}

\bibitem{wang2018esrgan}
Wang, X., Yu, K., Wu, S., Gu, J., Liu, Y., Dong, C., Loy, C.C., Qiao, Y., Tang,
  X.: Esrgan: Enhanced super-resolution generative adversarial networks. ECCV
  (2018)

\bibitem{WangBSS04SSIM}
Wang, Z., Bovik, A.C., Sheikh, H.R., Simoncelli, E.P.: Image quality
  assessment: from error visibility to structural similarity. {IEEE} Trans.
  Image Processing  \textbf{13}(4),  600--612 (2004)

\bibitem{Winkler2020CondNormalizingFlowSR}
Winkler, C., Worrall, D.E., Hoogeboom, E., Welling, M.: Learning likelihoods
  with conditional normalizing flows. arxiv  \textbf{abs/1912.00042} (2019),
  \url{http://arxiv.org/abs/1912.00042}

\bibitem{YangY13SimpleFuncSR}
Yang, C., Yang, M.: Fast direct super-resolution by simple functions. In: ICCV.
  pp. 561--568 (2013). \doi{10.1109/ICCV.2013.75},
  \url{https://doi.org/10.1109/ICCV.2013.75}

\bibitem{Yang2019PointFlow}
Yang, G., Huang, X., Hao, Z., Liu, M., Belongie, S.J., Hariharan, B.:
  Pointflow: 3d point cloud generation with continuous normalizing flows. ICCV
  (2019)

\bibitem{YangWHM08SRAsSparseRepresentationOfRawPatches}
Yang, J., Wright, J., Huang, T.S., Ma, Y.: Image super-resolution as sparse
  representation of raw image patches. In: CVPR (2008).
  \doi{10.1109/CVPR.2008.4587647}

\bibitem{YangWHM10SRViaSparseRep}
Yang, J., Wright, J., Huang, T.S., Ma, Y.: Image super-resolution via sparse
  representation. {IEEE} Trans. Image Processing  \textbf{19}(11),  2861--2873
  (2010). \doi{10.1109/TIP.2010.2050625},
  \url{https://doi.org/10.1109/TIP.2010.2050625}

\bibitem{URDGN}
Yu, X., Porikli, F.: Ultra-resolving face images by discriminative generative
  networks. In: ECCV. pp. 318--333 (2016). \doi{10.1007/978-3-319-46454-1\_20}

\bibitem{zhang2018unreasonable}
Zhang, R., Isola, P., Efros, A.A., Shechtman, E., Wang, O.: The unreasonable
  effectiveness of deep features as a perceptual metric. CVPR  (2018)

\bibitem{zhang2019ranksrgan}
Zhang, W., Liu, Y., Dong, C., Qiao, Y.: Ranksrgan: Generative adversarial
  networks with ranker for image super-resolution (2019)

\end{thebibliography}

\clearpage

% Appendix
\setcounter{section}{0}
\renewcommand{\thesection}{\Alph{section}}

\section{Architecture Details}
\label{sec:architecture-details}

In this section, we give additional details about our SRFlow architecture. 
The construction of a flow-based architecture requires the flow layers to be invertible and have a tractable Jacobian log-determinant. Since super-resolution of diverse images has to be able to cope with different input sizes, we also ensure that our  architecture is fully convolutional. We can therefore train our network on smaller patches, and directly apply it to the full image during testing.
The computational time of our approach is $1.13$ seconds for super-resolving one $256 \times 256$ input LR image with a scale factor of $4\times$ on an Nvidia V100 GPU.

\subsection{Low-resolution Image Encoding}
Our SRFlow network is conditioned on the encoding of the low-resolution image $\bu = \gT(\bx)$. To this end, we employ the RRDB-based architecture, described in the paper. It employs several RRDB-blocks with a channel dimension of 64, operating in the resolution of the input LR image. The final conditioning output $\bu = \gT(\bx)$ is achieved by concatenating the activations from 5 equally spaced RRDB blocks, resulting in a dimensionality of 320. 

\subsection{The Affine Injector Layer}
Our affine injector layer provide a direct means of conditioning all dimensions of the flow feature-map $\bh^n$ on the LR encoding as,
\begin{equation}
    \bh^{n+1} = \exp\!\big(f_{\bT,\text{s}}^n(\bu)\big) \cdot \bh^n + f_{\bT,\text{b}}(\bu) \,.
\end{equation}
The  scale and bias are extracted using non-invertible networks $f_{\bT,\text{s}}(\bu)$ and $f_{\bT,\text{b}}(\bu)$ respectively. The input $\bu$ is first bilinearly resized to the resolution of the corresponding flow-level. A conv-ReLU block first reduces the dimensionality to 64. Another conv-ReLU block is then applied with 64-dimensional output. The output of $f_{\bT,\text{s}}(\bu)$ and $f_{\bT,\text{b}}(\bu)$ are then achieved by two separate conv-layers applied to the same 64-dimensional input. For these layers, we employ the zero-initialization strategy proposed in~\cite{KingmaD18Glow}. All convolutions have a $3\times3$ kernel.

\subsection{Conditional Affine Coupling}
This building block allows applying complex unconstrained conditional learned functions that act on the normalizing flow, without harming its invertibility. This is made possible by bypassing half of the activations and applying an affine transformation to the other half~\cite{DinhKB14NICENonLinIndependentComponetsEst}. This transformation depends on the bypassed half $\bh^n_A$ and conditional features $\bu$ as,
\begin{equation}
    \label{eq:coupling}
    \begin{cases}
    \bh^{n+1}_A = \bh^n_A \\
    \bh^{n+1}_B = \exp\big(f_{\bT,\text{s}}^n(\bh^n_A; \bu)\big) \cdot \bh^n_B + f_{\bT,\text{b}}^n(\bh^n_A; \bu)
    \end{cases} \,.
\end{equation}

This expression can be easily inverted~\cite{DinhKB14NICENonLinIndependentComponetsEst}.
The network architectures of $f_{\bT,\text{s}}$ and $f_{\bT,\text{b}}$ are similar to those of the Affine Injector, described above. The only difference is that the two inputs $\bh^{n}_A$ and $\bu$ are initially concatenated after $\bu$ is resized to the resolution of $\bh^{n}_A$.

\begin{figure}[b]
\centering
\includegraphics[width=0.8\linewidth]{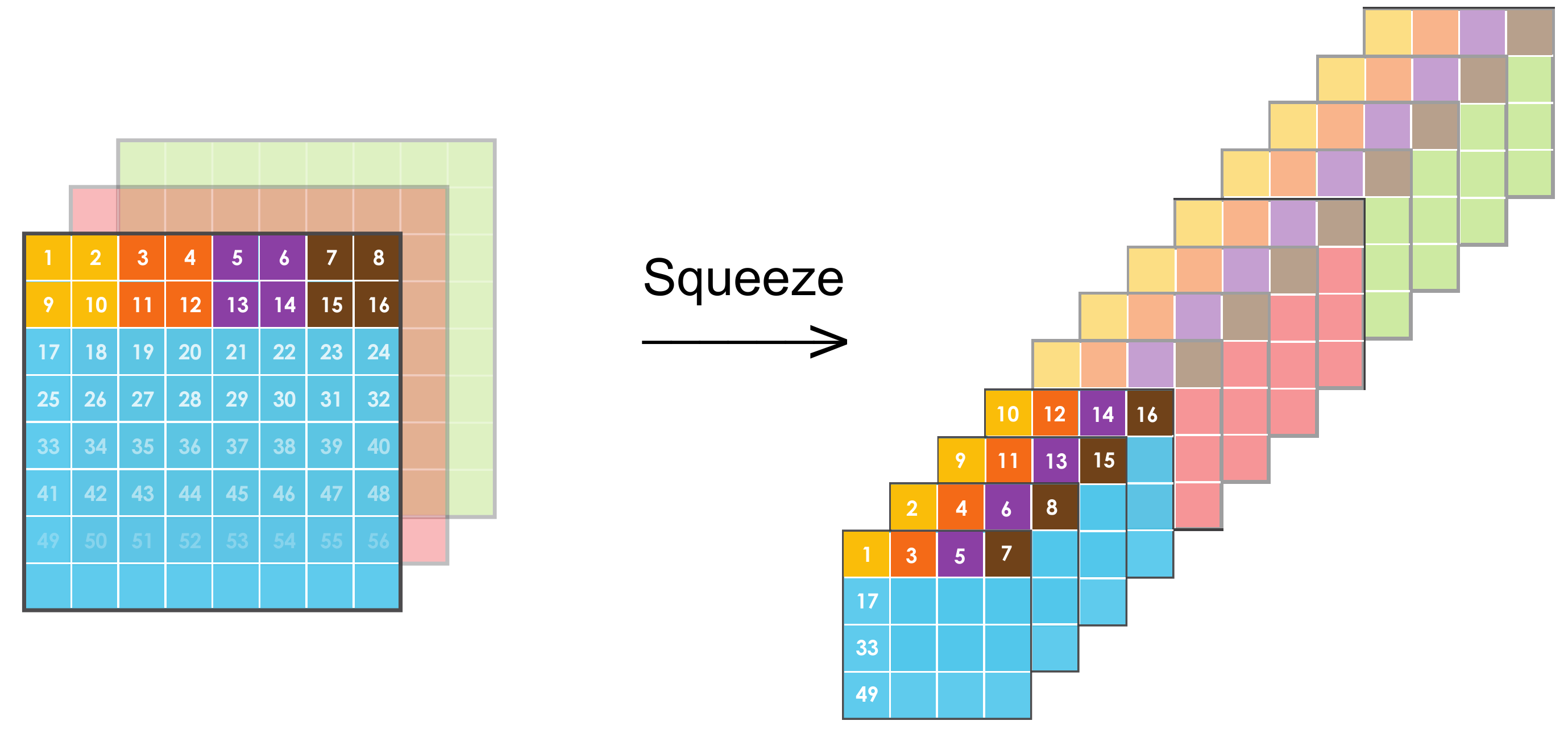}

\vspace{0mm}
\caption{Visualization of the Squeeze Operation.}
\vspace{0mm}
\label{fig:squeeze_layer}
\end{figure}

\subsection{Squeeze Operation}
This layer reshapes the activation map to half the width and height. In order to preserve the locality, neighboring pixels are stacked as seen in Figure~\ref{fig:squeeze_layer}.

\subsection{Activation Norm}
The Activation Norm (Actnorm) is a normalization layer. Unlike Batchnorm, it does not require synchronization among the elements of a batch. It simply consists of a learned scaling and bias factor for each dimension of the feature map. Thus it helps distributed learning on multiple GPUs.

\section{Training Details}
\label{sec:training-details}

In this section, we give additional details about the training procedure for our SRFlow. 
We employ the Adam optimizer with a starting learning rate of $5\cdot10^{-4}$. This learning rate is halved at $50\%, 75\%, 90\%$ and $95\%$ of the total number of training iterations.
During the first $50\%$ of the training iterations, the pre-trained weights of the LR encoder $\gT$ are frozen in a warm-up phase. In the latter $50\%$, all parameters of the SRFlow network, including $\gT$, are optimized jointly with the same learning rate.

As has been observed in e.g.~\cite{KingmaD18Glow}, adding slight random noise to the target image helps the training process and leads to better visual results. We therefore add Gaussian noise with a standard deviation of $\sigma=\frac{4}{\sqrt{3}}$ to the high-resolution image. In contrast to~\cite{KingmaD18Glow}, we do not employ 5-bit quantization.

\section{Detailed Quantitative Analysis}
\label{sec:quant_analysis}

In this section, we provide additional quantitative analysis of our approach.

\subsection{Influence of the Sampling Temperature}

\begin{figure}[t]
\centering%
\newcommand{\size}{0.19}%
\newcommand{\img}[1]{%
\includegraphics[width=\size\linewidth]{figures/heat_examples/celebA/000/#1}
\includegraphics[width=\size\linewidth]{figures/heat_examples/celebA/030/#1}
\includegraphics[width=\size\linewidth]{figures/heat_examples/celebA/060/#1}
\includegraphics[width=\size\linewidth]{figures/heat_examples/celebA/090/#1}
\includegraphics[width=\size\linewidth]{figures/heat_examples/celebA/gt/#1}
}
\img{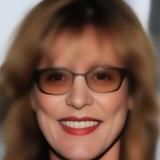}
\img{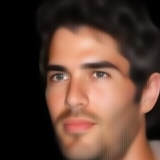}
\vspace{0mm}
\resizebox{\linewidth}{!}{
    \begin{tabular}{ C{2.5cm} C{2.5cm} C{2.5cm} C{2.5cm} C{2.5cm} }
    $\tau = 0$ & $\tau = 0.3$ & $\tau = 0.6$ & $\tau = 0.9$ & Ground-Truth \\
\end{tabular}
}%
\vspace{-3mm}
\caption{Super-resolved images sampled with different temperatures $\tau$.}
\vspace{0mm}
\label{fig:heat_examples}
\end{figure}

\begin{figure}[t]
\centering
\newcommand{\size}{0.45}
\includegraphics[width=\size\linewidth,valign=t]{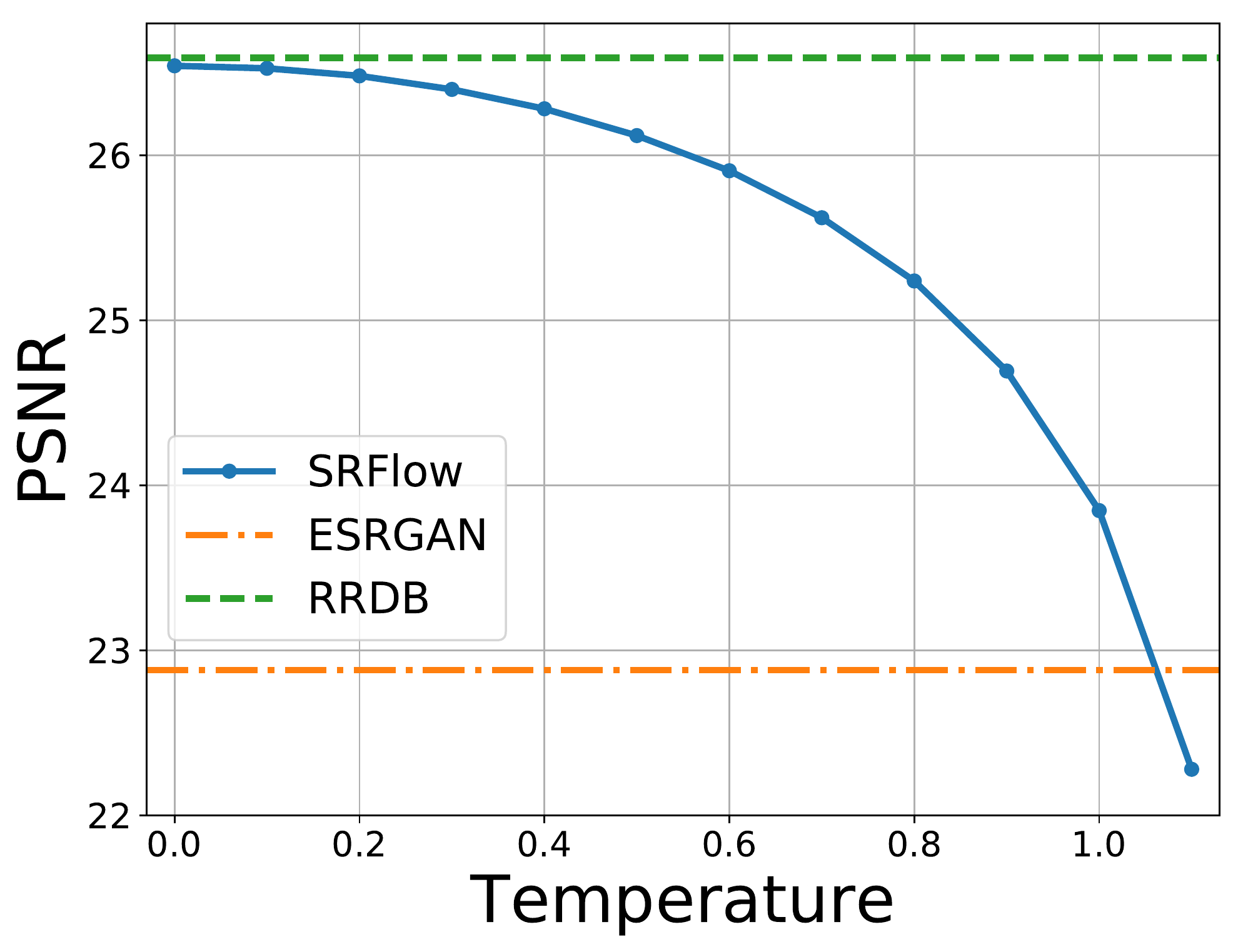}
\includegraphics[width=\size\linewidth,valign=t]{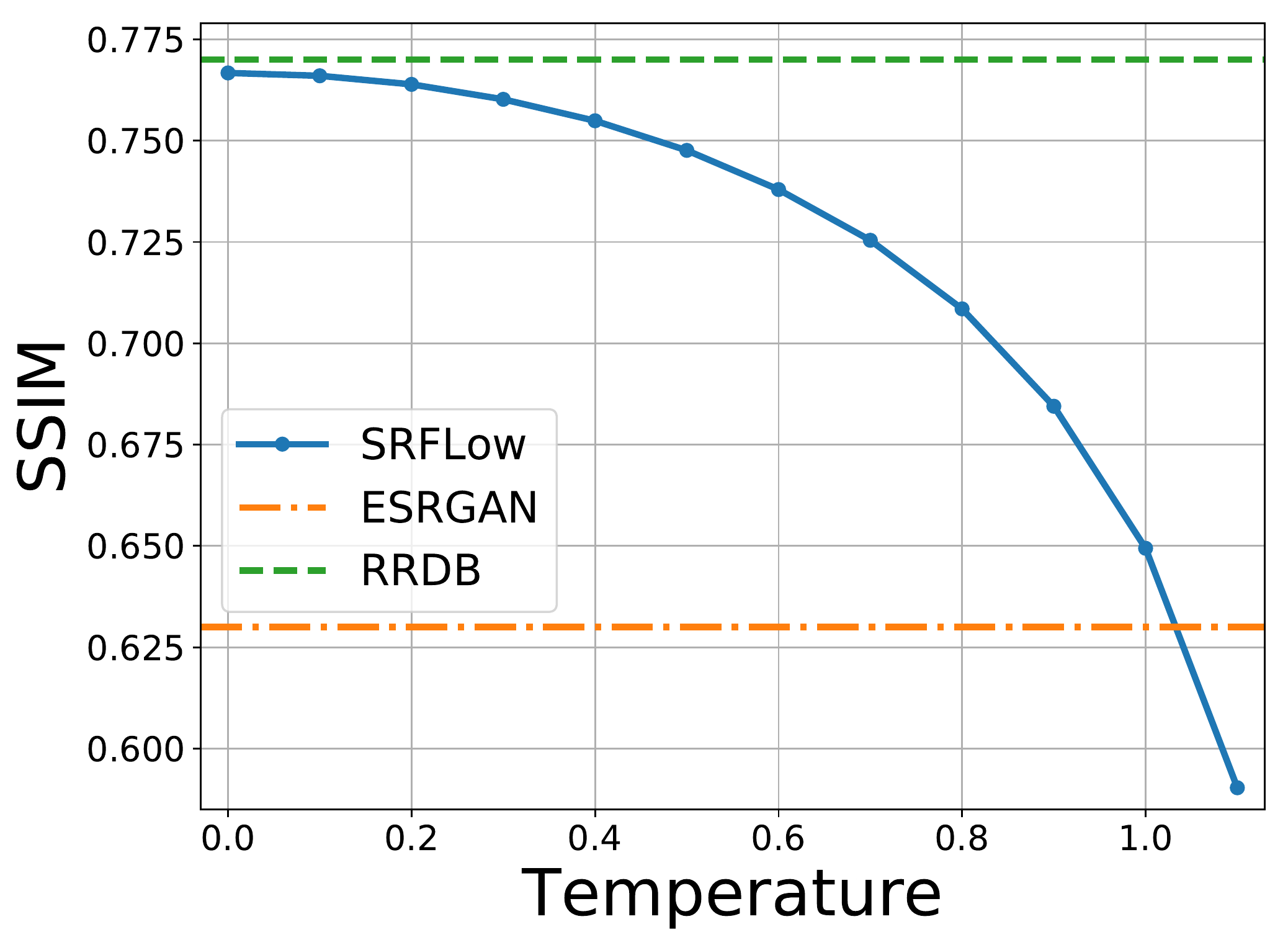}
\includegraphics[width=\size\linewidth,valign=t]{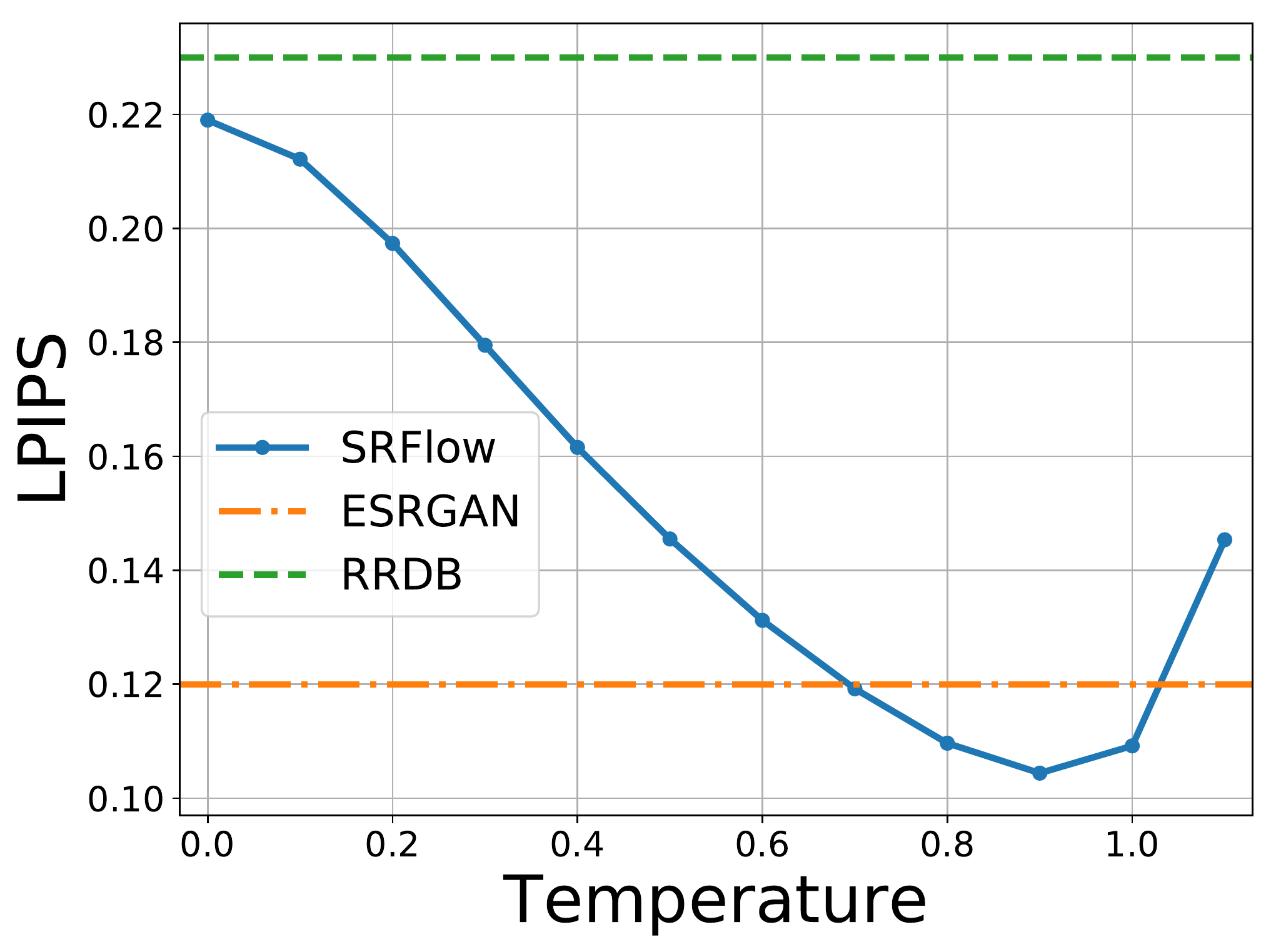}
\includegraphics[width=\size\linewidth,valign=t]{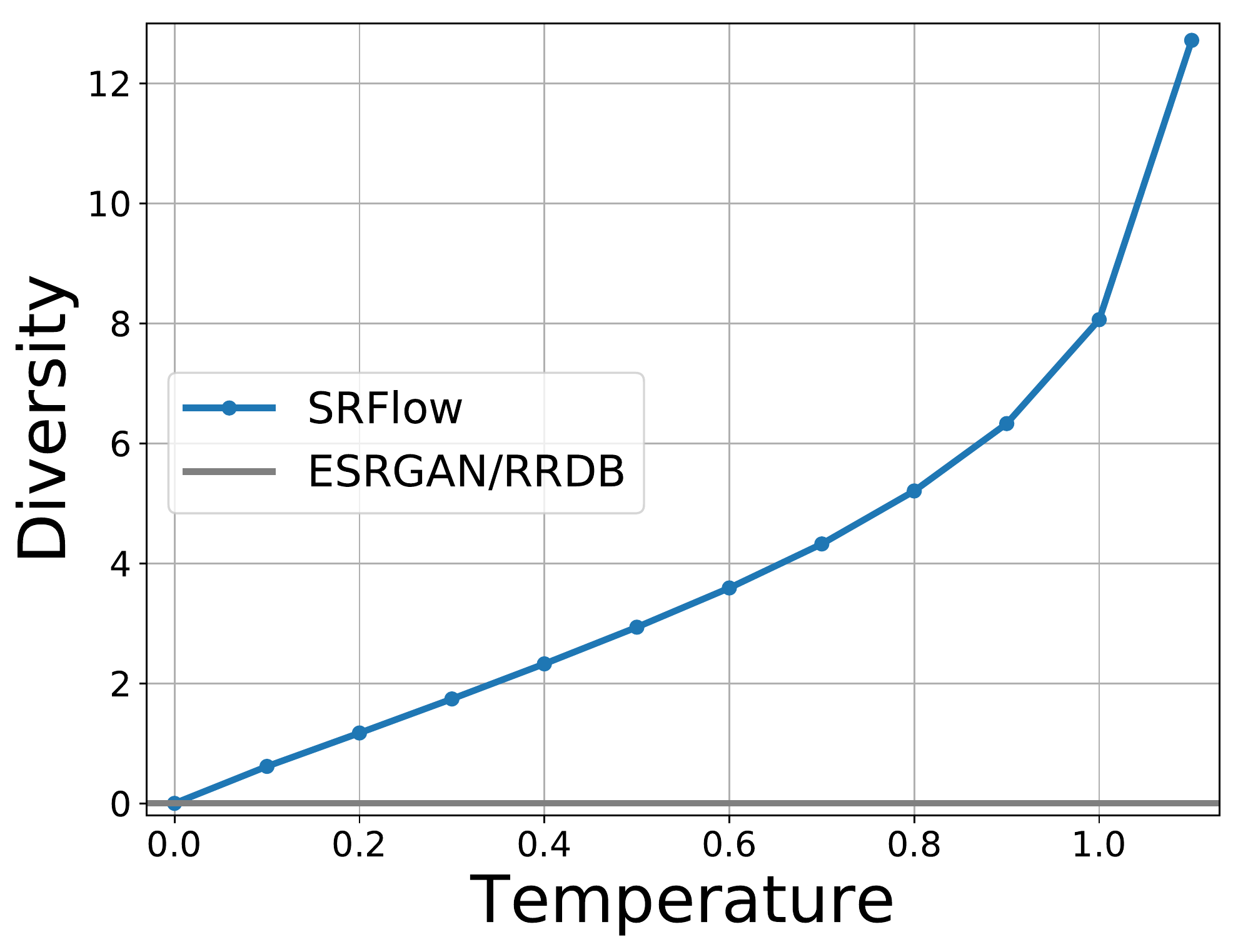}

\vspace{-3mm}
\caption{Analysis of the sampling temperature $\tau$ in terms of PSNR, SSIM, LPIPS and sample diversity on CelebA ($8\times$). Results of RRDB~\cite{wang2018esrgan} and ESRGAN~\cite{wang2018esrgan} are provided for reference.}
\vspace{-3mm}
\label{fig:heats_celeba}
\end{figure}

\begin{figure}[t]
\centering
\newcommand{\size}{0.45}
\includegraphics[width=\size\linewidth,valign=t]{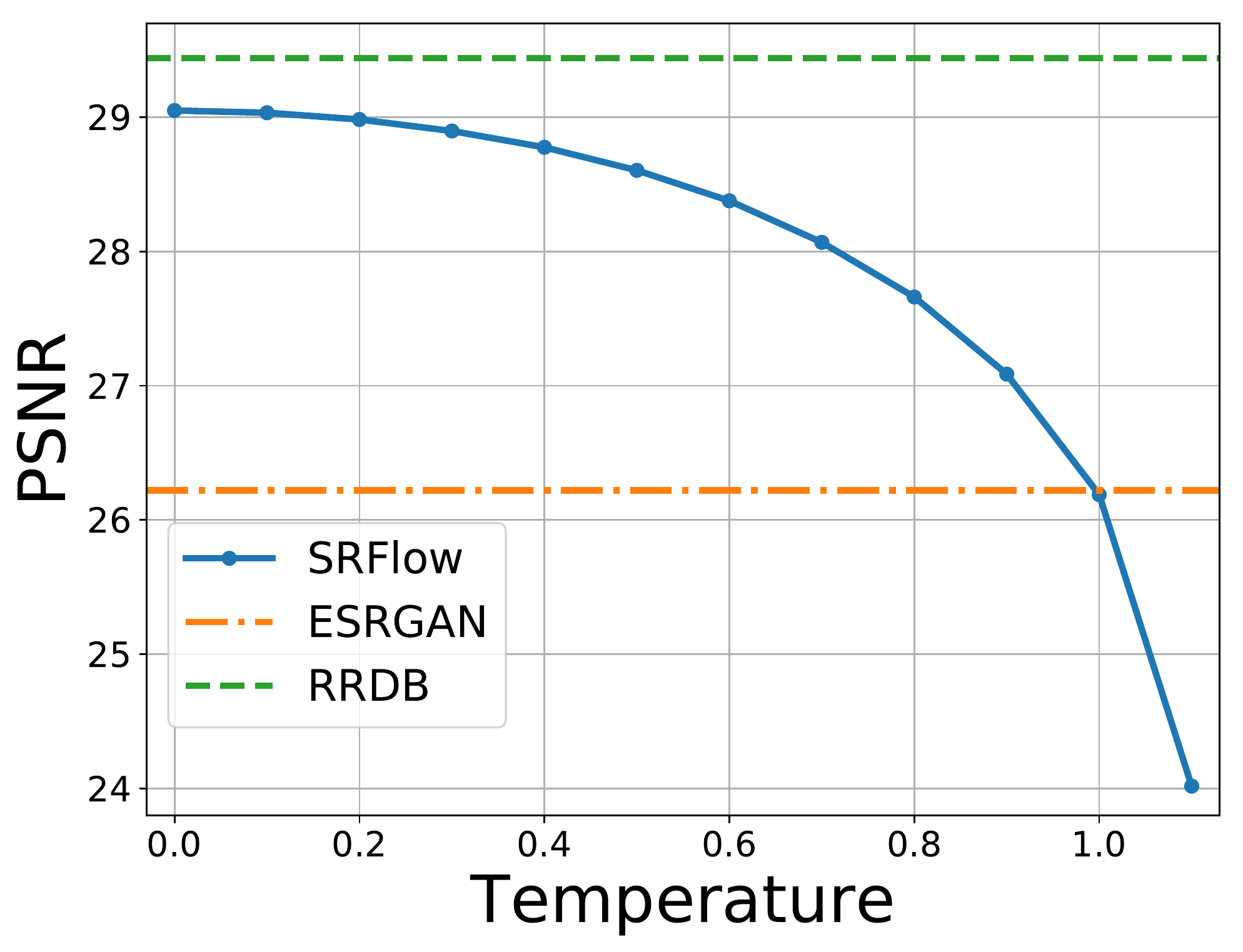}
\includegraphics[width=\size\linewidth,valign=t]{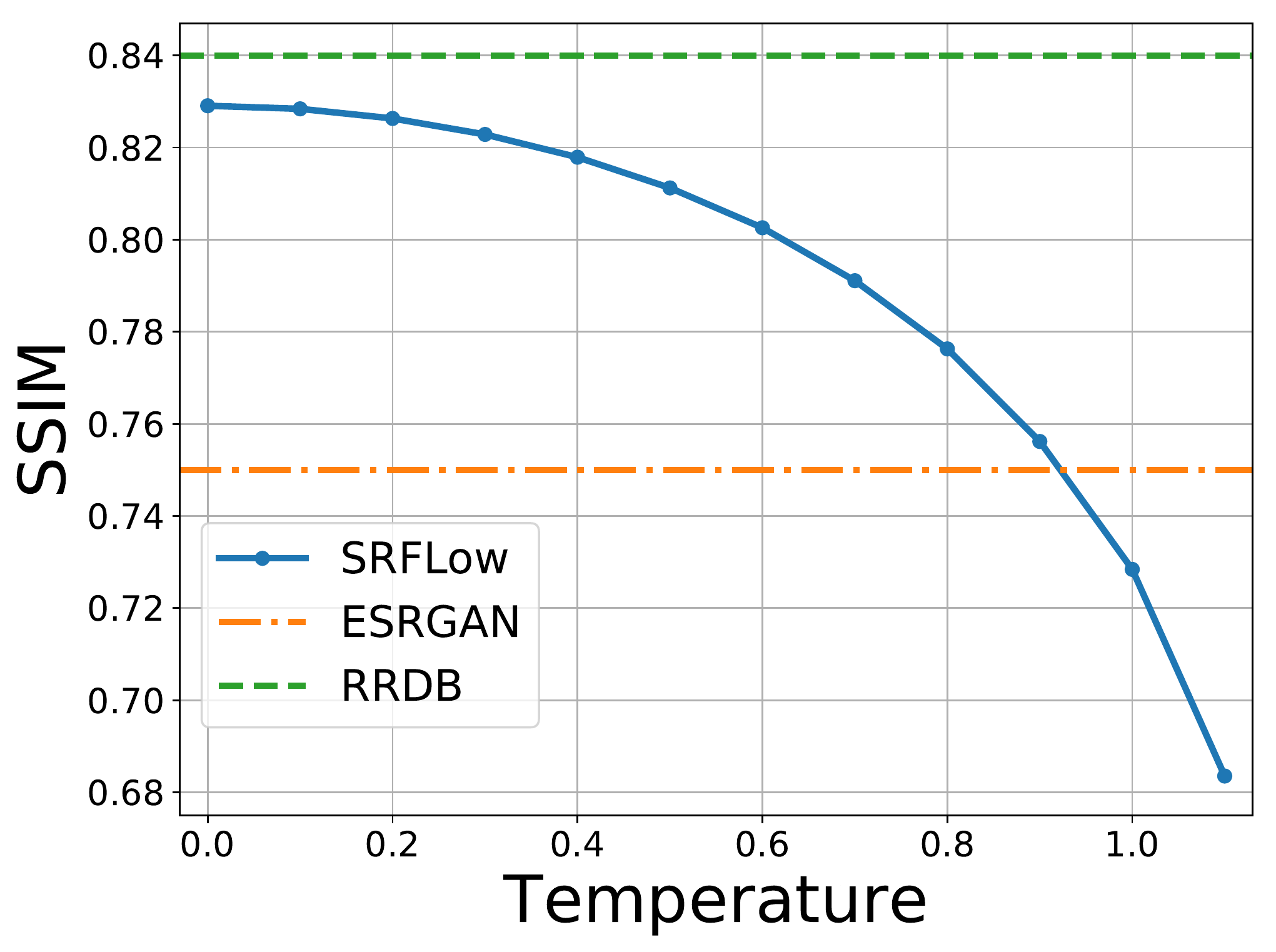}
\includegraphics[width=\size\linewidth,valign=t]{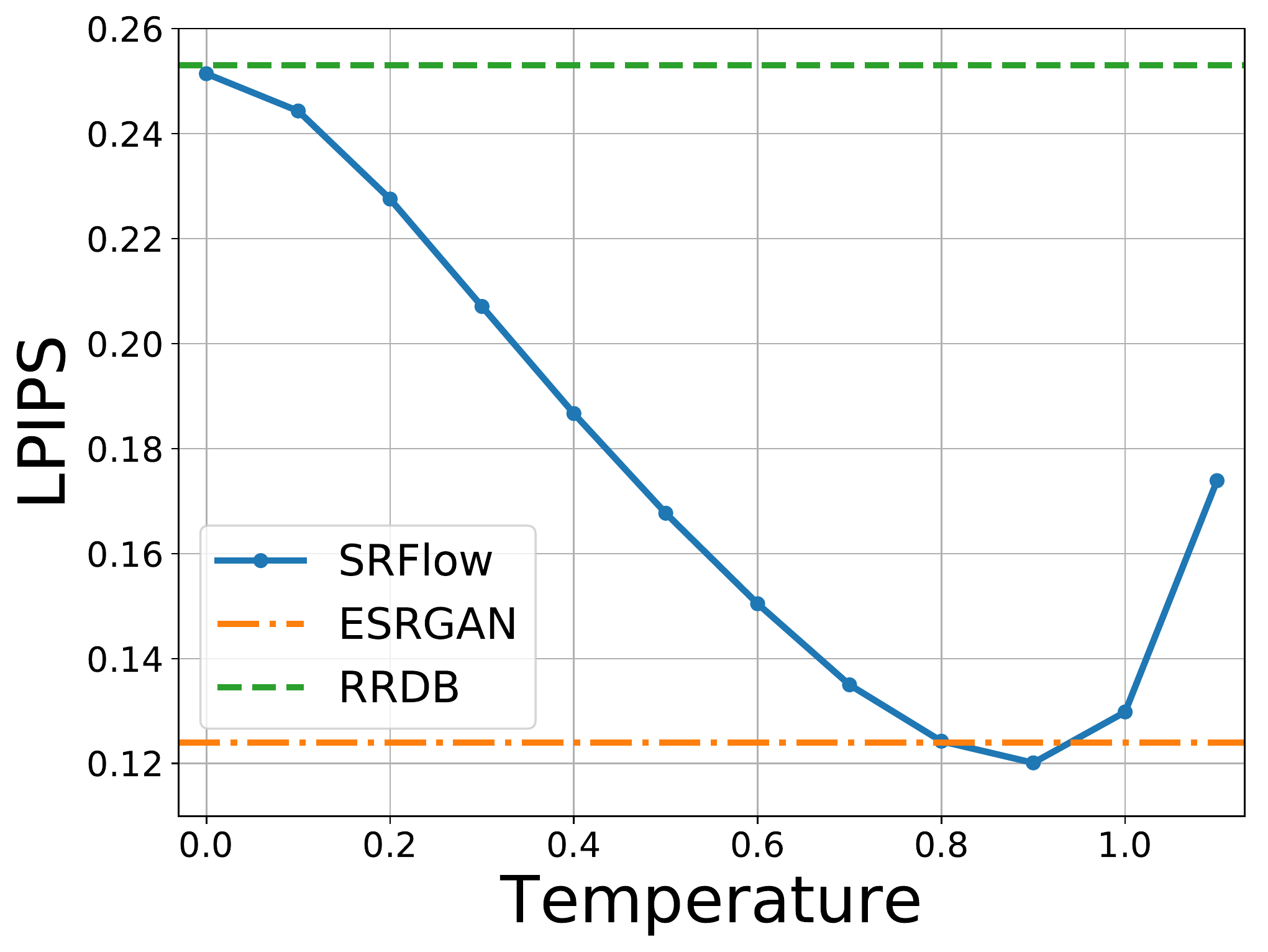}
\includegraphics[width=\size\linewidth,valign=t]{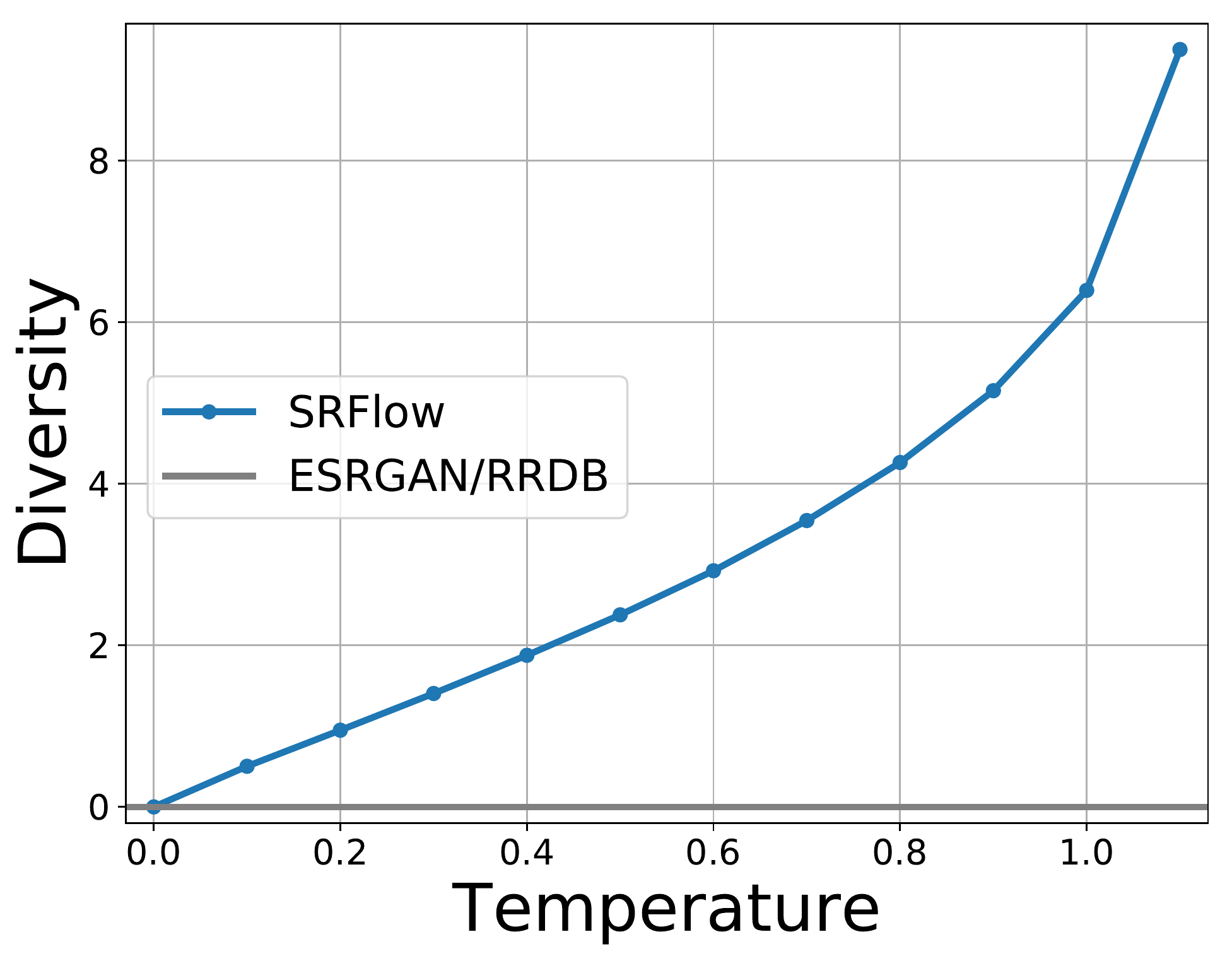}

\vspace{-3mm}
\caption{Analysis of the sampling temperature $\tau$ in terms of PSNR, SSIM, LPIPS and sample diversity on DIV2K ($4\times$). Results of RRDB~\cite{wang2018esrgan} and ESRGAN~\cite{wang2018esrgan} are provided for reference.}
\vspace{-3mm}
\label{fig:heats_div2k4}
\end{figure}

\begin{figure}[t]
\centering
\newcommand{\size}{0.45}
\includegraphics[width=\size\linewidth,valign=t]{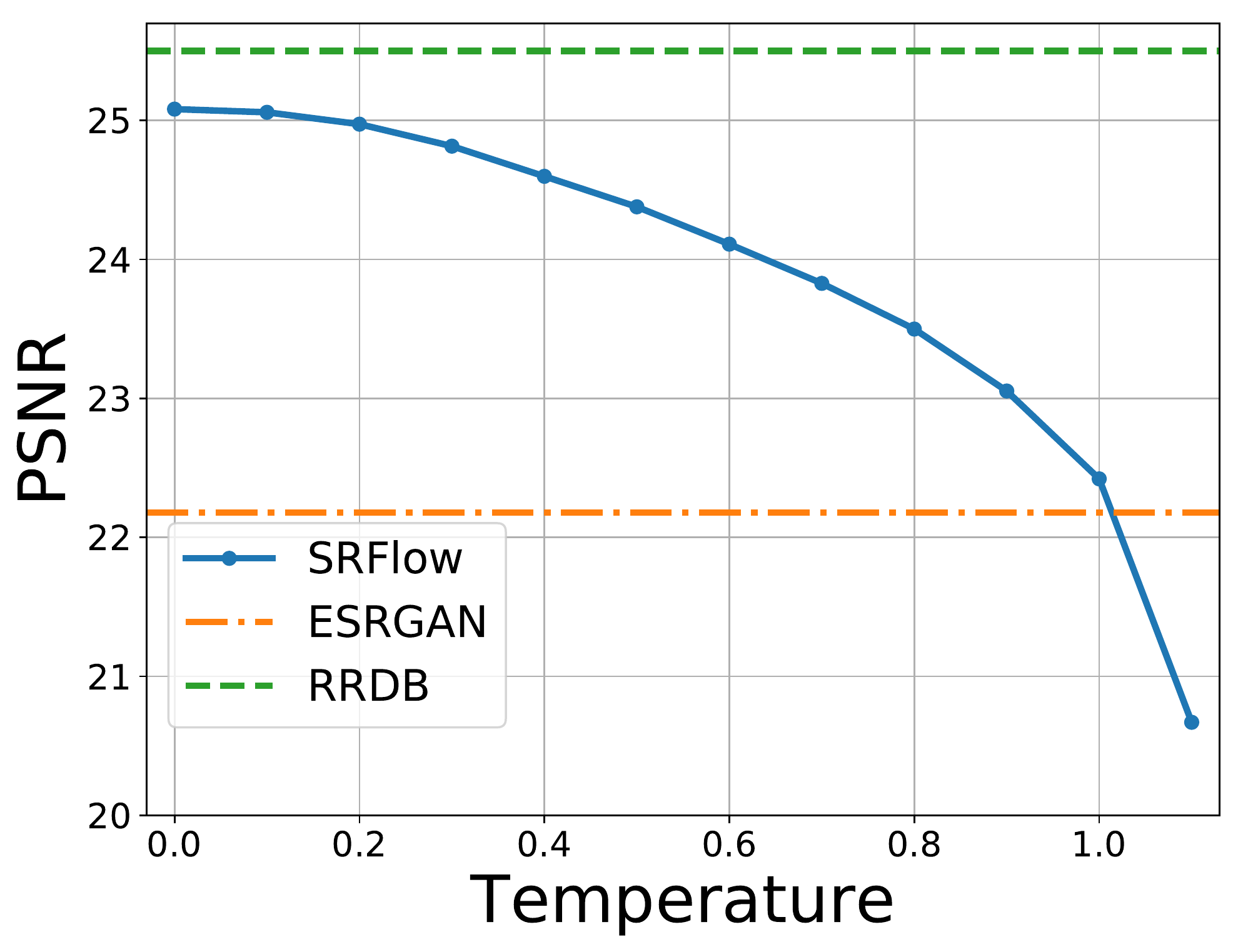}
\includegraphics[width=\size\linewidth,valign=t]{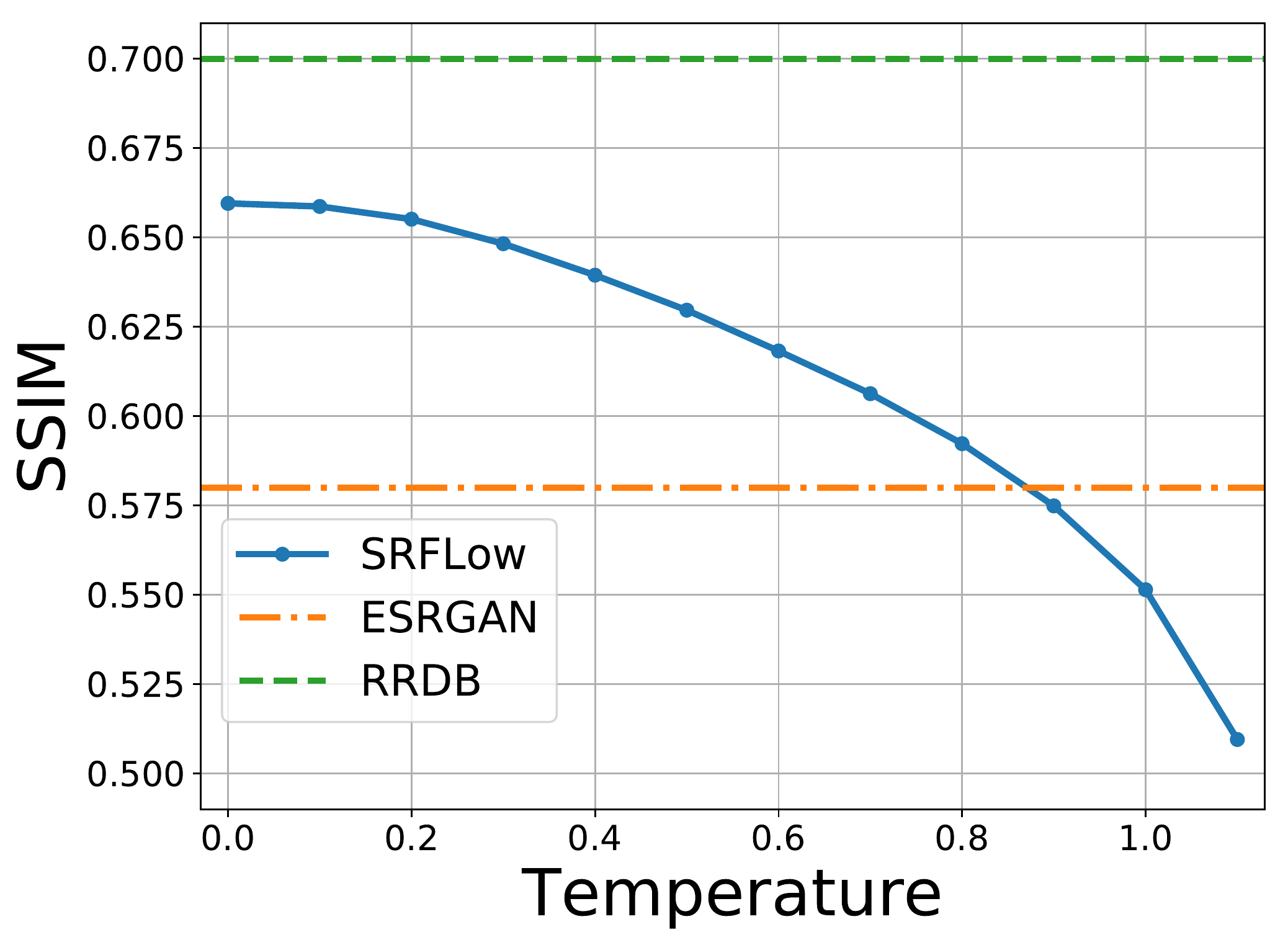}
\includegraphics[width=\size\linewidth,valign=t]{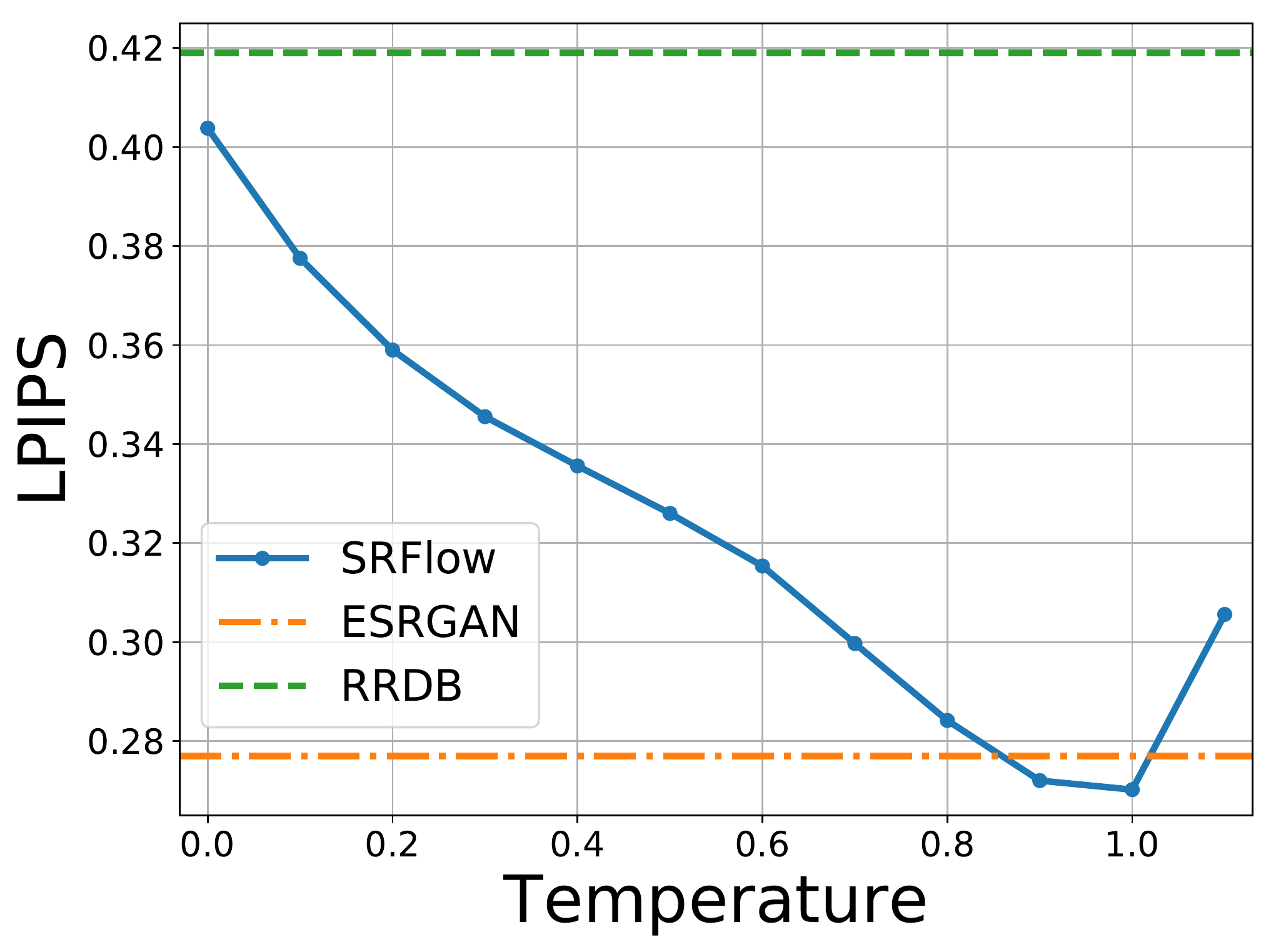}
\includegraphics[width=\size\linewidth,valign=t]{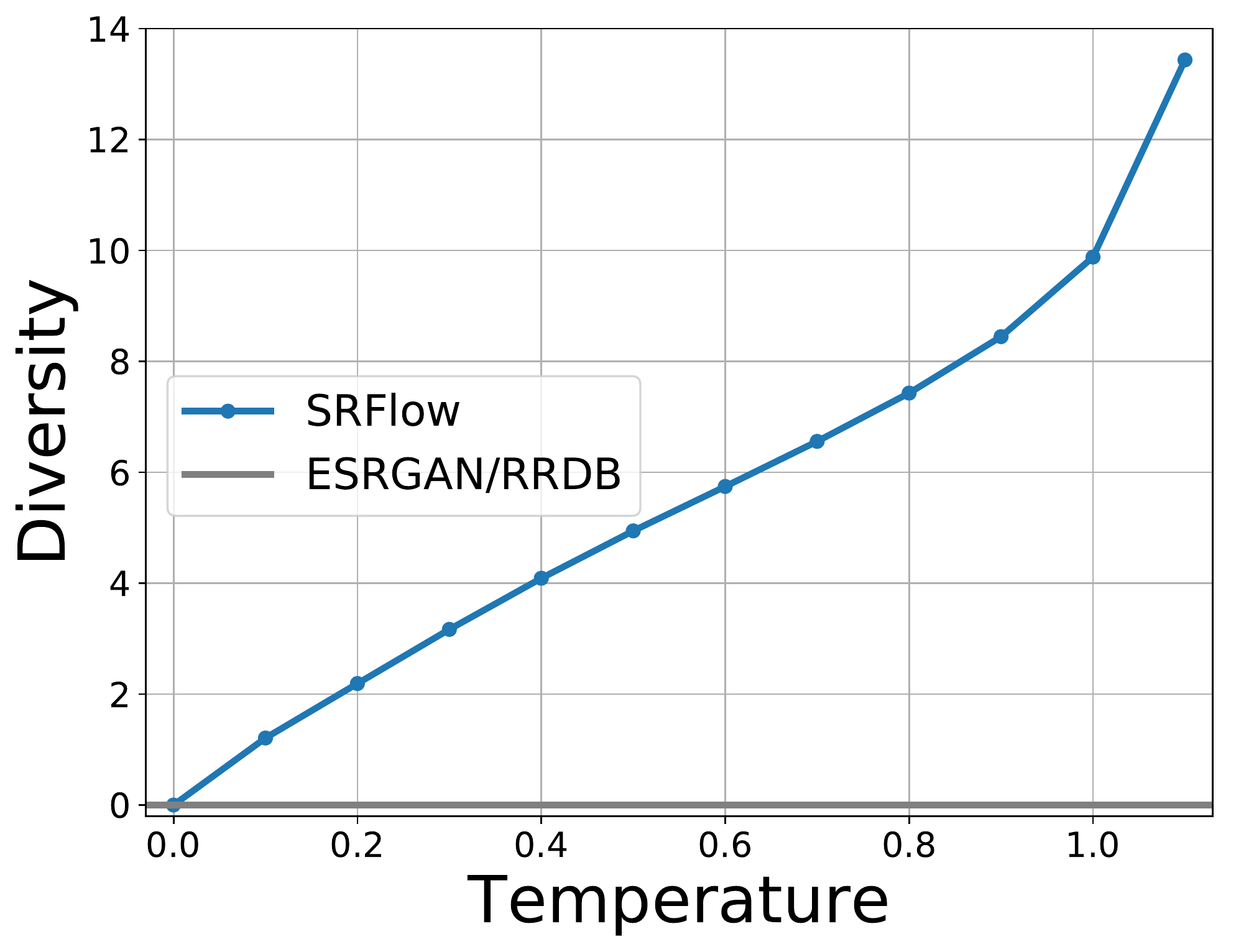}

\vspace{-3mm}
\caption{Analysis of the sampling temperature $\tau$ in terms of PSNR, SSIM, LPIPS and sample diversity on DIV2K ($8\times$). RRDB~\cite{wang2018esrgan} and ESRGAN~\cite{wang2018esrgan} are used as reference.}
\vspace{-5mm}
\label{fig:heats_div2k8}
\end{figure}

Here, we analyze the impact of the sampling temperature $\tau$ used during inference. It controls the variance of the Gaussian latent variable used when sampling SR images as $\by = \fT^{-1}(\bz; \bx),\, \bz \sim \mathcal{N}(0, \tau)$.
As described in Section~4.1 of the main paper, a slightly reduced temperature $\tau < 1$, increases the image quality.
When further decreasing the temperature to $\tau = 0$, the sampling process becomes deterministic. We analyze the effect of the sampling temperature $\tau$ on the main performance metrics, and on the sampling diversity itself.
Results are shown in Figures~\ref{fig:heats_celeba},~\ref{fig:heats_div2k4}~and~\ref{fig:heats_div2k8}. A temperature $\tau=0$ generates predictions with high fidelity, in terms of PSNR and SSIM. However, the results are blurry, as seen in Figure~\ref{fig:heat_examples}, explaining the poor perceptual quality (LPIPS) for this setting. Increasing the temperature leads to a drastic improvements in perceptual quality in terms of LPIPS distance. This is also clearly seen in the visual results in Figure~\ref{fig:heat_examples}.
We also plot how the sampling diversity improves with increased temperature $\tau$ in terms of pixel-wise variance.

\clearpage

\subsection{Perception--Distortion analysis}

\begin{figure}[t]
\centering
\newcommand{\size}{0.46}
\subfloat[CelebA]{%
\includegraphics*[trim=0 0 0 0,width=\size\linewidth]{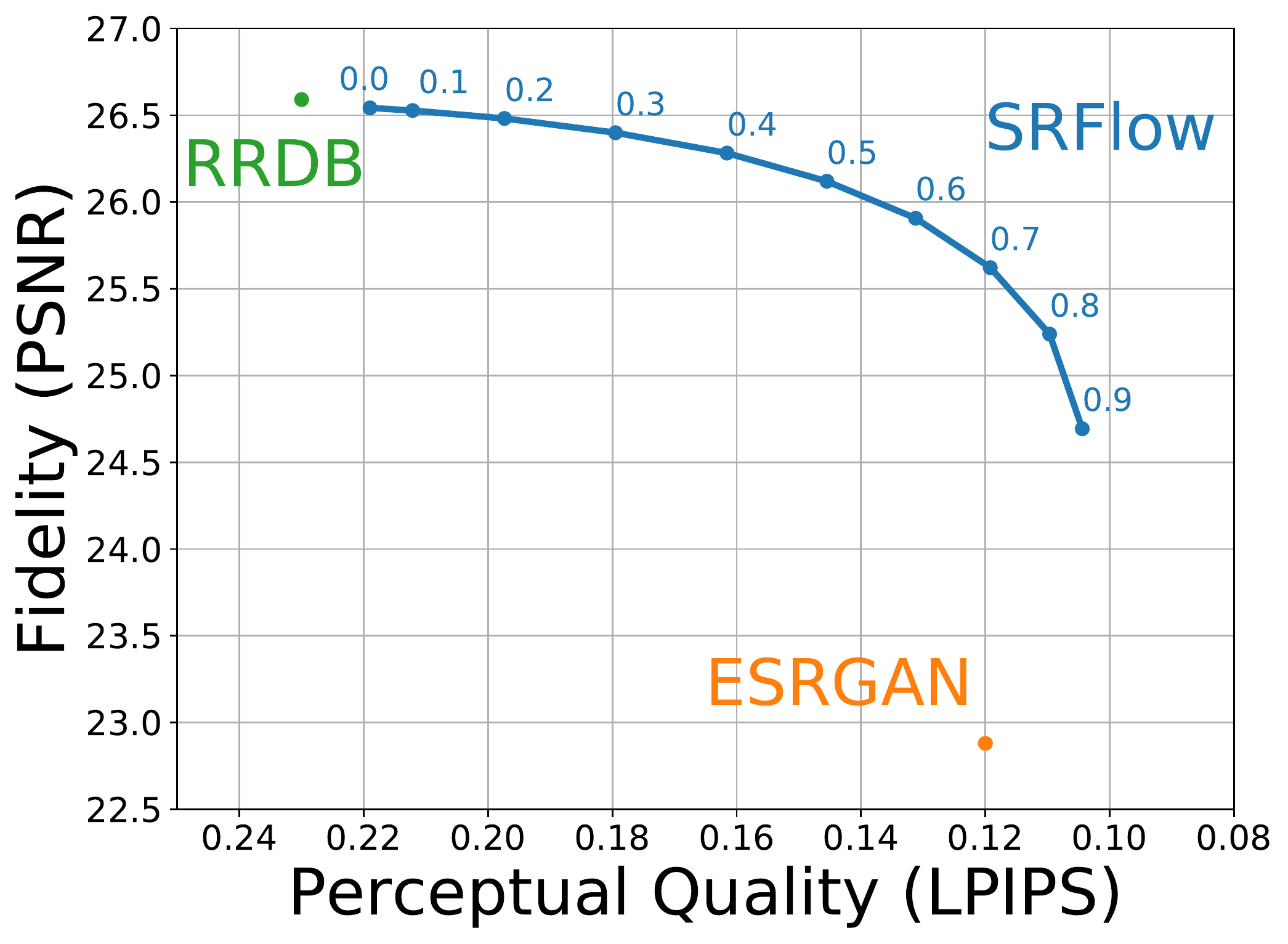}
}
\subfloat[DIV2K ($4\times$)]{%
\includegraphics*[trim=0 0 0 0,width=\size\linewidth]{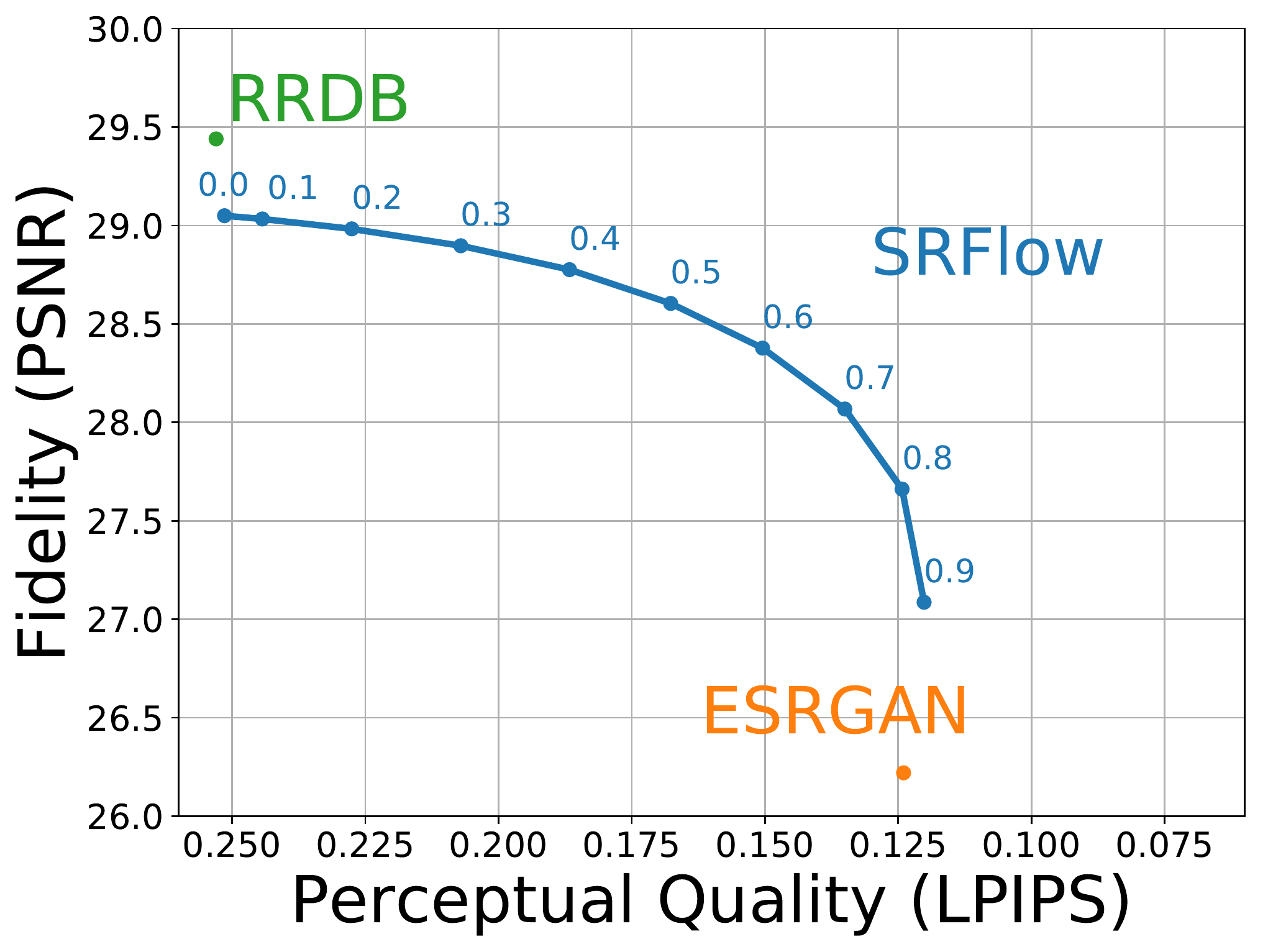}
}\\
\subfloat[DIV2K ($8\times$)]{%
\includegraphics*[trim=0 0 0 0,width=\size\linewidth]{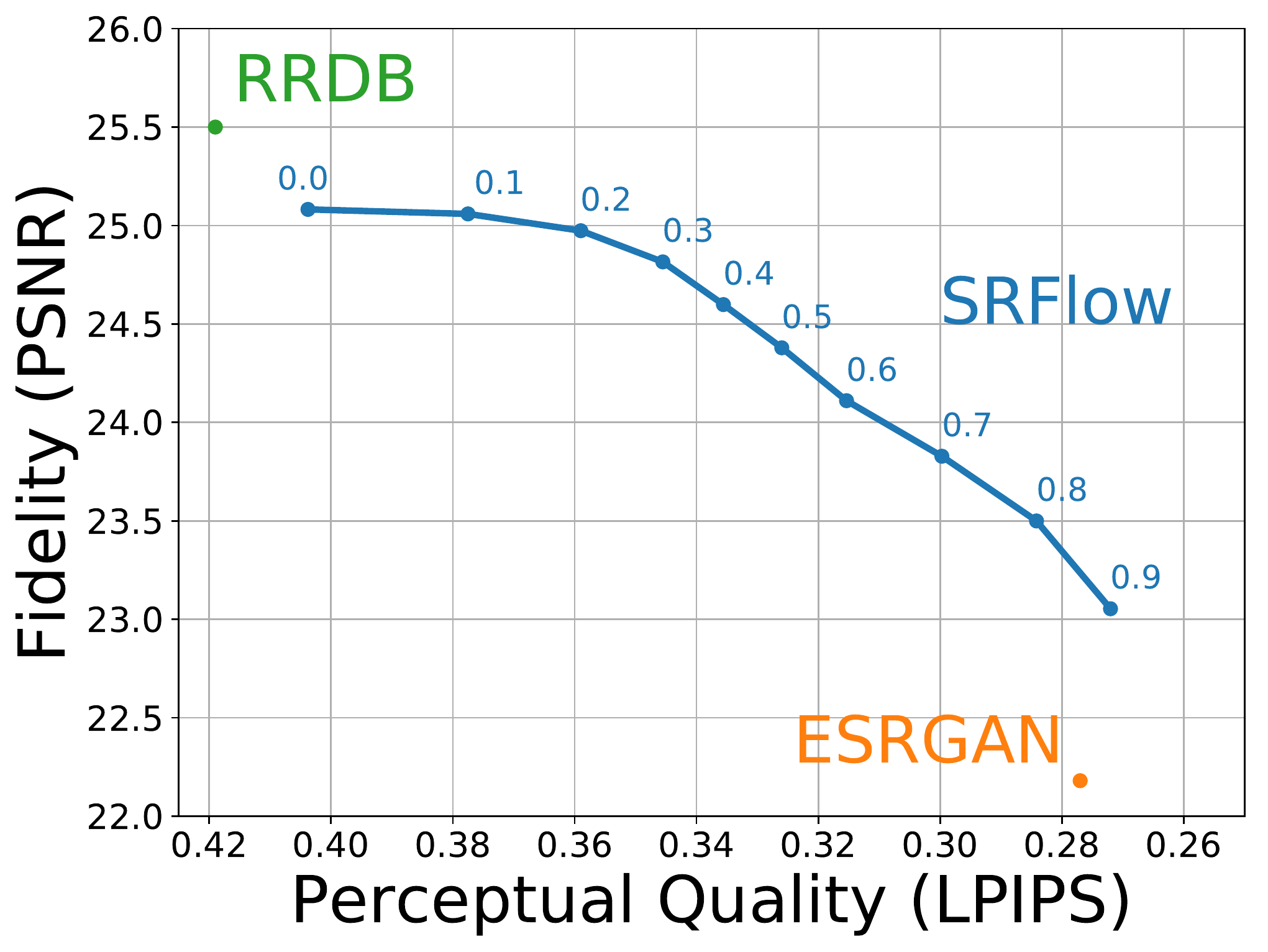}
}
\vspace{-3mm}
\caption{Analysis of the trade-off between perceptual quality and fidelity (distortion). SRFlow allows the trade-off to be controlled by varying the sampling  temperature $\tau$. In comparison, RRDB~\cite{wang2018esrgan} and ESRGAN~\cite{wang2018esrgan} provide only a single operating point each.}
\vspace{0mm}
\label{fig:perception_distortion}
\end{figure}

Here, we analyze the perception--distortion trade-off provided by our SRFlow. This trade off is an important choice decision for super-resolution \mbox{methods~\cite{ledig2017photo,BlauM18PerceptionDistTradeoff}.}
While most techniques do not allow to influence the super-resolution process during inference, SRFlow provides an effective means of controlling this trade-off using the sampling temperature $\tau$. We analyze this by plotting the perceptual quality (LPIPS) vs.\ the distortion (PSNR) with respect to the ground-truth in Figure~\ref{fig:perception_distortion}. We plot the results for different $\tau$ for SRFlow. Our approach provides different alternative trade-offs. It achieves similar PSNR compared to the $L_1$-loss trained RRDB~\cite{wang2018esrgan} for $\tau=0$. On the other hand, SRFlow provides similar or better perceptual quality compared to ESRGAN~\cite{wang2018esrgan} for $\tau \geq 0.8$, while achieving superior fidelity (PSNR).

\subsection{Impact of LR-Encoder Initialization}

To efficiently compare different variants of SRFlow, we reduced training time by pretraining the LR-Encoder $\gT$. As shown in Table~\ref{tab:rrdb_pretraining}, the perceptual quality is comparable, while the fidelity is slightly higher, compared to using a randomly initalized LR-Encoder. The default SRFlow network was trained for 200k steps and uses a pretrained LR-Encoder, which was trained for 200k steps. The model without pretraining was trained for 300k iterations to make up for the missing pretraining. Since the main bottleneck during training is the calculation of the log determinant, this reduces training time.

\begin{table}[b]
    \centering%
    \caption{Quantitative comparison on CelebA between training the SRFlow model with and without first pretraining the LR-Encoder $\gT$.}\vspace{-3mm}
    \begin{tabular}{l@{~~~}ccc}
        \toprule
         &  PSNR & SSIM & LPIPS \\
        \midrule
        Pretrained LR-Encoder & 25.24 & 0.71 & 0.110 \\
        Without pretrained LR-Encoder &  25.06 & 0.70 & 0.108 \\
        \bottomrule
    \end{tabular}
    \vspace{0mm}
    \label{tab:rrdb_pretraining}
\end{table}

\clearpage

\subsection{Oracle Analysis of the Sampling Space}

\begin{figure}[t]
\centering
\newcommand{\size}{0.13}
\hspace{0.13\linewidth}\includegraphics[width=\size\linewidth]{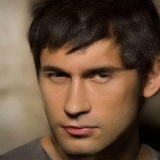}
\includegraphics[width=\size\linewidth]{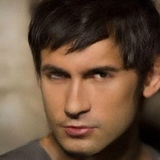}
\includegraphics[width=\size\linewidth]{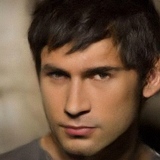}
\includegraphics[width=\size\linewidth]{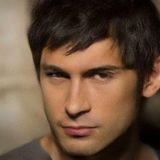}
\includegraphics[width=\size\linewidth]{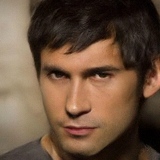}
\includegraphics[width=\size\linewidth]{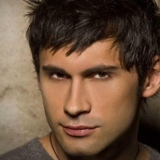}
\resizebox{\linewidth}{!}{
    \begin{tabular}{ C{2cm} C{2cm} C{2cm} C{2cm} C{2cm} C{2cm} C{2cm} }
  \#Samples $n$ & 1     & 10    & 100   & 1000  & 10000 & GT \\
  LPIPS      & 0.108 & 0.105 & 0.099 & 0.098 & 0.093 & 0
\end{tabular}
}
\vspace{-5mm}
\caption{Best of $n$ super-resolved ($8\times$) images in terms of the LPIPS metric.}
\vspace{-1mm}
\label{fig:oracle}
\end{figure}

\begin{figure}[t]
\centering
\newcommand{\size}{0.42}
\includegraphics[width=\size\linewidth]{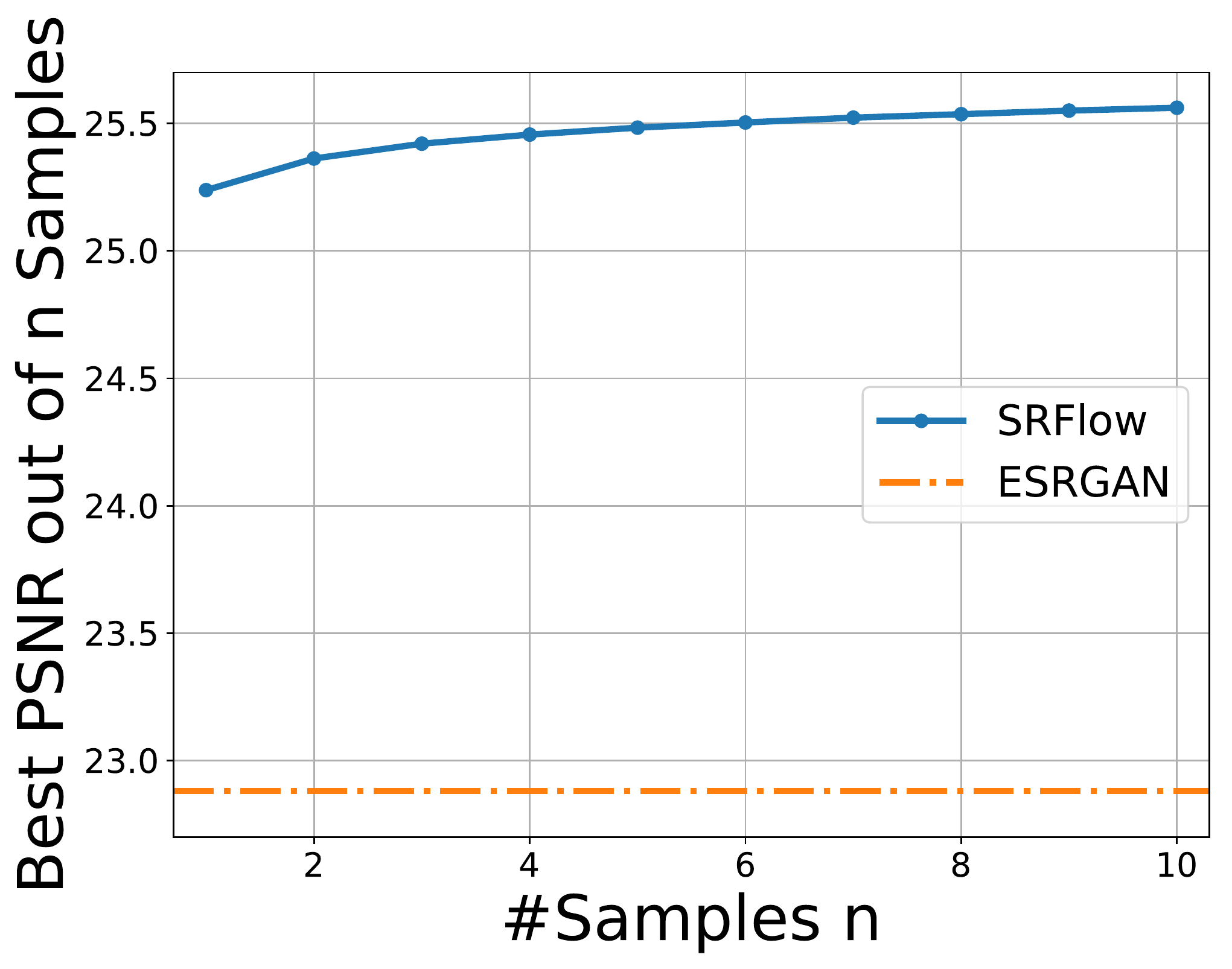}~~~
\includegraphics[width=\size\linewidth]{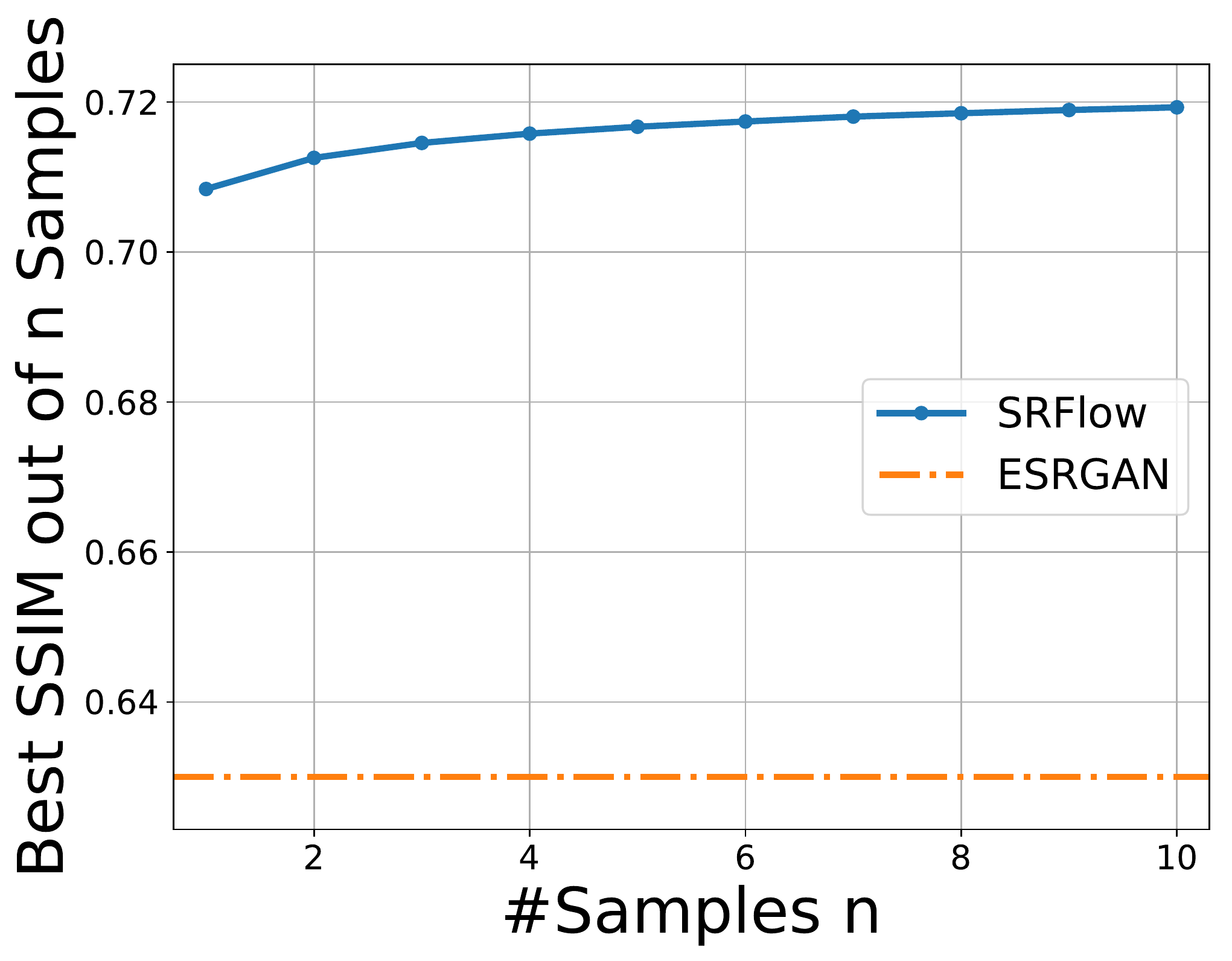}
\includegraphics[width=\size\linewidth]{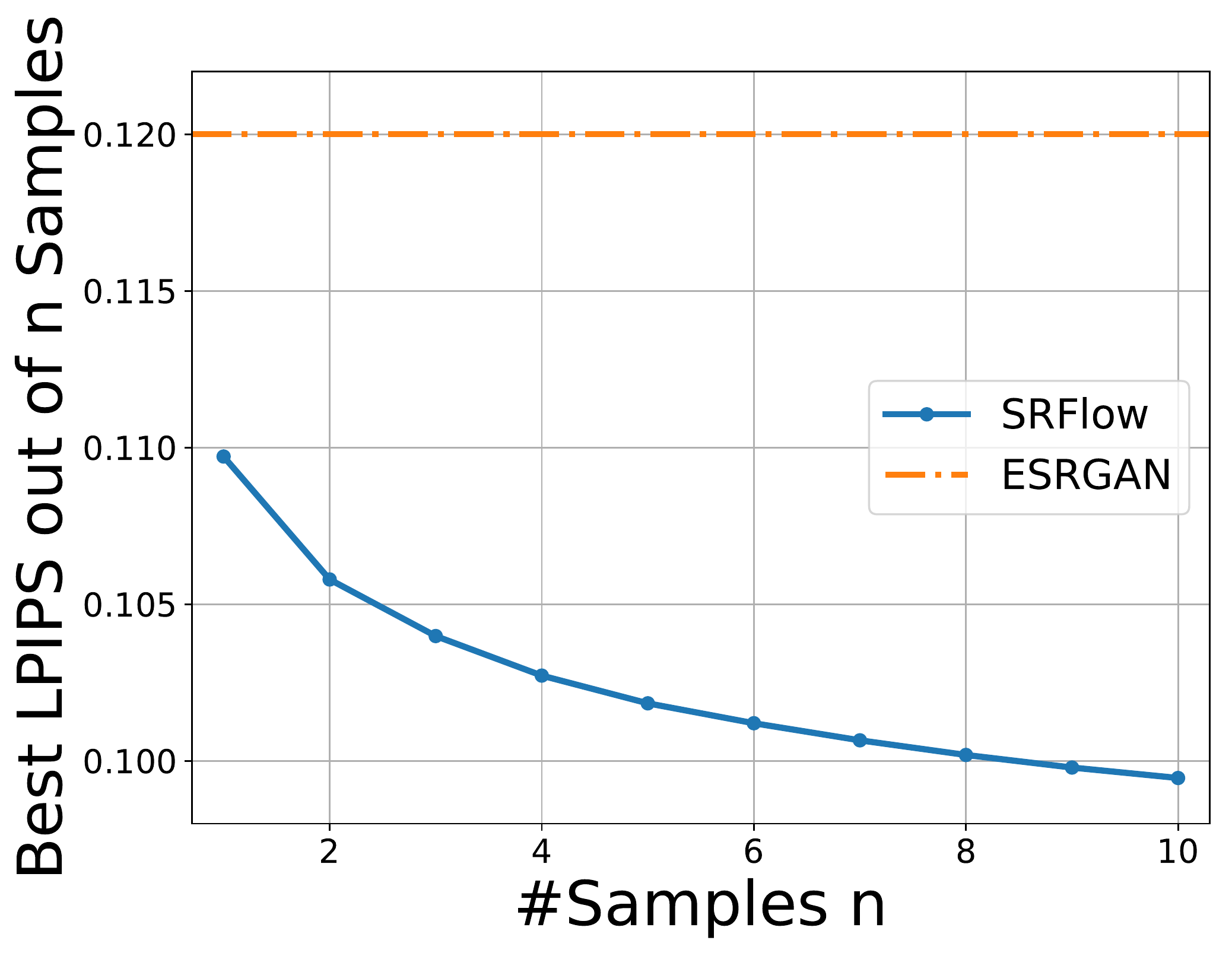}
\vspace{-4mm}
\caption{Analysis of the improvement in performance metrics when choosing the best out of $n$ samples. The performance of ESRGAN~\cite{wang2018esrgan} is included for reference.}
\vspace{-5mm}
\label{fig:oracle-results}
\end{figure}

As opposed to other state-of-the-art super-resolution approaches, SRFlow can be used to sample many variants of plausible super-resolutions. To further demonstrate the potential of this property, we analyze the performance of our SRFlow when selecting the best result among $n$ random samples. Results, using a sampling temperature of $\tau = 0.8$, are shown in Figure~\ref{fig:oracle-results}. The results are computed over the full CelebA test set of 5000 images. The best result w.r.t.\ the ground-truth in each plot is selected based on the corresponding performance metric for $n=1, \ldots, 10$ samples. This results shows that the perceptual quality in particular benefits from the oracle selection. This might be explained by our temperature setting, which forces the model to prefer perceptual quality over fidelity. It demonstrates that SRFlow provides a rich and diverse space of super-resolved images, from which solutions can be sampled. It provides the opportunity for improving the predictions of SRFlow by rejecting lower quality samples. A visual example is shown in Figure~\ref{fig:oracle}, when selecting the best out of $n$ samples using the LPIPS distance.

\clearpage

\begin{table}[t]

    \centering
    \caption{SRFlow results for image denoising on CelebA and DIV2K. Measurements for original images with Gaussian noise $\sigma=20$, images that were super-resolved after downsampling, and restored images that use our latent space normalization approach, which also exploits the original HR image. We use the SRFlow model trained for $8\times$ on CelebA and $4\times$ on DIV2k}\vspace{-2mm}
    
    \label{tab:denosing}
\end{table}

\subsection{Image Restoration}

\begin{figure}[t]
\centering
\newcommand{\size}{0.155}
\newcommand{\img}[2]{
\includegraphics[width=\size\linewidth]{figures/denoising/images/gt/#1}
\includegraphics[width=\size\linewidth]{figures/denoising/images/sr/#1}
\includegraphics[width=\size\linewidth]{figures/denoising/images/enhanced/#1}
\includegraphics[width=\size\linewidth]{figures/denoising/images/gt/#2}
\includegraphics[width=\size\linewidth]{figures/denoising/images/sr/#2}
\includegraphics[width=\size\linewidth]{figures/denoising/images/enhanced/#2}
}
\img{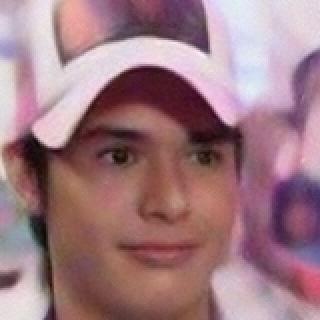}{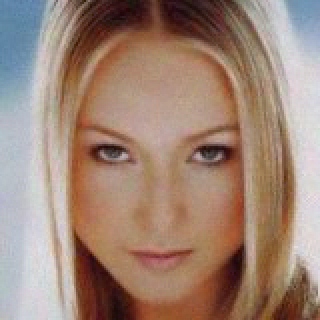}
\img{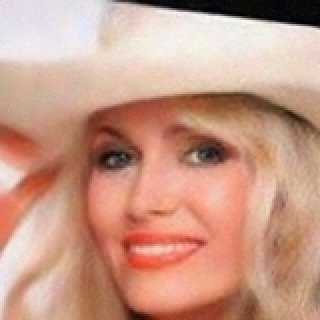}{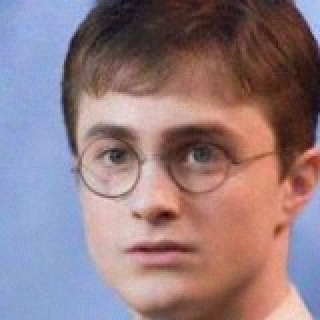}
\resizebox{\linewidth}{!}{
    \begin{tabular}{ C{2.5cm} C{2.5cm} C{2.5cm} C{2.5cm} C{2.5cm} C{2.5cm} C{2.5cm} }
        Original (Noise $\sigma = 20$) & Direct SR & Restored & Original (Noise $\sigma = 12$ + JPEG) & Direct SR & Restored
\end{tabular}
}
\vspace{-5mm}
\caption{Image restoration examples on CelebA images with different degradations. Directly super-resolving ($8\times$) the LR of the original removes noise but does not preserve details. Our SRFlow restoration also directly employs the original image by performing latent space normalization.} \vspace{-3mm}
\label{fig:denoising_celeba}
\end{figure}

\begin{figure}[t]
\centering
\newcommand{\size}{0.32}
\newcommand{\img}[2]{
\includegraphics[width=\size\linewidth]{figures/denoising_div2k/images/g20_j100_h09_Df2k4DIV2K_#1_gt}
\includegraphics[width=\size\linewidth]{figures/denoising_div2k/images/g20_j100_h09_Df2k4DIV2K_#1_sr}
\includegraphics[width=\size\linewidth]{figures/denoising_div2k/images/g20_j100_h09_Df2k4DIV2K_#1_enhanced}
\vspace{1mm}
\includegraphics[width=\size\linewidth]{figures/denoising_div2k/images/g20_j100_h09_Df2k4DIV2K_#2_gt}
\includegraphics[width=\size\linewidth]{figures/denoising_div2k/images/g20_j100_h09_Df2k4DIV2K_#2_sr}
\includegraphics[width=\size\linewidth]{figures/denoising_div2k/images/g20_j100_h09_Df2k4DIV2K_#2_enhanced}
}
\img{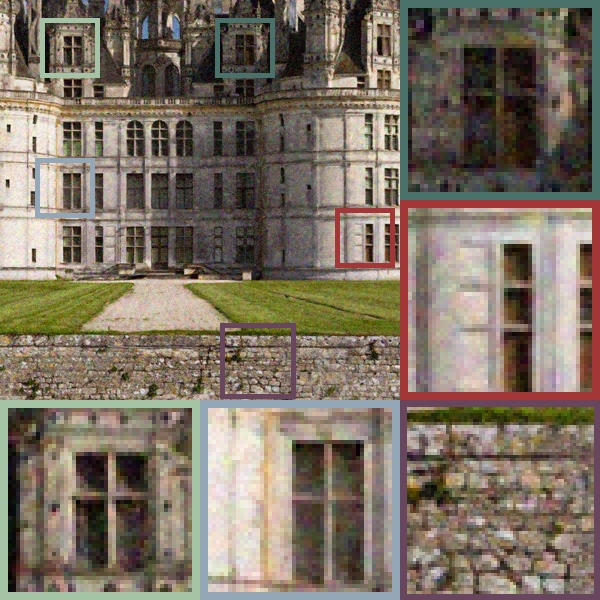}{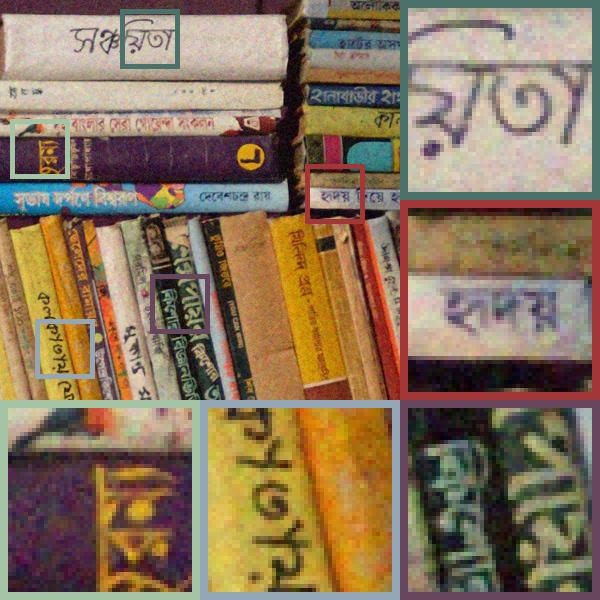}
\resizebox{\linewidth}{!}{
    \begin{tabular}{ C{2.5cm} C{2.5cm} C{2.5cm} C{2.5cm} C{2.5cm} C{2.5cm} C{2.5cm} }

        Original & Direct SR & Restored
\end{tabular}
}
\vspace{-4mm}
\caption{Image denoising examples on DIV2k images. Directly super-resolving ($4\times$) the LR of the original removes noise but does not preserve details. Our SRFlow restoration also directly employs the original image by performing latent space normalization.} \vspace{0mm}
\vspace{-3mm}
\label{fig:denoising_div2k}
\end{figure}

We provide additional quantitative and qualitative results for image restoration, described in Section~4.5. Table~\ref{tab:denosing} shows quantitative results for the task of image denoising when using white Gaussian noise with standard deviation $\sigma=20$. We report performance metrics w.r.t.\ the clean ground-truth for the original noisy image, when just super-resolving the down-sampled image, and when using our restoration approach based on latent space normalization, as described in Section~4.5. Despite only being trained for the task of super-resolving clean images, our approach provides promising results for image denoising. This demonstrates the strong image posterior learned by our SRFlow. We show visual examples on CelebA and DIV2K in Figure~\ref{fig:denoising_celeba} and Figure~\ref{fig:denoising_div2k} respectively.

\clearpage

\section{Visual Results}
\label{sec:visual_results}

In this section, we provide additional visual results.

\begin{figure}
\centering%
\newcommand{\size}{0.135}%
\newcommand{\img}[1]{%
\includegraphics[width=0.019375\linewidth]{figures/sotaCMbic8/#1/lq}~~~~~~%
\includegraphics[width=\size\linewidth]{figures/sotaCMbic8/#1/RRDBn23}~~%
\includegraphics[width=\size\linewidth]{figures/sotaCMbic8/#1/ESRGANn23px2e-1}~~%
\includegraphics[width=\size\linewidth]{figures/sotaCMbic8/#1/Christmas}~~~~~~%
\includegraphics[width=\size\linewidth]{figures/sotaCMbic8/#1/Prog}~~%
\includegraphics[width=\size\linewidth]{figures/sotaCMbic8/#1/ChristmasP}~~~~~~%
\includegraphics[width=\size\linewidth]{figures/sotaCMbic8/#1/gt}%
}%
\img{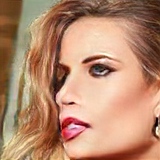}
\img{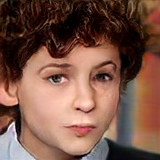}
\img{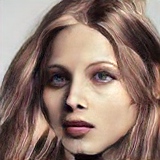}
\img{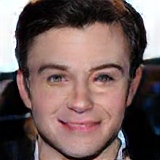}
\img{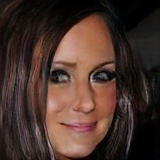}
\img{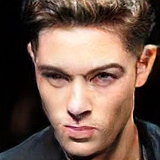}
\img{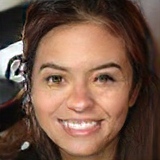}
\img{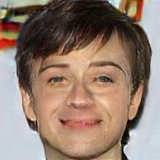}
\img{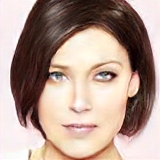}
\vspace{-1mm}

\resizebox{\linewidth}{!}{
    \begin{tabular}{ C{0.4cm} C{2.5cm} C{2.5cm} C{2.5cm} C{2.5cm} C{2.5cm} C{2.5cm} }
        LR & RRDB~\cite{wang2018esrgan}~~ & ESRGAN~\cite{wang2018esrgan} & \textbf{SRFlow} & ProgFSR~\cite{Kim19ProgressFSR} & \textbf{SRFlow} & Ground-Truth
\end{tabular}
}
\vspace{-3.5mm}
\caption{Comparison of our SRFlow with state-of-the-art for $8\times$ face super-resolution on CelebA. The three columns with super-resolutions on the left are trained and applied on bicubic downsampled images. The next two colums employ the bilinear kernel~\cite{Kim19ProgressFSR}.}
\vspace{-5mm}
\label{fig:sotaCMbic8-ap}
\end{figure}

\subsection{State-of-the-Art for Face Super-Resolution}
\label{ssec:sotaCeleba}

Additional examples that compare SRFlow with state-of-the-art  for face super-resolution on CelebA are shown in  Figure~\ref{fig:sotaCMbic8-ap}. For fair comparison, we also show SRFlow results when trained and applied on the same bilinear downsampling kernel as ProgFSR~\cite{Kim19ProgressFSR}. Our approach provides superior perceptual quality and better fidelity compared to the GAN-based methods.

\clearpage

\subsection{State-of-the-Art General Super-Resolution}

\begin{figure}[t]
\centering%
\newcommand{\size}{0.118}%
\newcommand{\img}[1]{%
\includegraphics[width=\size\linewidth]{figures/sotaDiv2k4-app/images/#1_lq.png}
\includegraphics[width=\size\linewidth]{figures/sotaDiv2k4-app/images/#1_Bicubic_Df2k4.png}
\includegraphics[width=\size\linewidth]{figures/sotaDiv2k4-app/images/#1_EDSR.png}
\includegraphics[width=\size\linewidth]{figures/sotaDiv2k4-app/images/#1_RRDBn23.png}
\includegraphics[width=\size\linewidth]{figures/sotaDiv2k4-app/images/#1_ESRGAN.png}
\includegraphics[width=\size\linewidth]{figures/sotaDiv2k4-app/images/#1_RankSRGAN.png}
\includegraphics[width=\size\linewidth]{figures/sotaDiv2k4-app/images/#1_Christmas.png}
\includegraphics[width=\size\linewidth]{figures/sotaDiv2k4-app/images/#1_gt.png}%

}
\img{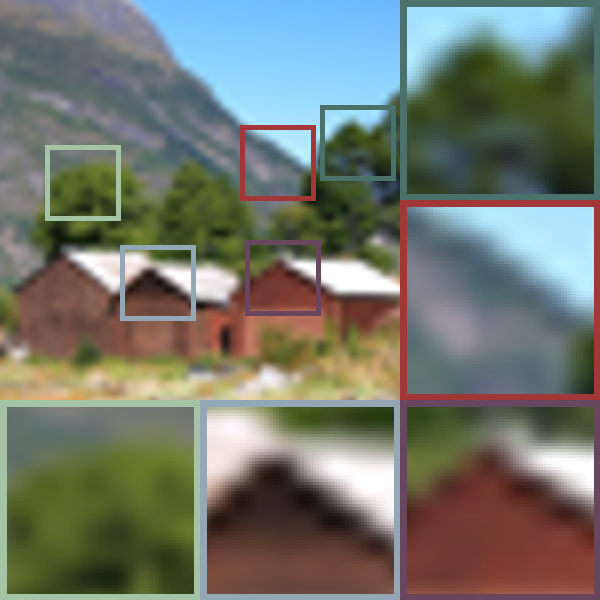}
\img{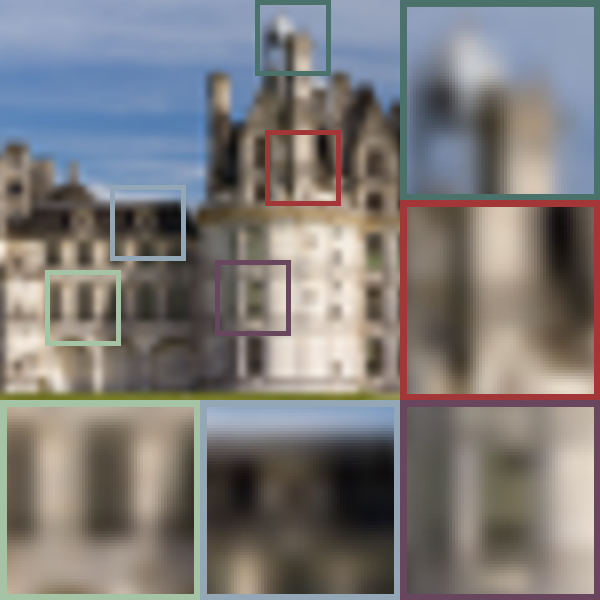}
\img{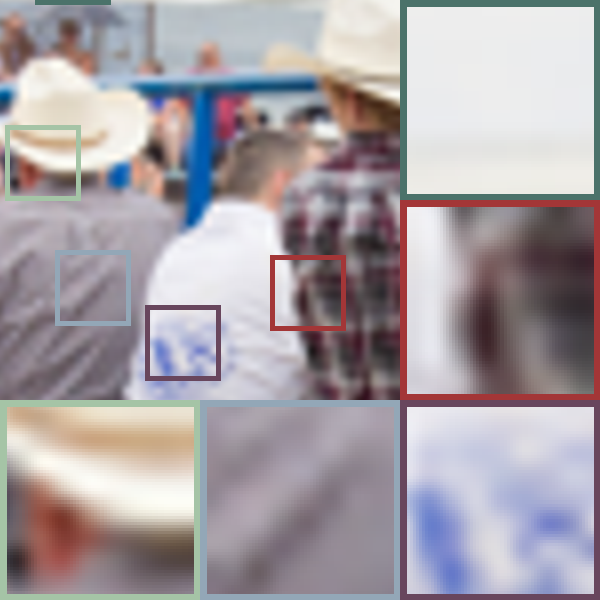}
\img{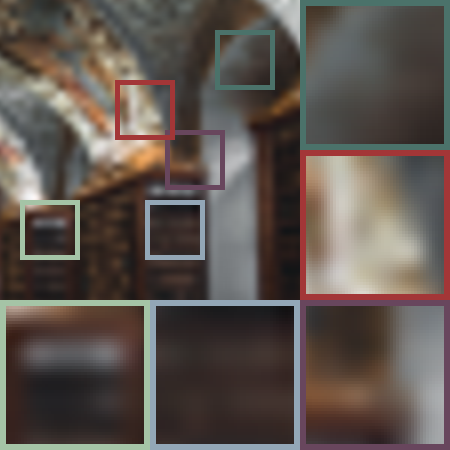}%
\vspace{0mm}
\resizebox{\linewidth}{!}{
    \begin{tabular}{ C{2.5cm} C{2.5cm} C{2.5cm} C{2.5cm} C{2.5cm} C{2.5cm} C{2.5cm} C{2.5cm} }
        Low Resolution & Bicubic & EDSR \cite{lim2017EDSR} & RRDB \cite{wang2018esrgan} & ESRGAN \cite{wang2018esrgan} & RankSRGAN~\cite{zhang2019ranksrgan} & \textbf{SRFlow} \small{$\tau=0.9$} & Ground Truth
\end{tabular}
}%
\vspace{0mm}
\caption{Comparison to state-of-the-art for general super-resolution on the DIV2k $4\times$ validation set.}
\vspace{0mm}
\label{fig:sotaDiv2k4-app}
\end{figure}

\begin{figure}[t]
\centering
\newcommand{\size}{0.155}
\newcommand{\img}[1]{
\includegraphics[width=\size\linewidth]{figures/sotaDiv2k8/images/#1_lq}
\includegraphics[width=\size\linewidth]{figures/sotaDiv2k8/images/#1_Bicubic_Df2k8}
\includegraphics[width=\size\linewidth]{figures/sotaDiv2k8/images/#1_RRDBn23}
\includegraphics[width=\size\linewidth]{figures/sotaDiv2k8/images/#1_ESRGAN}
\includegraphics[width=\size\linewidth]{figures/sotaDiv2k8/images/#1_Christmas}
\includegraphics[width=\size\linewidth]{figures/sotaDiv2k8/images/#1_gt}
}
\img{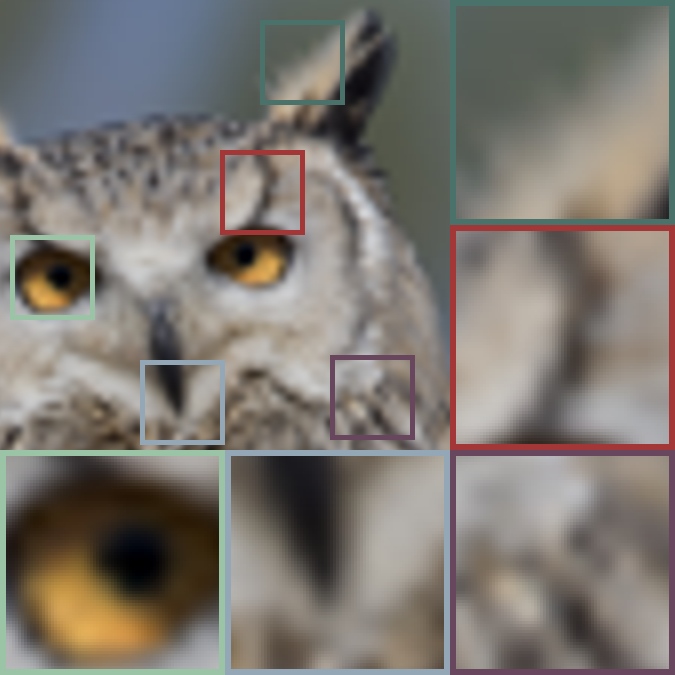}
\img{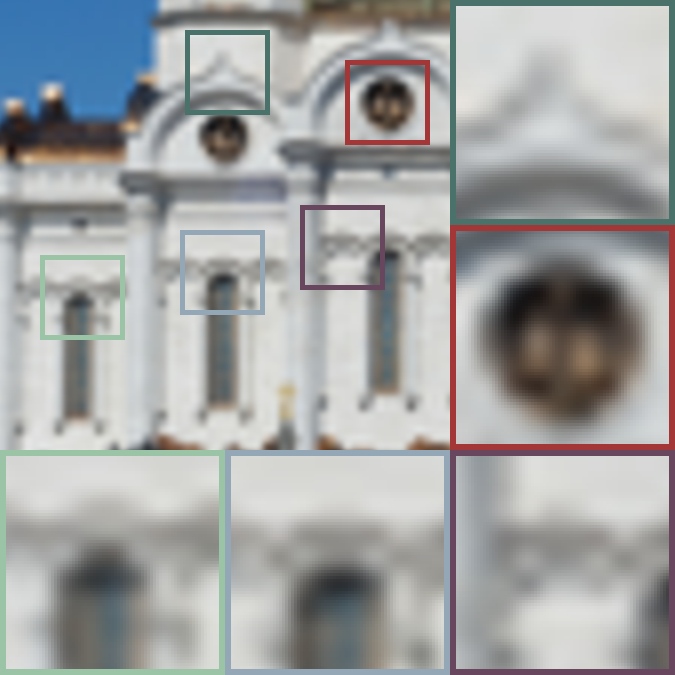}
\img{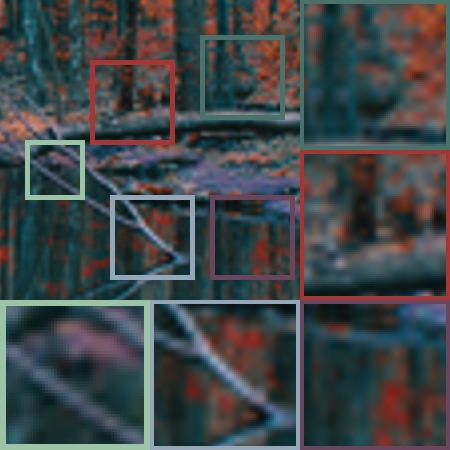}
\img{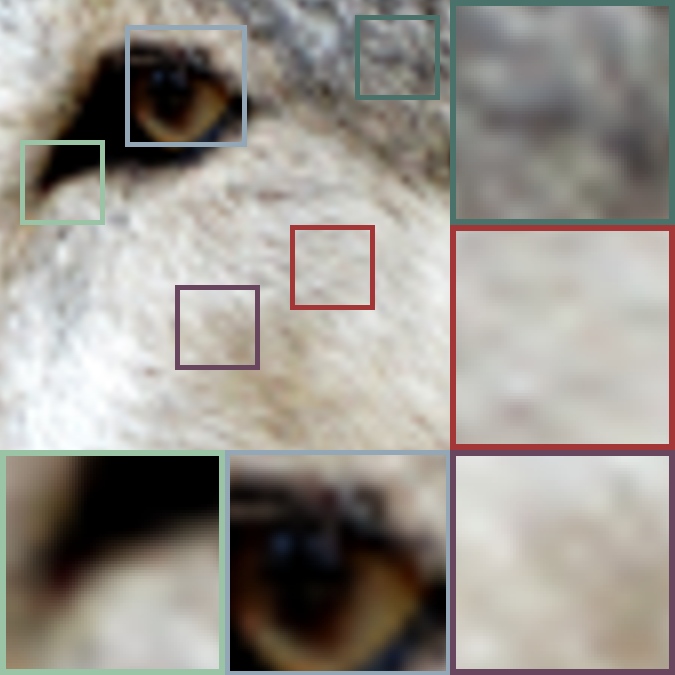}
\img{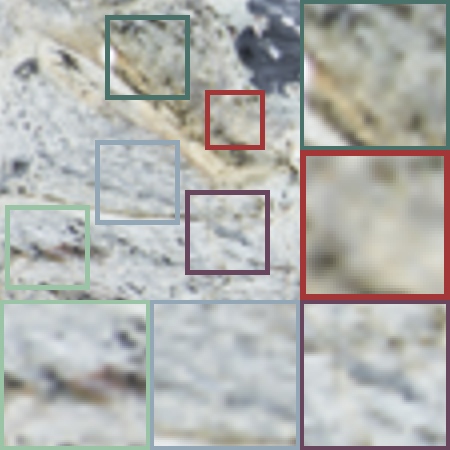}
\resizebox{\linewidth}{!}{
    \begin{tabular}{ C{2cm} C{2cm} C{2cm} C{2cm} C{2cm} C{2cm} C{2cm} }
        Low Resolution & Bicubic & RRDB & ESRGAN & \textbf{SRFlow} & Ground-Truth
\end{tabular}
}
\vspace{0mm}
\caption{Comparison to state-of-the-art for general super-resolution on the DIV2k $8\times$ validation set.}
\vspace{0mm}
\label{fig:sotaDiv2k8}
\end{figure}

We provide more visual examples for the experiments on DIV2K, comparing SRFlow with with state-of-the-art super-resolution methods.
In Figure~\ref{fig:sotaDiv2k4-app} illustrates results for $4\times$.
In addition, we provide results for DIV2K $8\times$ in Figure~\ref{fig:sotaDiv2k8}.
SRFlow achieves perceptual quality similar or better than ESRGAN in most cases.
Moreover, our approach do not suffer from the hallucination artifacts typically seen in GAN-based methods.

\subsection{Stochastic Face Super-Resolution}

\begin{figure}
\centering

\includegraphics[width=\linewidth]{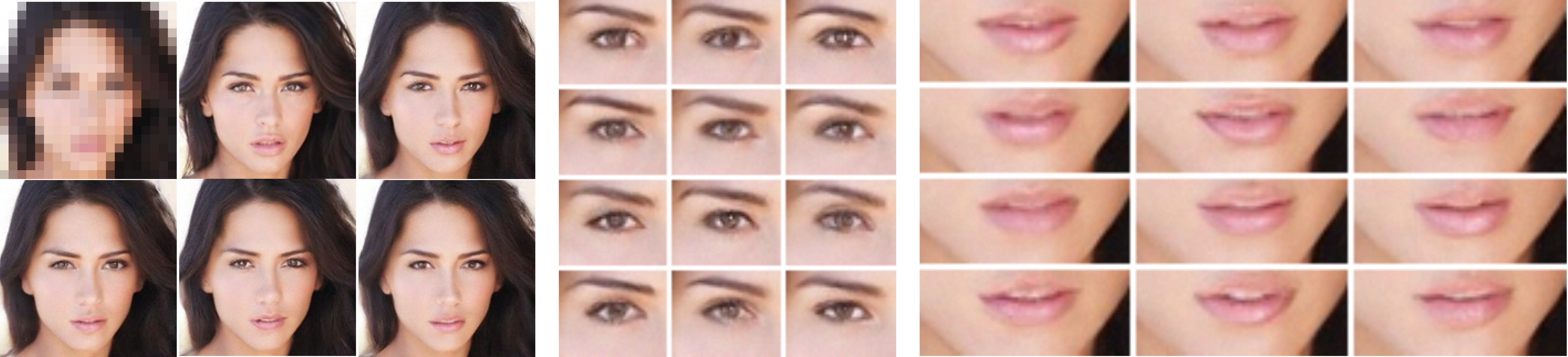}

\vspace{-2mm}
\caption{Random SR samples generated by SRFlow using the given LR image on CelebA ($8\times$).}
\vspace{-2mm}
\label{fig:sampling_CMbic8}
\end{figure}

\begin{figure}
\centering

\includegraphics[width=\linewidth]{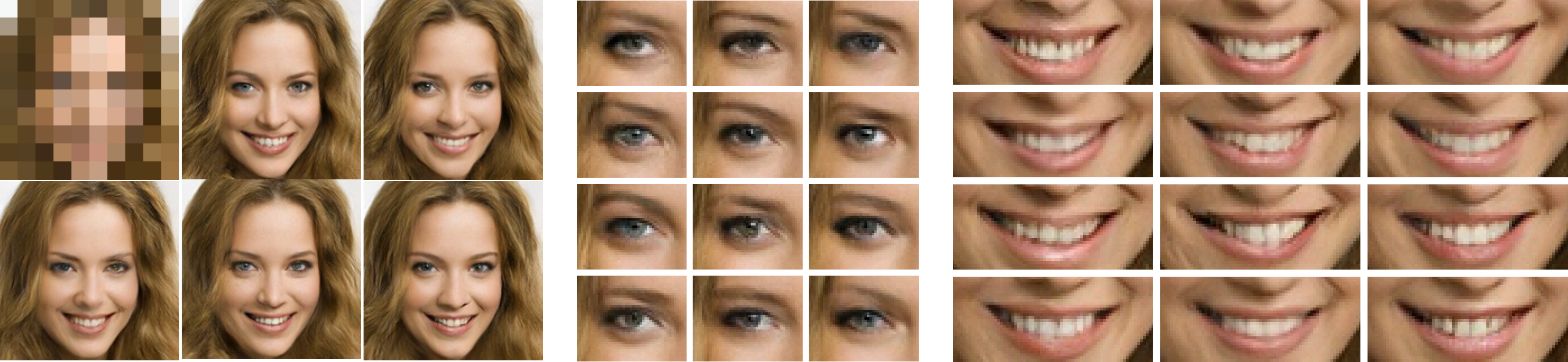}

\vspace{-2mm}
\caption{Random SR samples generated by SRFlow using the given LR image on CelebA ($16\times$).}
\vspace{-2mm}
\label{fig:sampling_CMbic16}
\end{figure}

Here we provide additional examples to show the variety when sampling SR images with our default temperature $\tau = 0.8$ for CelebA.
As seen for $8\times$ super-resolution sampling in Figure~\ref{fig:sampling_CMbic8}, the low resolution image still contains significant information about facial characteristics.
This bounds the diversity of super-resolution in order to be consistent.
On the other hand in Figure~\ref{fig:sampling_CMbic16} we show $16\times$ super-resolution which is much more free while still being consistent to the low-resolution.
Therefore one can observe a much higher variety.

\subsection{Stochastic General Super-Resolution}

\begin{figure}[htbp]
\centering

\newcommand{\size}{0.19}
\includegraphics[width=\size\linewidth]{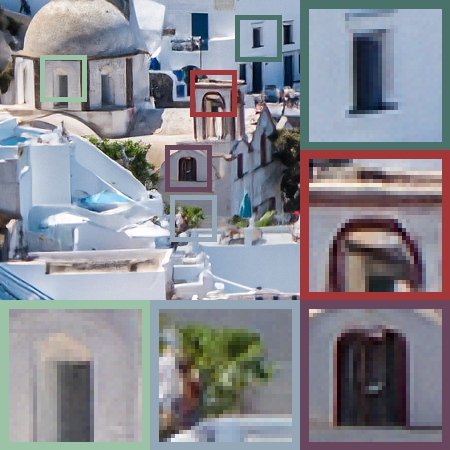}
\includegraphics[width=\size\linewidth]{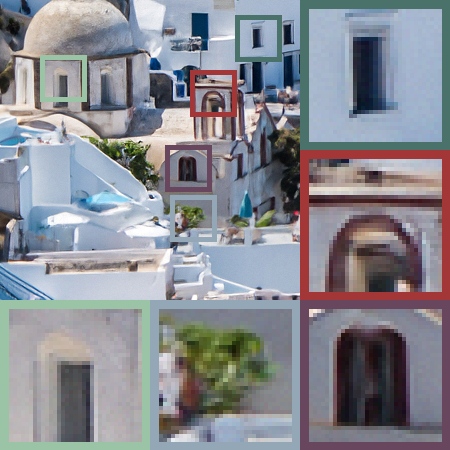}
\includegraphics[width=\size\linewidth]{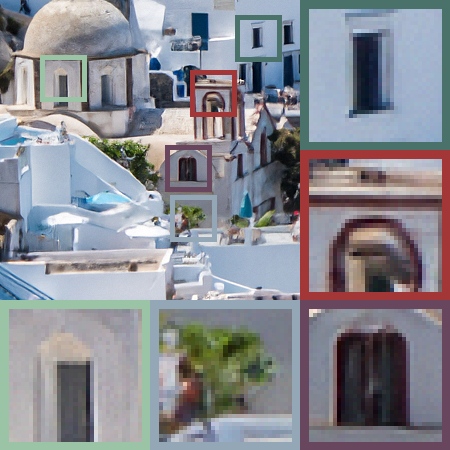}
\includegraphics[width=\size\linewidth]{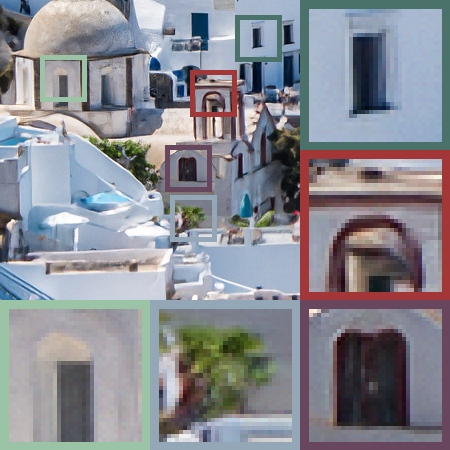}
\includegraphics[width=\size\linewidth]{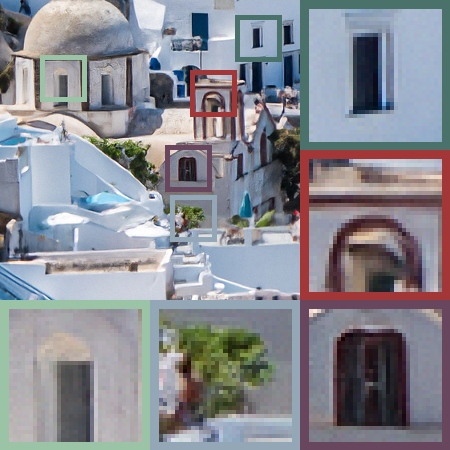}

\vspace{0mm}
\caption{Random SR samples generated by SRFlow using the given LR image on DIV2K ($4\times$).}
\vspace{0mm}
\label{fig:diversity_div2k4}
\end{figure}
\begin{figure}[htbp]
\centering

\newcommand{\size}{0.19}
\includegraphics[width=\size\linewidth]{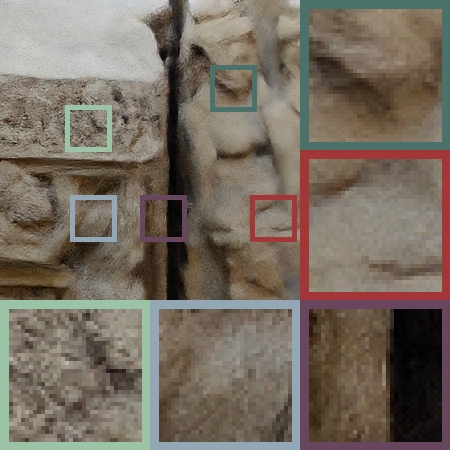}
\includegraphics[width=\size\linewidth]{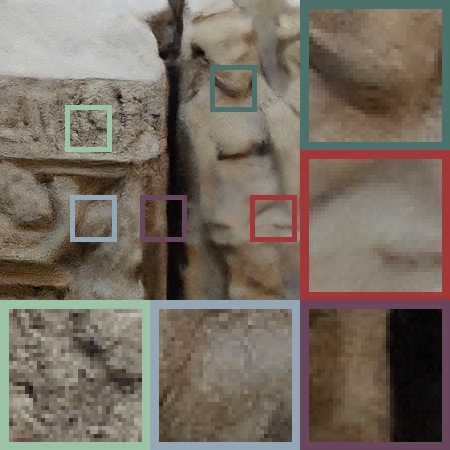}
\includegraphics[width=\size\linewidth]{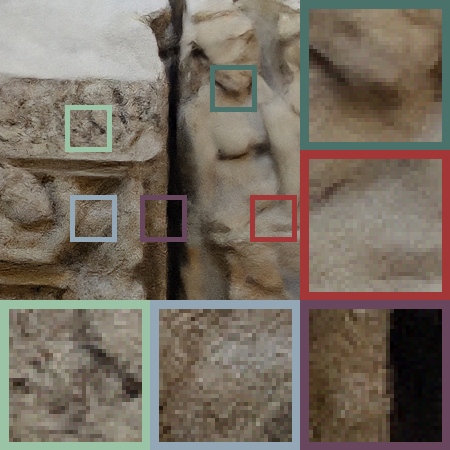}
\includegraphics[width=\size\linewidth]{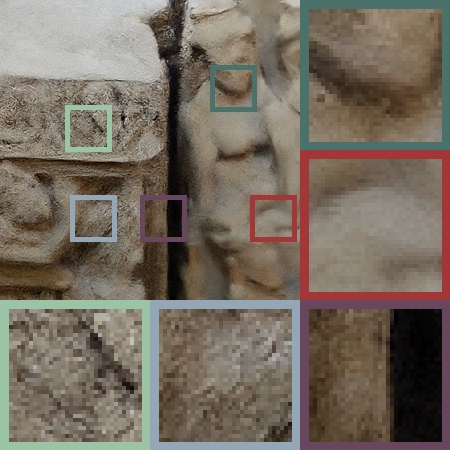}
\includegraphics[width=\size\linewidth]{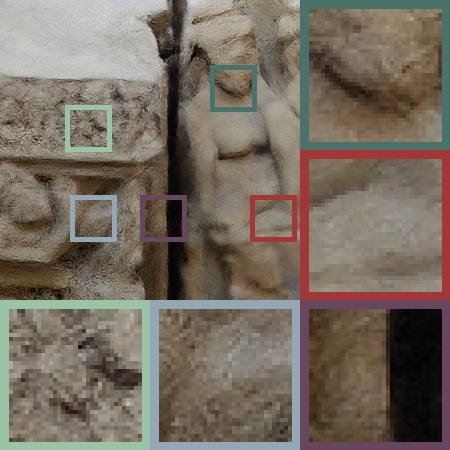}

\vspace{0mm}
\caption{Random SR samples generated by SRFlow using the given LR image on DIV2K ($8\times$).}
\vspace{0mm}
\label{fig:diversity_div2k8}
\end{figure}

In analogy to the visual sampling experiments for CelebA, we show results for the same procedure applied to DIV2K. An example for the variety of upscaling factor $4\times$ is shown in Figure~\ref{fig:diversity_div2k4}. For example, one can observe that the door in the lower right sometimes looks more like an archway and other examples more square. In addition we show the results for $8\times$ upsampling in Figure~\ref{fig:diversity_div2k8}. There it can be observed that the texture of the stones varies from being smooth to being rough.

\subsection{Image Content Transfer}
\begin{figure}[t]
\centering%
\newcommand{\sizeL}{0.24}%
\newcommand{\free}{0.013}%
\newcommand{\imggt}[1]{%
\includegraphics[width=\sizeL\linewidth]{figures_sup/rgbGtTrans/#1_source}\hspace{\free\linewidth}%
\includegraphics[width=\sizeL\linewidth]{figures_sup/rgbGtTrans/#1_target}\hspace{\free\linewidth}%
\includegraphics[width=\sizeL\linewidth]{figures_sup/rgbGtTrans/#1_rgb}\hspace{\free\linewidth}%
\includegraphics[width=\sizeL\linewidth]{figures_sup/rgbGtTrans/#1_mod}
}%
\imggt{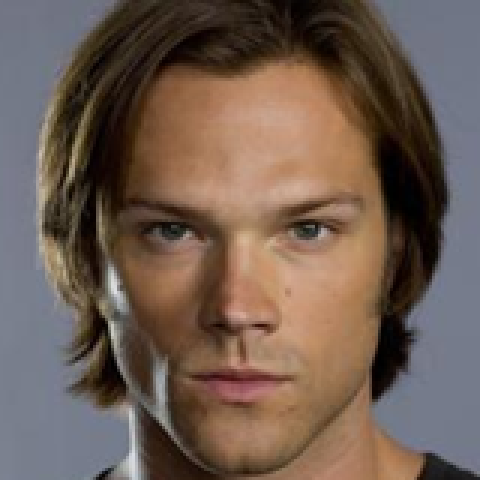}
\imggt{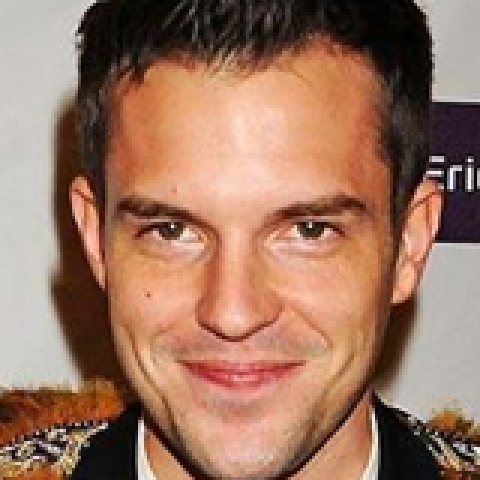}
\resizebox{\linewidth}{!}{
    \hspace{-2mm}\begin{tabular}{@{}C{2cm} C{2cm} C{2cm} C{2cm}}
    Source & Target $\by$ & Input $\tilde{\by}$ & Transferred $\hat{\by}$
\end{tabular}}

\caption{Image content transfer for an existing HR image (top) and an SR prediction (bottom). Content from the source is applied directly to the target. By applying latent space normalization in our SRFlow, the content is integrated.}%
\label{fig:sup-content-transfer}
\end{figure}

Additional examples for image content transfer are depicted in Figure~\ref{fig:sup-content-transfer}. For this task we trained SRFlow with random shifts of 4px in HR to obtain a higher flexibility.

\end{document}